\documentclass[final]{cvpr}
\pdfoutput=1
\usepackage{times}
\usepackage{epsfig}
\usepackage{graphicx}
\usepackage{amsmath}
\usepackage{amssymb}
\usepackage{subcaption}
\usepackage{booktabs}       % professional-quality tables
\usepackage[numbers]{natbib}
\usepackage[pagebackref=true,breaklinks=true,colorlinks,bookmarks=false]{hyperref}

\def\cvprPaperID{4394} % *** Enter the CVPR Paper ID here
\def\confYear{CVPR 2021}

\begin{document}

%%%%%%%%% TITLE
\title{Query-Free Adversarial Transfer via Undertrained Surrogates}

\author{Chris Miller\\
  Department of Computer Science\\
  Dartmouth College\\
  Hanover, NH 03755 \\
  {\tt\small Chris.20@Dartmouth.edu}
  \and
  Soroush Vosoughi\\
  Department of Computer Science\\
  Dartmouth College\\
  Hanover, NH 03755 \\
  {\tt\small Soroush.Vosoughi@Dartmouth.edu}
}

\maketitle

%%%%%%%%% ABSTRACT
\begin{abstract}
    Deep neural networks are vulnerable to adversarial examples---minor perturbations added to a model's input which cause the model to output an incorrect prediction. We introduce a new method for improving the efficacy of adversarial attacks in a black-box setting by undertraining the surrogate model which the attacks are generated on. Using two datasets and five model architectures, we show that this method transfers well across architectures and outperforms state-of-the-art methods by a wide margin. We interpret the effectiveness of our approach as a function of reduced surrogate model loss function curvature and increased universal gradient characteristics, and show that our approach reduces the presence of local loss maxima which hinder transferability. Our results suggest that finding strong single surrogate models is a highly effective and simple method for generating transferable adversarial attacks, and that this method represents a valuable route for future study.
\end{abstract}

%%%%%%%%% BODY TEXT

\section{Introduction}
Previous work has shown that deep learning models are vulnerable to adversarial perturbations \cite{szegedy2013intriguing}. These are small modifications to an input image which cause the model to output an incorrect prediction. 

Understanding adversarial examples is an important task. When deep models are applied in security-conscious domains such as autonomous driving, healthcare, and fraud detection, their vulnerabilities to attack become vulnerabilities which can threaten individual health and safety. This threat becomes especially salient with the advent of adversarial examples which can fool models in the physical world \cite{kurakin2016adversarial,eykholt2018robust}. 

Currently the most consistently successful method for defending models is adversarial training, which entails training on adversarial examples in addition to (or instead of) clean images \cite{goodfellow2014explaining}. However, adversarial training methods are typically computationally expensive, reduce model accuracy on clean inputs, and only provide limited security, suggesting that better methods are needed \cite{goodfellow2014explaining, madry2017towards}. 

\section{Adversarial transfer}
Adversarial attacks can be grouped into white and black-box attacks. In white-box attacks, the attacker has access to the parameters and architecture of the target model, allowing them to utilize model gradients and losses. In black-box attacks, the adversary has no access to the parameters of the target model, and may or may not have access to its architecture or outputs (predictions and logits) for a given input.

Szegedy et al. showed that adversarial examples have the ability to transfer between models, enabling an example generated on a seen model (known as the surrogate model) to fool an unseen model (known as the target model) \cite{szegedy2013intriguing}. This effect means that keeping a model's parameters and query access private is not an effective way to protect a model from adversarial attack. As long as an adversary has the ability to train a model on a compatible dataset, the adversary can use their own model as a surrogate to produce adversarial examples which may then fool the target classifier. 

Many methods for producing effective black-box transfer rely on access to the outputs of the target model \cite{bhagoji2018practical}. Access to target model outputs allows for a class of attacks known as gradient estimation attacks, in which the adversary attempts to estimate the gradients of the target model in order to approximate a white-box attack in a black-box setting (using, for example, a finite-difference method). Required access to the target model can range from the predicted label for an input to class-conditional probability predictions for all classes. Recent approaches to gradient-estimation attacks include \cite{cheng2019query}, \cite{cheng2019improving}, \cite{ilyas2019prior}, and \cite{tu2019autozoom}. Other approaches such as that of Moon et al. use query access to construct adversarial examples without gradient estimation \cite{moon2019parsimonious}. This approach is not always representative of real-world model access. In real-world settings, models may be secured by restricting access to outputs or limiting the permitted number of queries. In these settings, an adversary may not have access to the predictions for any inputs. This led Ilyas et al. to introduce more restrictive black-box settings in \cite{ilyas2018black}.

We use the restrictive zero-knowledge, zero-access setting, in which an adversary has no knowledge of model architecture or parameters, and has no access to model outputs. An adversary which can effectively fool a target model in this setting is extremely strong. One approach in this setting focuses on building attacks on ensembles of surrogate classifiers \cite{liu2016delving}. This approach reduces the overfit of attacks to the surrogate model, and increases their ability to transfer to the target model. A limitation of this approach is the computational requirements for an adversary to train multiple diverse classifiers and build adversarial examples using them in parallel. % add citation of ensemble surrogate classifiers and more limitations 

Recent attempts to produce more successful transfer attacks have focused on creating stronger attack generation methods. The current state of the art in query-free adversarial transfer is the Intermediate Level Attack (ILA) introduced by \citet{huang2019enhancing}. They show that enhancing a previously generated adversarial example by increasing its perturbation on a certain layer of the surrogate model substantially improves transfer to target models. 

Our work, by contrast, focuses on finding a more effective individual surrogate model. We show that generating simple attacks on a more effective surrogate produces stronger transferability than generating more sophisticated attacks on a less effective surrogate. We suggest that further research into finding highly effective surrogate models may be a promising avenue for producing strong transfer attacks and accurately assessing the true robustness of existing models to black-box attack. We provide extensive analysis on the CIFAR-10 dataset, and validate our results with analysis on ImageNet \cite{deng2009imagenet,krizhevsky2009learning}.
\section{Attack methods}
\label{sec:methodology_attacks}
A variety of attack strategies have been used in the past to produce transferable adversarial attacks. We evaluate our models against a variety of attacks, and show them below. Our results are based on standard implementations of the attacks released by Huang et al. and the Cleverhans team \cite{huang2019enhancing,papernot2016cleverhans}. We use $\epsilon = .05$ for all attacks. Note that because our work focuses on finding stronger surrogate models for adversarial transfer rather than finding stronger attack generation methods, we use previously developed attacks to evaluate our proposed method for generating transferable attacks.

The \textbf{Fast Gradient Sign Method} (FGSM) was introduced by Goodfellow et al. as a simple method for producing adversarial examples efficiently \cite{goodfellow2014explaining}. The adversarial perturbation is generated by scaling the sign of the model's gradient by $\epsilon$, and this perturbation is added to the original image to form the adversarial example. The adversarial image, $\hat{x}$, is defined by \ref{eq:fgsm}, where $\nabla \ell(x)$ is the gradient of the model loss with respect to input $x$.

\begin{equation} \label{eq:fgsm}
    \hat{x} = x + \epsilon \cdot \textrm{sign}(\nabla \ell (x))
\end{equation}

\textbf{Iterative FGSM} (I-FGSM), also referred to as the Basic Iterative Method, is a simple extension to the FGSM attack, introduced by \cite{kurakin2016adversarial}. This method applies FGSM repeatedly to produce a more finely targeted adversarial example. The attack is defined in Equation~\ref{eq:ifgsm}, where $\hat{x_n}$ indicates the adversarial example produced by $n$ steps of I-FGSM and $\alpha$ indicates the learning rate. Here the function Clip restricts the adversarial example to remain within the $\epsilon$-ball surrounding $x$.
\begin{equation} \label{eq:ifgsm}
    \hat{x_n} = \textrm{Clip}_{x, \epsilon}(\hat{x_{n-1}} + \alpha \cdot \textrm{sign}(\nabla \ell (x_{n-1})))
\end{equation}
We use a learning rate (the epsilon value during each iteration) of $.005$ and $20$ iterations. These values were determined empirically via grid search to produce the strongest transfer.

\textbf{Momentum I-FGSM} (MI-FGSM) was introduced by \citet{dong2018boosting} to enhance iterative attack transferability. The authors find that incorporating a momentum term in when calculating I-FGSM increases the stability of the attack by reducing its susceptibility to being trapped in a local loss maximum. We use a learning rate of $.005$ and $20$ iterations of attack, with a decay $\mu = 0.9$.

The \textbf{Transferable Adversarial Perturbation} (TAP) attack, introduced by Zhou et al., uses intermediate feature representations to generate an adversarial example \cite{zhou2018transferable}. The TAP attack attempts to maximize the distance between the original image and the adversarial image in the intermediate feature maps of the surrogate model. The authors also show that applying smooth regularization to the resulting perturbation improves transfer between models. 

The \textbf{Intermediate Level Attack} (ILA) was introduced by Huang et al. in 2019 as the state of the art in query-free transfer attacks \cite{huang2019enhancing}. This method takes a predefined adversarial example, created using another method, and enhances its perturbation on intermediate layer representations in the surrogate model. The attack uses the predefined example as a guide towards an adversarial direction. We refer to an ILA attack which enhances an example produced by FGSM as ILA-enhanced FGSM, and follow the same convention for other ILA attacks. For ILA-enhanced iterative attacks, we follow the methodology of the original paper and use ten iterations of the original attack followed by ten iterations of ILA enhancement. For each surrogate, we evaluated ILA attacks based on each possible layer, and found that the optimal source layer to enhance perturbations on was consistent for all epochs of the surrogate model. We find that the optimal layer to target is block 4 for ResNet18 and SENet18, block 0 for MobileNetV2, block 9 for GoogLeNet, and block 6 for DenseNet121, and we report results for ILA based on these layers. These parameters are consistent with the optimal target layers shown by Huang et al. \cite{huang2019enhancing}.

\section{Intermediate epoch transferability}
\subsection{Approach}
\label{sec:approach}
Prior work has often focused on attack generation strategies which improve transferability. These include algorithmic approaches such as MI-FGSM, which introduced a momentum term to prevent the attack from being caught in local loss maxima (creating perturbations which produce high loss on the surrogate model, but not on a target model, due to the imperfect decision boundary of the surrogate), and vr-IGSM, which introduced local gradient smoothing for the same reason \cite{dong2018boosting, wu2018understanding}. They also include novel approaches such as input diversity, in which the attacker transforms the adversarial image during the attack to find a more generalizable perturbation \cite{xie2019improving}.

We instead focus on finding a surrogate model with a generalizable yet low complexity decision boundary. The surrogate model must be generalizable to approximate the decision boundary of the data manifold well (and thus approximate the decision boundary of the target model well). It must also be low complexity, to limit the effect of local loss maxima on the generated attack. If we avoid the surrogate model overfitting to the training data, we can in turn ensure that it learns a highly generalizable, low-complexity function. Learning this type of function maximizes our chance of an attack on the surrogate model generalizing well to the target model, as shown by \cite{demontis2019adversarial}. 
% To maximize generalization of the surrogate model while minimizing decision boundary complexity, we aim to minimize overfit of the surrogate model on the training data.
These observations give rise to a simple method for producing highly transferable attacks: undertraining the surrogate model. 

\subsection{Undertrained Models}

We define a fully trained model as one which has achieved the lowest validation set loss given its architecture, initialization, and training procedure. This definition encompasses the typical training procedure for deep learning models: train a model until validation set loss stops improving, then select the model with the lowest validation set loss (the fully trained model) for final use.

We define an undertrained model using two conditions. First, an undertrained model is the state of a model with a higher validation set loss than the fully trained model. This means that an undertrained model is by definition a less effective model for the initial task. Second, an undertrained model has undergone fewer training steps (been trained for less epochs) than the fully trained model. This condition is applied because we wish to exclude overfit models, which fulfill the first condition of an undertrained model but are unlikely to fulfil the conditions of strong generalization and low complexity defined in Section~\ref{sec:approach}.

%We define an undertrained surrogate model as the condition of the model at an earlier epoch of training with a lower validation set accuracy than the final, fully trained model (ie, the model at the point which has the highest validation set accuracy). We require both conditions, as an epoch with lower validation set accuracy that comes after the epoch with the best accuracy is likely to be overtrained rather than undertrained, and by definition of the fully trained epoch no other epoch can have a higher validation set accuracy. 

\begin{figure}
    \centering
    \begin{minipage}[t]{\columnwidth}
        \raggedleft
        \includegraphics[width=\columnwidth]{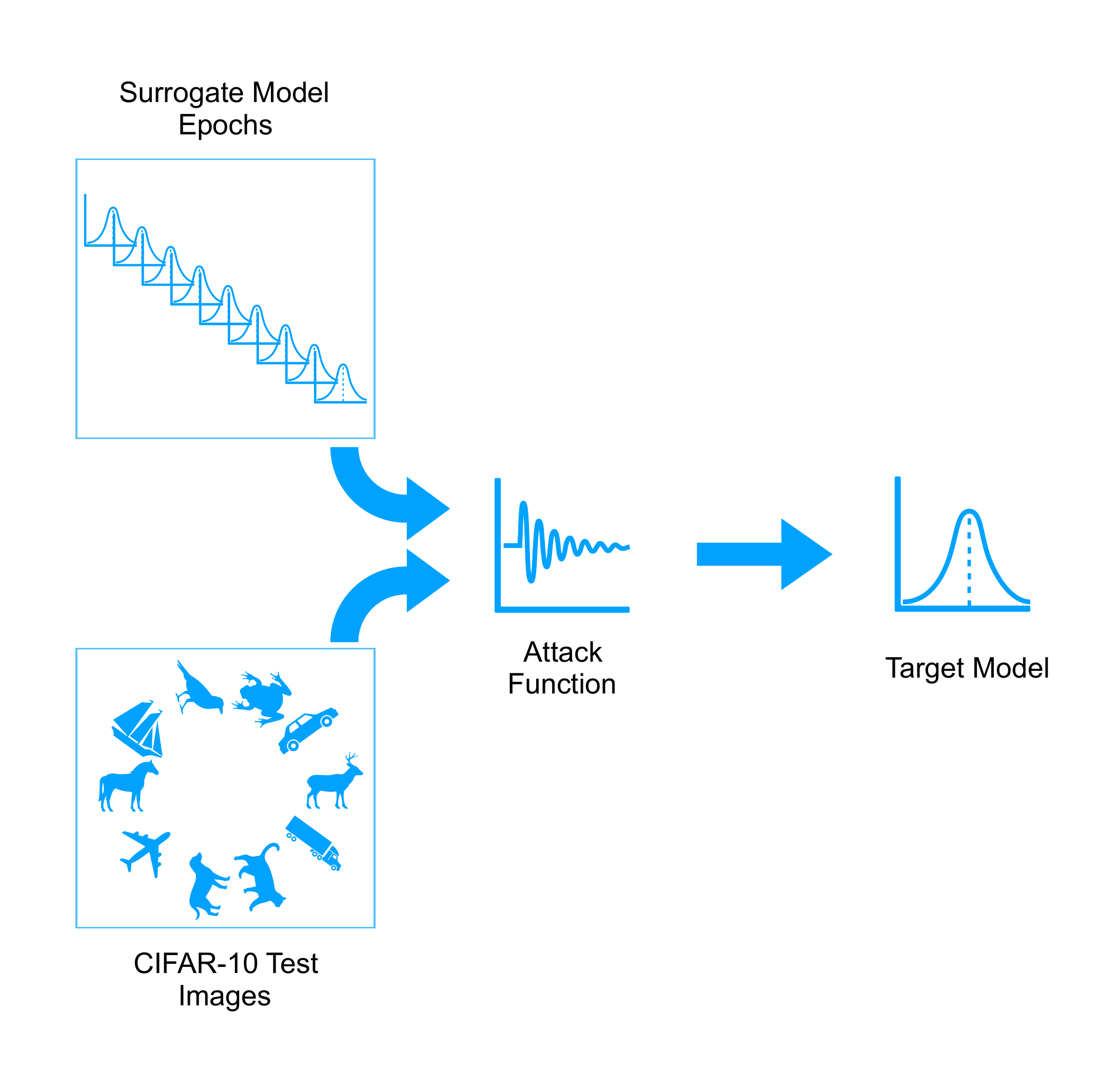}
        \caption{Overview of our methodology for evaluating undertrained adversarial transfer to a target model. Test set images and individual epochs of each surrogate model are inputs to a given attack function. The attack function generates adversarial images, which are then evaluated against a target model.}
         
        \label{fig:methodology}
    \end{minipage}%
    \newline
    \begin{minipage}[t]{\columnwidth}
        \raggedright
        \includegraphics[width=.9\columnwidth]{"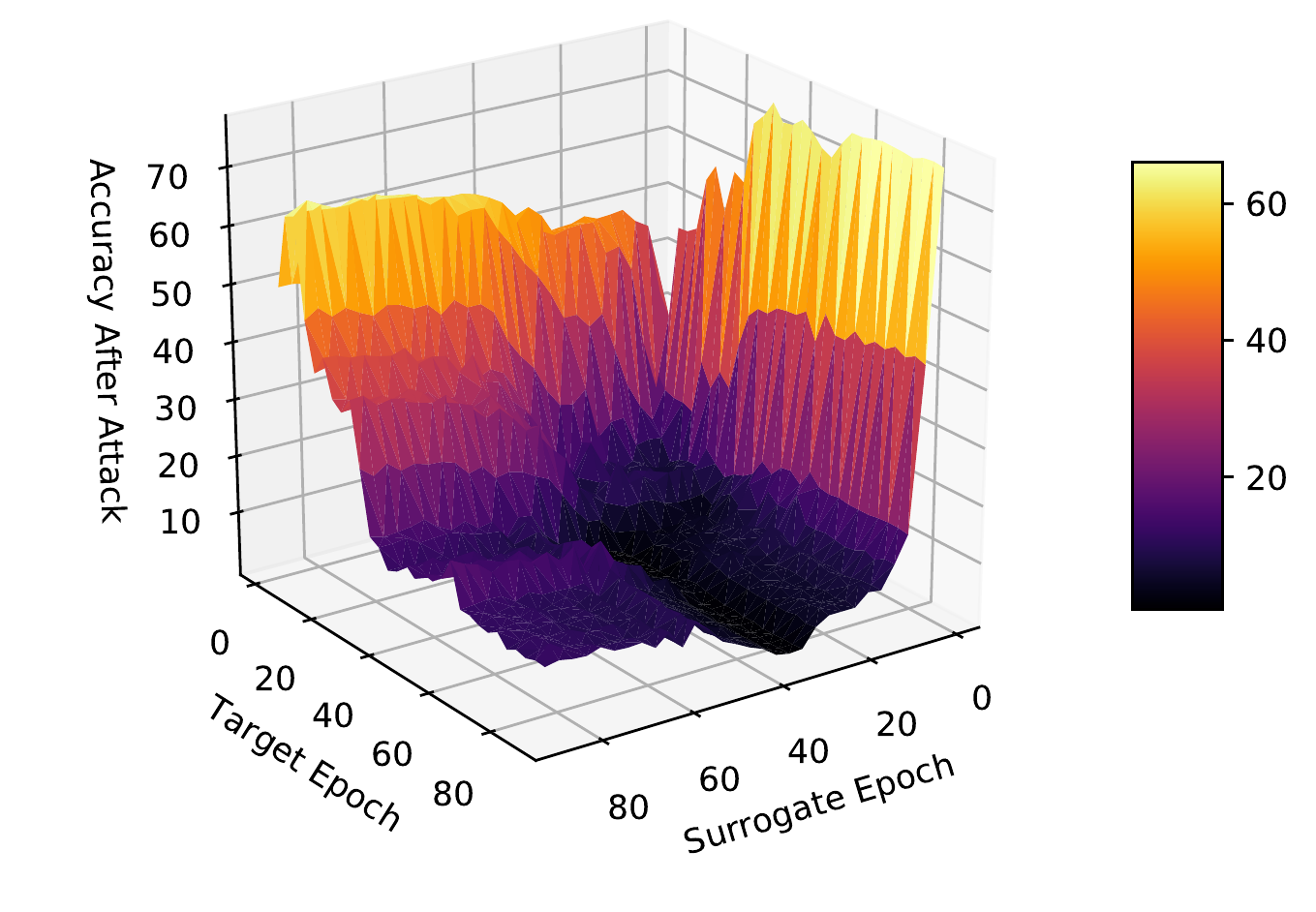"}
        \caption{ResNet18 post-attack accuracy on I-FGSM.}
        \label{fig:3d_transfer}
    \end{minipage}%
    \vspace{-6pt}
\end{figure}

To evaluate this approach, we train surrogate models fully as described in Section~\ref{sec:methodology_models} and save a copy of the model parameters after each epoch of training. This enables retroactive evaluation of the optimal point for building an adversarial attack. We show an overview of our transfer evaluation methodology in Figure~\ref{fig:methodology}.

\section{Experimental Setup}
\label{sec:methodology_models}
We chose to evaluate our approach across a wide variety of model architectures to produce an accurate assessment of transfer between both similar and different models. We report results using ResNet18 models, SENet18 models, GoogLeNet models, DenseNet121 models, and MobileNetV2 models as described by \cite{he2016deep,hu2018squeeze,szegedy2015going,huang2017densely, sandler2018mobilenetv2}. Model architectures are as they is defined in their original papers, with minor changes given by \cite{pytorchtraining} to accommodate the input size of CIFAR-10 images. All models are trained and evaluated on an individual NVIDIA Tesla P100 GPU.

We use the CIFAR-10 dataset for training and testing \cite{krizhevsky2009learning}. This dataset is composed of 60,000 images divided into ten disjoint classes. There are 6,000 images per class, and the dataset is divided into 50,000 training images and 10,000 test images. Each image is a full color $3\times 32\times 32$ image. 

For naturally trained models, we used the definitions and training scripts provided by \cite{pytorchtraining}. Our final trained accuracies are comparable to those reported by \cite{pytorchtraining}.

To ensure that our analysis is comparable to the current state of the art, we use the pretrained target models released by \cite{huang2019enhancing}. These models are a ResNet18 model, an SENet18 model, a DenseNet121 model, and a GoogLeNet model as described by \cite{he2016deep,hu2018squeeze,huang2017densely,szegedy2015going}. As with our surrogate models, these models are as defined by their original authors with the exception of minor modifications to support the $32\times 32$ input size of images. These models are implemented and trained using code released by \cite{pytorchtraining}, consistent with our training.

For adversarially trained models, we use fast adversarial training, introduced by \cite{wong2019fast}, with $\epsilon=.05$ to train ResNet18 models via adversarial training. Fast adversarial training introduces FGSM adversarial examples with random initializations into each minibatch while training, controlling for a phenomenon the authors refer to as catastrophic overfitting to produce robustness to iterative attack. Our results for these models are comparable to those reported by \cite{wong2019fast}, and show that the adversarial training method makes these models quite robust to adversarial examples. 

We describe the training procedures and final accuracies for the CIFAR10 models described above in SM in Section~S3.1 and evaluate the impact of hyperparameter selection in SM in Section~S4.

\section{Results}
\subsection{Naturally trained transfer}

To evaluate the effectiveness of this approach, we tested transfer between two separately trained ResNet18 models across epochs. Both models were trained using the surrogate model training procedure described in \ref{sec:methodology_models}. We evaluated transfer by generating attacks on every third epoch of the first model (ie, the epoch set \{1, 4, 7, ..., 88\}) and transferred them to the all epochs in the same set of the second model. Our results are shown in Figure~\ref{fig:3d_transfer}, where lower accuracy after attack indicates more effective transfer. Although intuition would suggest that attacks generated on a given surrogate epoch would transfer best to the same epoch of target model, our results show that this is not the case. The resulting accuracy landscape instead shows a distinct valley between surrogate epochs 20 and 40, where the same surrogate models transfer well to almost all target model epochs. Also significant is that these epochs outperform later epochs (those which are more fully trained) in transferring to target models. These results validate the hypothesis that undertraining can produce a surrogate model which generates significantly more transferable adversarial examples.

We expand our analysis by evaluating transfer from intermediate epochs of ResNet18, SENet18, MobileNetV2, GoogLeNet, and DenseNet121 models to separately trained target models from the same set of architectures. We consider this wide variety of architectures to evaluate the effectiveness of our approach in a true black-box setting, where the target model architecture is unknown and thus no guarantees can be made about the similarity of the target and surrogate model architectures. Note that the separate training means that attacks between the same architectures are not white box attacks, as the models have different parameters. We generate attacks on every other epoch of the surrogate model, and target the epoch of the target model with the lowest test loss (ie, the fully trained final model). For all models tested (CIFAR-10 and ImageNet), the fully trained epoch was between epochs 80 and 89. We evaluate transfer for a variety of attacks: FGSM, I-FGSM, MI-FGSM, ILA-enhanced I-FGSM, ILA-enhanced MI-FGSM, and Transferable Adversarial Perturbations. For all attacks, we use the parameters outlined in Section~\ref{sec:methodology_attacks}, and use the attack method to form adversarial examples based on all images in the validation set of the dataset.

We compare our approach of generating adversarial examples on intermediate epochs to the previous standard approach of generating adversarial examples on the best loss model. Our findings show that the intermediate-epoch approach produces examples which transfer substantially more successfully than previous approaches across a variety of attack styles. For all attacks evaluated, the intermediate epoch attack outperforms the best epoch attack. Our results for the best intermediate epoch compared to the previous approach using a ResNet18 surrogate model are given in Table~\ref{tab:resnet_transfer}, and we show results for all surrogate models for MI-FGSM attacks in Figure~\ref{fig:natural_transfer}. We provide equivalent graphs for all other attacks in the supplementary material (SM) in Figures~S1-S5.

\begin{figure*}
    \centering
    \includegraphics[width=\textwidth]{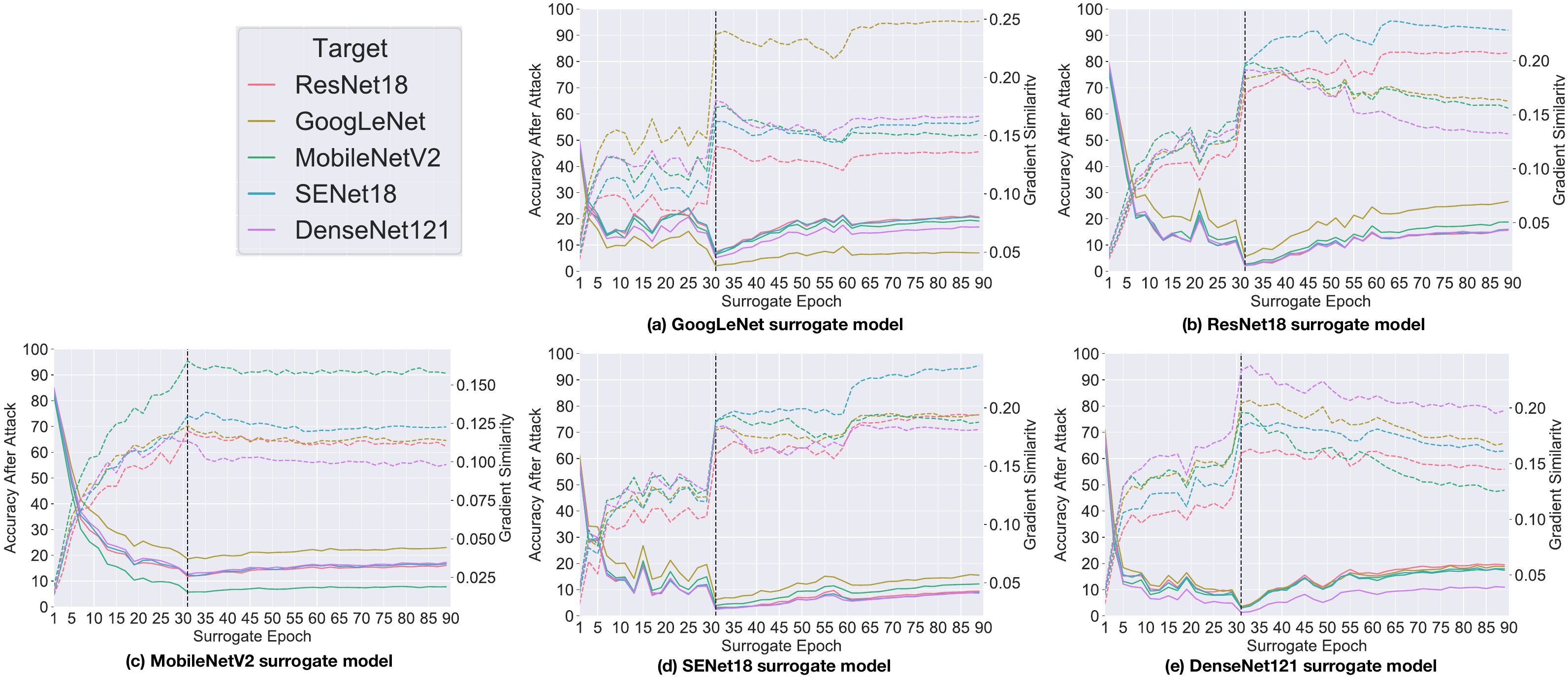}
    \caption{Solid lines indicates post-attack accuracy for MI-FGSM transfer attacks from naturally trained models to target models ($\epsilon=.05$). Lower accuracy indicates better transfer. Dashed colored lines indicate gradient similarity between surrogate and target at each epoch. The black vertical lines indicate the surrogate epoch with the best transferability for most attacks.}
    \label{fig:natural_transfer}
\end{figure*}

\begin{table*}[]
    \centering
    \caption{Top-1 accuracy (\%) (show as Undertrained/Fully Trained) after attack by target model for CIFAR-10 attacks generated using an intermediate surrogate epoch of a ResNet18 model and the surrogate epoch with the lowest validation loss. $\epsilon = .05$. Lower accuracy indicates better transfer.}
    \begin{tabular}{lccccc}
        \toprule
        Target & MI-FGSM & I-FGSM & ILA MI-FGSM & ILA I-FGSM & TAP \\
        \midrule
        ResNet18 & \textbf{2.23}/15.63 & \textbf{2.20}/22.40 & \textbf{2.41}/10.39 & \textbf{2.36}/9.98 & \textbf{13.44}/22.83 \\
        SENet18 & \textbf{2.23}/16.00 & \textbf{2.79}/24.43 & \textbf{2.69}/10.87 & \textbf{2.88}/11.57 & \textbf{12.92}/22.36 \\
        MobileNetV2 & \textbf{2.74}/18.76 & \textbf{2.79}/22.36 & \textbf{2.82}/11.99 & \textbf{2.62}/10.24 & \textbf{12.76}/22.93 \\
        GoogLeNet & \textbf{5.62}/26.64 & \textbf{5.61}/35.42 & \textbf{6.35}/17.09 & \textbf{6.04}/16.76 & \textbf{17.18}/27.96 \\
        DenseNet121 & \textbf{2.24}/15.87 & \textbf{2.13}/21.60 & \textbf{2.57}/10.29 & \textbf{2.44}/9.42 & \textbf{13.67}/21.74 \\
         \bottomrule
    \end{tabular}
    \label{tab:resnet_transfer}
\end{table*}

\begin{table*}[]
    \centering
    \caption{Top-1 accuracy (\%) (show as Undertrained/Fully Trained) after attack by target model for ImageNet MI-FGSM attacks generated using an intermediate surrogate epoch and the surrogate epoch with the lowest validation loss. $\epsilon = .05$. Lower accuracy indicates better transfer.}
    \begin{tabular}{lccccc}
        \toprule
         Source & & & Target & & \\
         \midrule
         & ResNet18 & SENet18 & MobileNetV2 & GoogLeNet & DenseNet121\\
         \midrule
        ResNet18 & \textbf{2.06}/4.43 & \textbf{2.33}/5.49 & \textbf{11.62}/15.59 & \textbf{18.33}/24.14 & \textbf{12.55}/15.10 \\
        SENet18 & \textbf{4.52}/8.47 & \textbf{2.53}/5.63 & \textbf{14.55}/18.03 & \textbf{21.94}/28.21 & \textbf{15.78}/19.05 \\
        MobileNetV2 & 21.07/\textbf{20.93} & \textbf{20.63}/21.03 & 5.29/\textbf{4.55} & \textbf{33.63}/34.33 & 31.50/\textbf{29.59} \\
        GoogLeNet & \textbf{12.28}/19.64 & \textbf{14.14}/21.62 & \textbf{15.41}/21.79 & \textbf{10.09}/18.10 & \textbf{19.54}/25.15 \\
        DenseNet121 & \textbf{5.31}/17.08 & \textbf{6.03}/18.07 & \textbf{11.25}/21.34 & \textbf{18.21}/30.73 & \textbf{2.56}/8.48 \\
         \bottomrule
    \end{tabular}
    \label{tab:resnet_transfer_imagenet}
\end{table*}

% \begin{table}[h]
%     \vspace{-18pt}
%     \centering
%     \caption{Accuracy (\%) (show as Undertrained/Fully Trained) after attack by target model for attacks generated using an intermediate surrogate epoch of a ResNet18 model and the surrogate epoch with the lowest validation loss. $\epsilon = .05$. Lower accuracy indicates better transfer.}
%     \begin{tabular}{lccccc}
%         \toprule
%         Target & MI-FGSM & I-FGSM & ILA MI-FGSM & ILA I-FGSM & TAP \\
%         \midrule
%         ResNet18 & \textbf{2.23}/15.63 & \textbf{2.20}/22.40 & \textbf{2.41}/10.39 & \textbf{2.36}/9.98 & \textbf{13.44}/22.83 \\
%         SENet18 & \textbf{2.23}/16.00 & \textbf{2.79}/24.43 & \textbf{2.69}/10.87 & \textbf{2.88}/11.57 & \textbf{12.92}/22.36 \\
%         MobileNetV2 & \textbf{2.74}/18.76 & \textbf{2.79}/22.36 & \textbf{2.82}/11.99 & \textbf{2.62}/10.24 & \textbf{12.76}/22.93 \\
%         GoogLeNet & \textbf{5.62}/26.64 & \textbf{5.61}/35.42 & \textbf{6.35}/17.09 & \textbf{6.04}/16.76 & \textbf{17.18}/27.96 \\
%         DenseNet121 & \textbf{2.24}/15.87 & \textbf{2.13}/21.60 & \textbf{2.57}/10.29 & \textbf{2.44}/9.42 & \textbf{13.67}/21.74 \\
%          \bottomrule
%     \end{tabular}
%     %\vspace{-14pt}
%     \label{tab:resnet_transfer}
% \end{table}

We show that an intermediate epoch MI-FGSM attack produces the strongest results across most surrogate-target combinations, with the exception of the ResNet18 surrogate, where it performs similarly to an intermediate-epoch I-FGSM attack. We discuss potential reasons for I-FGSM performing similarly to MI-FGSM on the ResNet18 surrogate in \ref{sec:explain}. We selected the strongest surrogate (ResNet18) by choosing the surrogate which produced attacks which performed the best on average across all target models. We choose to focus on the MI-FGSM attack, since it performs the best across the full range of surrogate models. MI-FGSM attacks based on the ResNet18 surrogate reduced accuracy on ResNet18 by 97.65\%, on GoogLeNet by 94.07\%, on MobileNetV2 by 97.02\%, on SENet18 by 97.64\%, and on DenseNet121 by 97.66\%.
%reduced accuracy on GoogLeNet by 89.22 percentage points, on ResNet18 by 92.53 percentage points, on SENet by 92.35 percentage points, on MobileNetV2 by 91.32 percentage points, and on DenseNet121 by 93.35 percentage points. 
These results emphasize the ability of undertrained surrogate attacks to generalize across a wide variety of target model architectures in a black-box setting. To the best of our knowledge, this makes an undertrained MI-FGSM attack (UMI-FGSM) the state of the art in query-free transfer attacks. Our results also show that the optimal surrogate epoch for transfer is consistent across target models for all attacks except TAP. This indicates that an adversary can select the strongest surrogate by evaluating performance on other models, without requiring any query access to the target model, fitting our case of the black-box setting with zero query access.

To validate that our fully trained models are representative of models typically used to generate adversarial attacks, we generate attacks using the CIFAR-10 models released by \citet{huang2019enhancing} (which we use as target models in this work) and transfer them to other target models. We also include our MobileNetV2 target model for completeness, although this was not a model released by Huang et al. We report results in SM in Table~S3. Our results show that attacks generated using these models as surrogates are significantly less effective than attacks generated using our undertrained surrogate models at transferring to the same targets, validating our analysis.

These results show that undertraining a surrogate model is an effective strategy for producing adversarial examples which can transfer well across varied model architectures.

To evaluate whether or not our results are limited to small datasets such as CIFAR-10, we evaluate our approach on the ImageNet dataset \cite{deng2009imagenet}. We find that the effectiveness of an undertrained attack is also present for ImageNet-trained models, showing that this approach is applicable to datasets with large image counts, class counts, and image sizes. We show results for the transferability of MI-FGSM attacks between ImageNet models in Table~\ref{tab:resnet_transfer_imagenet}, and note that we generally achieve reductions of twenty to seventy percent in post-attack accuracy. We do not see this effect strongly for the MobileNetV2 surrogate model, where the undertrained model performs equivalently to the fully trained model. We suggest this is due to the results shown in Figure~\ref{fig:loss_curvature}, which show that the MobileNetV2 architecture does not display a local minimum for local loss curvature at undertrained epochs. These results explains the lower efficacy of undertrained MobileNetV2 surrogate models on both ImageNet and CIFAR-10 (as shown in Figure~\ref{fig:natural_transfer}), and support our conclusion that this effect is partially caused by low local loss curvature for undertrained epochs. We report experimental setup and final accuracies of ImageNet models in SM in Section~S3.2, and report full results of adversarial transfer in SM in Section~S2.
\vspace{-5pt}

\subsection{Adversarially trained transfer}
\vspace{-3pt}

\begin{figure*}[]
\centering
\vspace{-6pt}
    \begin{subfigure}[b]{.99\columnwidth}
        \raggedleft
        \includegraphics[width=.65\columnwidth]{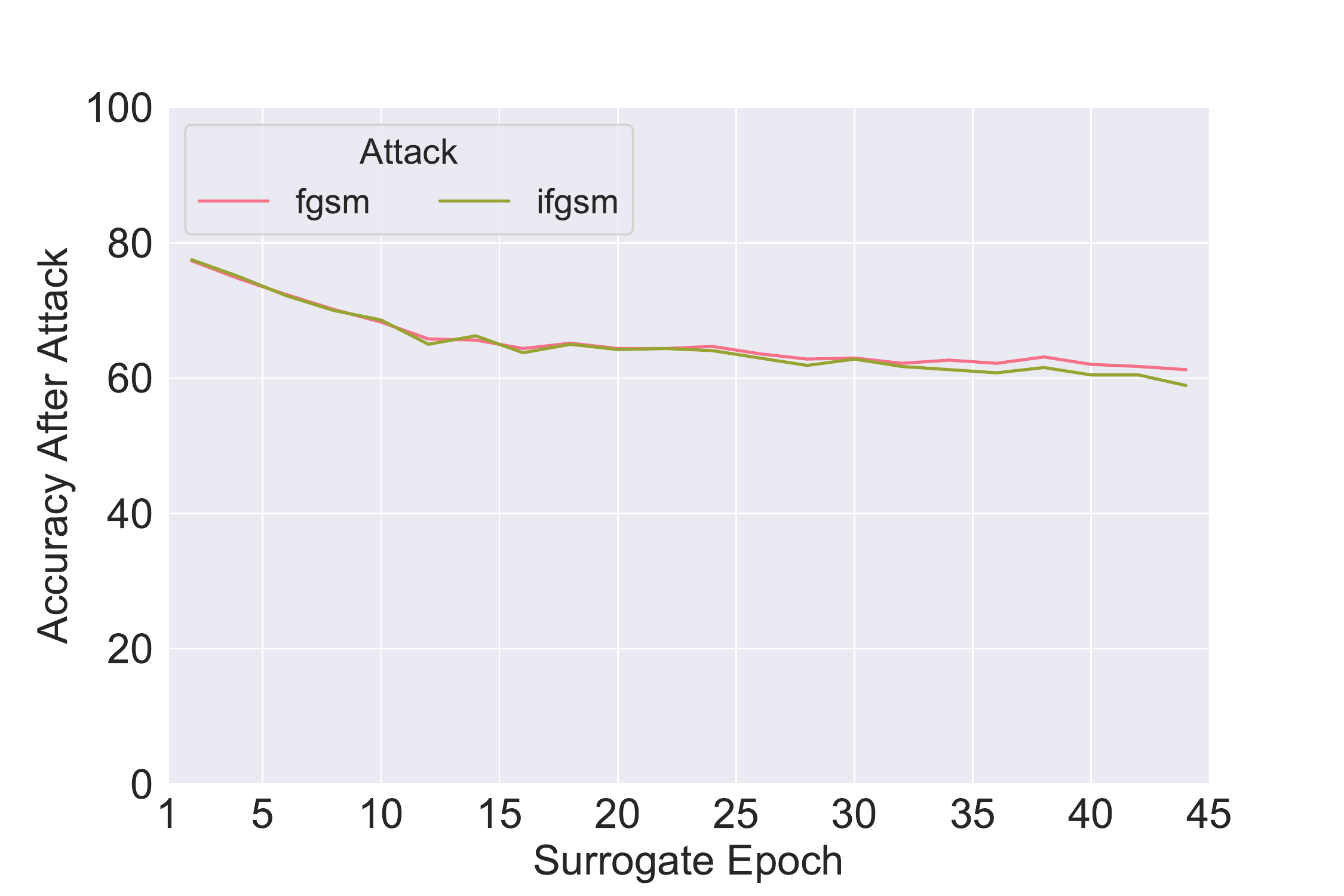}
        \caption{FGSM Trained (Robust)}
        \label{fig:adv_ifgsm_transfer}
    \end{subfigure}
    %\hfill
    \begin{subfigure}[b]{.99\columnwidth}
        \raggedright
        \includegraphics[width=.65\columnwidth]{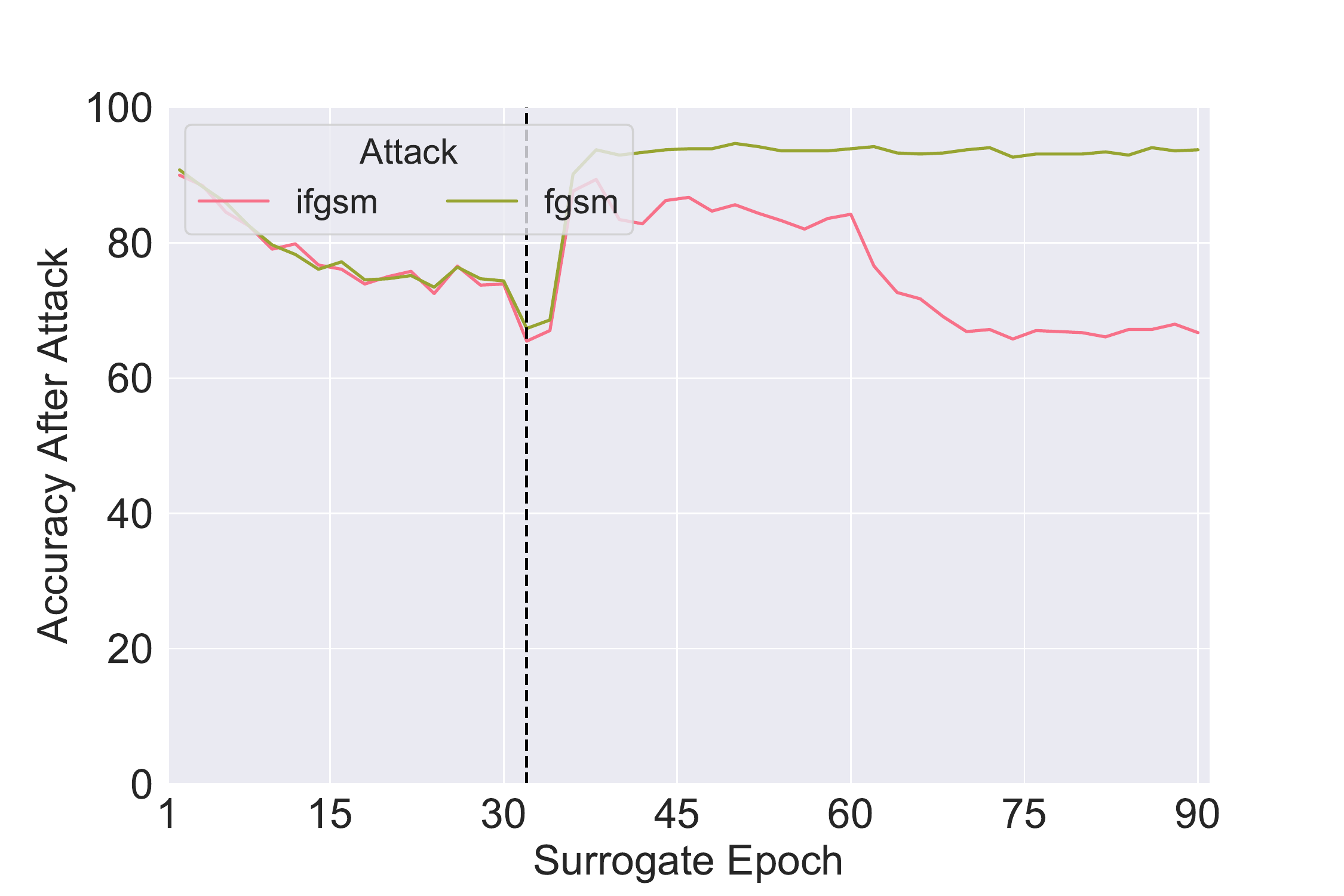}
        \caption{FGSM Trained (Non-Robust)}
        \label{fig:adv_fgsm_transfer}
    \end{subfigure}
    
    \caption{Accuracy of an adversarially trained ResNet18 model against transferred I-FGSM examples from a second adversarially trained ResNet18 model by epoch of the surrogate model, broken out by type of adversarial training. Lower accuracy indicates better transfer.}
    \label{fig:adv_transfer}
    \vspace{-8pt}
\end{figure*}

We next investigated the impact of adversarial training on this phenomenon, providing results for transferred attacks between two ResNet18 models adversarially trained using the Fast-FGSM method introduced by \cite{wong2019fast}, as discussed in Section~\ref{sec:methodology_attacks}. Some prior work has been done here by Vivek \etal, who showed intermediate-epoch adversarial efficacy for non-robust adversarially trained models trained with FGSM examples (discussed there in the context of reducing the cost of ensemble adversarial training) \cite{vivek2018gray}. Our results show that intermediate epoch transferability is restricted to non-robust models (models which are not secure against iterative attacks) such as naturally trained or non-robust FGSM trained models. We report full results on adversarially trained models in Figure~\ref{fig:adv_transfer}. Our results also call into question the conclusion of Vivek \etal that this effect is caused by adversarial training producing models which generate weak adversaries, since we find that this effect is not present in robust models.

\section{Explaining surrogate-based transferability} \label{sec:explain}
Here we evaluate potential causes of improved transferability for certain surrogates, and build an explanatory model for transferability.

\textbf{Gradient similarity} is an important factor in transferability. Models which produce similar gradients on input images will produce similar adversarial images when using gradient-based attacks. We suggest that by undertraining the surrogate classifier, we retain more universal characteristics in the gradients (ie, the model gradients have relatively high similarity to gradients of other models with different architectures), ensuring that the surrogate model gradients will be similar to any target classifier which learns the data manifold regardless of architecture. To evaluate this effect, we calculated the cosine similarity gradients for surrogate and target models on clean images. We calculate gradients on test set images for surrogate and target models. We then reshape each image gradient into a vector, and take the $l_2$ normalized dot product of the target and surrogate model gradients, averaging this value across all images. Here $x_s^i$ represents the reshaped gradient of image $i$ with respect to surrogate model loss and $x_t^i$ represents the same quantity with respect to target model loss:

\begin{equation}
Similarity = \frac{1}{n}\sum_{i=1}^n(\frac{x_s^i x_t^i}{|x_s^i|_2 |x_t^i|_2}).
\end{equation}

We show results for all surrogate and target models in Figure~\ref{fig:natural_transfer}. For all surrogate-target pairs evaluated, gradient similarity is negatively correlated to the post-attack accuracy with $p < .001$. We report correlations in SM in Table~S3. However, the results shown in Figure~\ref{fig:natural_transfer} raise several questions which suggest that gradient correlation is not a perfect proxy for transferability. Note that while the similarity plots for all surrogate/target models exhibit sharp spikes in gradient similarity after epoch 30, which matches transferability improving at that epoch, in many cases gradient similarity does not decline appreciably after the epochs of maximal transferability. Relying on solely gradient direction similarity, we might expect that later epochs would transfer better for some models. These results confirm those of Liu \etal in showing that while gradient similarity between a surrogate and target model is linked with adversarial transferability, other factors appear to be at play \cite{liu2016delving}. 

\begin{figure}
    \centering 
    \includegraphics[width=.8\columnwidth]{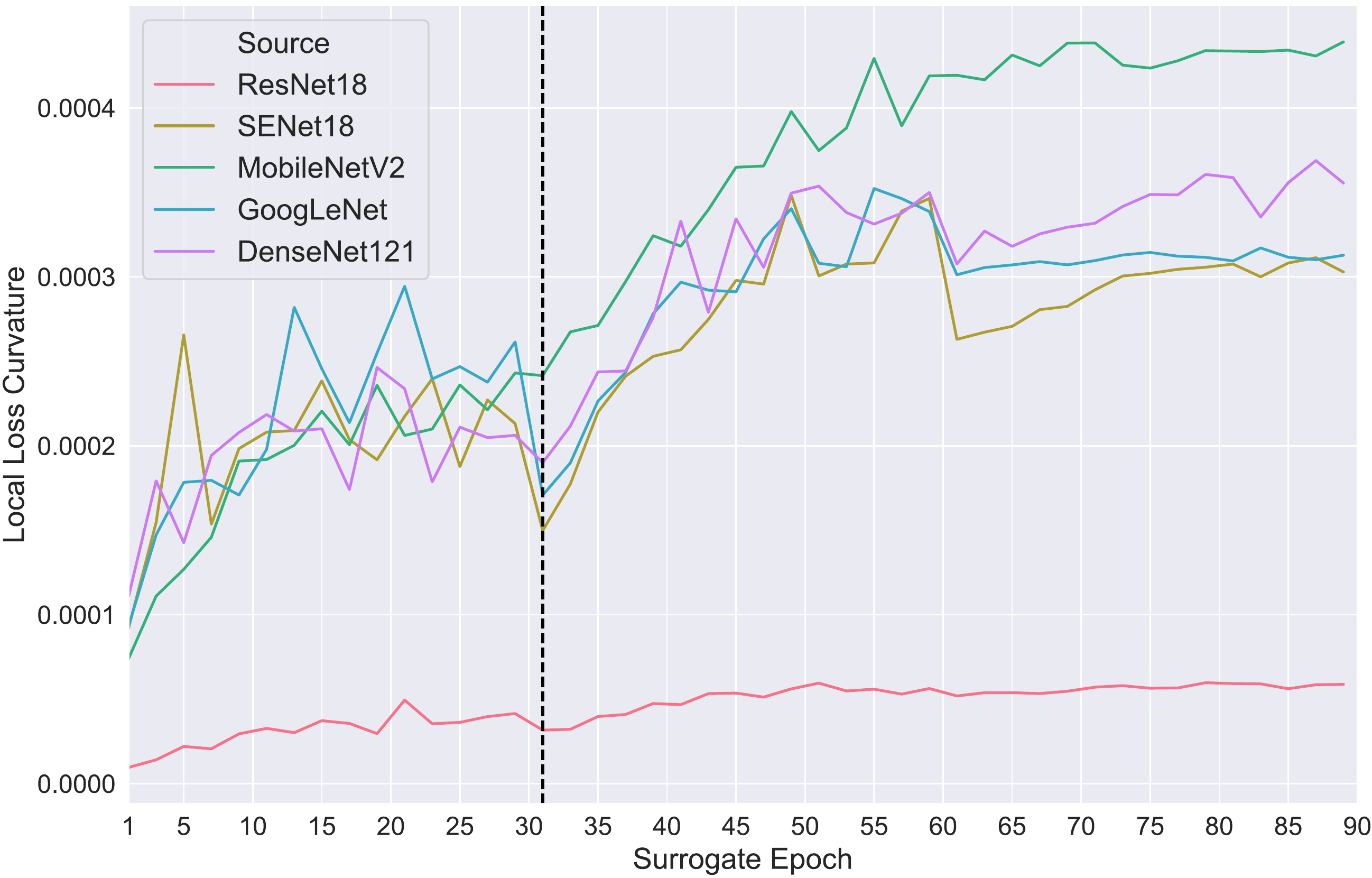}
    \caption{Loss curvature approximation for all surrogate models.}
    \label{fig:loss_curvature}
    %\vspace{-10pt}
\end{figure}

\textbf{Loss function curvature} represents the second part of our explanatory model. Our secondary goal in undertraining is to reduce decision boundary complexity, limiting the effect of local loss maxima on the surrogate model. While decision boundary complexity is computationally challenging to directly quantify, the topology of the decision of a network is closely related to its loss landscape, which we can measure locally \cite{liu2020geometry}. Some prior work has also suggested that local loss function smoothness is related to adversarial transferability, and it also appears to be highly correlated with adversarial robustness \cite{wu2018understanding, moosavi2019robustness}. We evaluate the hypothesis that intermediate epochs have lower complexity than fully trained models. We follow \citet{moosavi2019robustness} in using a finite difference approximation of local loss curvature for a random subset of test data and take the average magnitude of the result for each epoch. Our results show a local minimum in loss curvature at the optimal intermediate epoch for all models except MobileNetV2 (which Figure~\ref{fig:natural_transfer} shows to have the least presence of this effect). For all models, the optimal intermediate epoch curvature is significantly lower than the final model curvature. Figure~\ref{fig:loss_curvature} shows the local loss curvature by epoch for all surrogate models.

To confirm that a reduction in our metric of reduced loss curvature is accurately reflecting a reduction in local loss maxima, we note that the improved performance of MI-FGSM over I-FGSM is based on the ability of MI-FGSM to escape local loss maxima \cite{dong2018boosting}. We thus expect that MI-FGSM will outperform I-FGSM more on models with more local loss maxima (ie, models with a more complex loss landscape). This suggests that if higher curvature indeed implies more local loss maxima and a more complex loss landscape, it will have strong positive correlation with how much MI-FGSM outperforms I-FGSM. We calculate the Pearson correlation coefficient between curvature and how strongly MI-FGSM outperformed I-FGSM (ie, the accuracy of target models on MI-FGSM examples minus their accuracy on I-FGSM examples) for each surrogate model. We found highly statistically significant positive correlation between curvature and how strongly MI-FGSM outperformed I-FGSM (mean coefficient = $0.86$,  $p < .001$ for all surrogate models). This indicates that our approach is successfully producing a surrogate model with a less complex loss landscape, allowing I-FGSM to approach the transferability of MI-FGSM.

These results also suggest an explanation for why the ResNet18 surrogate model outperforms other models in intermediate transfer. Its complexity at all epochs is significantly lower than that of the other models we consider. This result may also explain why ResNet18 is the only surrogate model architecture in which an intermediate I-FGSM attack performs as well or better than the MI-FGSM attack: the lower complexity of the ResNet18 model loss landscape reduces the impact of local loss maxima, eliminating the advantage of the MI-FGSM.

\subsection{Explanatory Model}

\begin{figure}
    \centering
    \captionof{table}{Linear model of post-attack accuracy for all target models ($R^2 = 0.68$)}
    \includegraphics[width=\columnwidth]{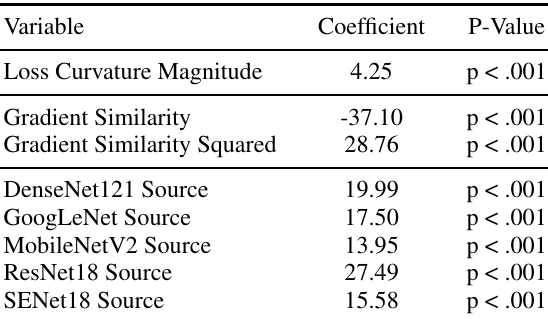}
    \label{tab:linmodel}
    %\vspace{-20pt}
\end{figure}
To evaluate the impact of gradient similarity and curvature, we fit an ordinary least squares linear regression model where the target variable is post-attack accuracy. The model includes gradient similarity, squared gradient similarity, curvature, and a constant for each source model as the independent variables (to account for the overall accuracy of the underlying architecture). We include squared gradient similarity due to the presence of nonlinear residuals in a model without squared gradient similarity. We report coefficients and significance measures in Table~\ref{tab:linmodel}. Our model confirms that gradient similarity and curvature are highly significant ($p < .001$) predictors of transferability. 

%\vspace{-18pt}
% \begin{table}
% \scriptsize
% \caption{Linear Model of Post-Attack Accuracy ($R^2 = 0.681$)}
% \centering
% \begin{tabular}{ lcc } \\
%  \toprule
%  Variable & Coefficient & P-Value \\
%  \midrule
%  Loss Curvature Magnitude & 4.25 & p < .001\\
%  \midrule
%  Gradient Similarity & -37.10 & p < .001\\ 
%  Gradient Similarity Squared & 28.76 & p < .001\\
%  \midrule
%  DenseNet121 Source & 19.99 & p < .001\\
%  GoogLeNet Source & 17.50 & p < .001\\
%  MobileNetV2 Source & 13.95 & p < .001\\
%  ResNet18 Source & 27.49 & p < .001\\
%  SENet18 Source & 15.58 & p < .001\\
%  \bottomrule
% \end{tabular}
% \label{tab:linmodel}
% \end{table}
\section{Discussion} 
We find that adversarial examples generated on an undertrained surrogate model transfer significantly more successfully than attacks generated on fully trained models. Our MI-FGSM attacks generated on undertrained models outperform the current state of the art in query-free black-box transfer. Our results clearly show that a new focus on finding strong single surrogate models with low local loss curvature can produce state-of-the-art results for adversarial transfer. We explain intermediate epoch transferability as the result of two effects: universal gradient characteristics and low loss function curvature. We show that gradient similarity and loss function curvature are highly significant ($p < .001$) predictors of transferability.

We also note that this result has important implications for the analysis of model robustness to black-box attack. Prior work has assumed that surrogate models trained with the same architecture and training procedure are the worst-case for adversarial transfer, as suggested by \citet{madry2017towards}. However, our results show that this is not the worst-case. An undertrained surrogate model---even one with a different architecture---can produce attacks which transfer more successfully than those based on fully trained models of the same architecture, reducing post-attack accuracy by more than 75\% compared to the previous assumption of worst-case black-box transfer. This indicates that prior robustness analyses underestimate the risk of black-box transfer attacks. Our work also finds that this effect is not present in robust models, though it is present to some degree in non-robust adversarially trained models, confirming some of the results of \citet{vivek2018gray}. Further work here may provide insights into how the training process of robust models differs from that of non-robust models.

\section{Conclusion and Future Work}
% New
Our results show that a simple approach focused on surrogate model, rather than attack method, outperforms prior methods for producing transferable adversarial attacks. We show that this surrogate-focused approach to adversarial example generation creates attacks which transfer well across architectures and models, while requiring no query access to the target model. The undertrained surrogate attack outperforms the prior state-of-the-art, ILA-enhanced MI-FGSM, by seven to ten percentage points, reducing target classifier performance to below random chance accuracy. Our findings indicate a gap in existing understanding of both adversarial transferability and intermediate epoch models, and show that stronger surrogate models represent an open area of investigation for improvements in transfer attacks. Our findings also reveal that the previous known worst-case scenario for black-box transfer (a surrogate model with the same architecture and training procedure) is not an accurate representation of the worst-case, and produces highly misleading estimates of model robustness to black-box transfer attacks. Evaluation of other strategies for producing strong surrogates may provide more insight into the mechanics of transferability and the strength of black-box attacks.

Our findings leave open many avenues for future work. First among these is how other choices of surrogate model architecture, regularization, and hyperparameters can impact adversarial transferability. Of the surrogate models we evaluate, ResNet18 produces the strongest transfer to the chosen target models. We suggest in Section~\ref{sec:explain} that this is due to the model's low local loss curvature compared to the other models evaluated. However, it is likely that a more effective surrogate architecture or training method exists. Work on this front would help to identify architectural attributes which produce more universal gradient characteristics and reduce loss curvature. Finally, we suggest that extension of this analysis to different tasks may provide context for how widespread this effect is.

{\small
\bibliographystyle{plainnat}
\bibliography{egbib}

\begin{thebibliography}{32}
\providecommand{\natexlab}[1]{#1}
\providecommand{\url}[1]{\texttt{#1}}
\expandafter\ifx\csname urlstyle\endcsname\relax
  \providecommand{\doi}[1]{doi: #1}\else
  \providecommand{\doi}{doi: \begingroup \urlstyle{rm}\Url}\fi

\bibitem[Bhagoji et~al.(2018)Bhagoji, He, Li, and Song]{bhagoji2018practical}
Arjun~Nitin Bhagoji, Warren He, Bo~Li, and Dawn Song.
\newblock Practical black-box attacks on deep neural networks using efficient
  query mechanisms.
\newblock In \emph{European Conference on Computer Vision}, 2018.

\bibitem[Cheng et~al.(2019{\natexlab{a}})Cheng, Le, Chen, Zhang, Yi, and
  Hsieh]{cheng2019query}
Minhao Cheng, Thong Le, Pin-Yu Chen, Huan Zhang, Jinfeng Yi, and Cho-Jui Hsieh.
\newblock Query-efficient hard-label black-box attack: An optimization-based
  approach.
\newblock In \emph{International Conference on Learning Representations},
  2019{\natexlab{a}}.

\bibitem[Cheng et~al.(2019{\natexlab{b}})Cheng, Dong, Pang, Su, and
  Zhu]{cheng2019improving}
Shuyu Cheng, Yinpeng Dong, Tianyu Pang, Hang Su, and Jun Zhu.
\newblock Improving black-box adversarial attacks with a transfer-based prior.
\newblock In \emph{Advances in Neural Information Processing Systems}, pages
  10932--10942, 2019{\natexlab{b}}.

\bibitem[Demontis et~al.(2019)Demontis, Melis, Pintor, Jagielski, Biggio,
  Oprea, Nita-Rotaru, and Roli]{demontis2019adversarial}
Ambra Demontis, Marco Melis, Maura Pintor, Matthew Jagielski, Battista Biggio,
  Alina Oprea, Cristina Nita-Rotaru, and Fabio Roli.
\newblock Why do adversarial attacks transfer? explaining transferability of
  evasion and poisoning attacks.
\newblock In \emph{USENIX Security Symposium}, 2019.

\bibitem[Deng et~al.(2009)Deng, Dong, Socher, Li, Li, and
  Fei-Fei]{deng2009imagenet}
Jia Deng, Wei Dong, Richard Socher, Li-Jia Li, Kai Li, and Li~Fei-Fei.
\newblock Imagenet: A large-scale hierarchical image database.
\newblock In \emph{2009 IEEE conference on computer vision and pattern
  recognition}, pages 248--255. Ieee, 2009.

\bibitem[Dong et~al.(2018)Dong, Liao, Pang, Su, Zhu, Hu, and
  Li]{dong2018boosting}
Yinpeng Dong, Fangzhou Liao, Tianyu Pang, Hang Su, Jun Zhu, Xiaolin Hu, and
  Jianguo Li.
\newblock Boosting adversarial attacks with momentum.
\newblock In \emph{IEEE Conference on Computer Vision and Pattern Recognition},
  pages 9185--9193, 2018.

\bibitem[Eykholt et~al.(2018)Eykholt, Evtimov, Fernandes, Li, Rahmati, Xiao,
  Prakash, Kohno, and Song]{eykholt2018robust}
Kevin Eykholt, Ivan Evtimov, Earlence Fernandes, Bo~Li, Amir Rahmati, Chaowei
  Xiao, Atul Prakash, Tadayoshi Kohno, and Dawn Song.
\newblock Robust physical-world attacks on deep learning visual classification.
\newblock In \emph{IEEE Conference on Computer Vision and Pattern Recognition},
  2018.

\bibitem[Goodfellow et~al.(2015)Goodfellow, Shlens, and
  Szegedy]{goodfellow2014explaining}
Ian~J Goodfellow, Jonathon Shlens, and Christian Szegedy.
\newblock Explaining and harnessing adversarial examples.
\newblock \emph{International Conference on Learning Representations}, 2015.

\bibitem[He et~al.(2016)He, Zhang, Ren, and Sun]{he2016deep}
Kaiming He, Xiangyu Zhang, Shaoqing Ren, and Jian Sun.
\newblock Deep residual learning for image recognition.
\newblock In \emph{IEEE Conference on Computer Vision and Pattern Recognition},
  2016.

\bibitem[Hu et~al.(2018)Hu, Shen, and Sun]{hu2018squeeze}
Jie Hu, Li~Shen, and Gang Sun.
\newblock Squeeze-and-excitation networks.
\newblock In \emph{IEEE Conference on Computer Vision and Pattern Recognition},
  2018.

\bibitem[Huang et~al.(2017)Huang, Liu, Van Der~Maaten, and
  Weinberger]{huang2017densely}
Gao Huang, Zhuang Liu, Laurens Van Der~Maaten, and Kilian~Q Weinberger.
\newblock Densely connected convolutional networks.
\newblock In \emph{IEEE Conference on Computer Vision and Pattern Recognition},
  pages 4700--4708, 2017.

\bibitem[Huang et~al.(2019)Huang, Katsman, He, Gu, Belongie, and
  Lim]{huang2019enhancing}
Qian Huang, Isay Katsman, Horace He, Zeqi Gu, Serge Belongie, and Ser-Nam Lim.
\newblock Enhancing adversarial example transferability with an intermediate
  level attack.
\newblock In \emph{IEEE International Conference on Computer Vision}, 2019.

\bibitem[Ilyas et~al.(2018)Ilyas, Engstrom, Athalye, and Lin]{ilyas2018black}
Andrew Ilyas, Logan Engstrom, Anish Athalye, and Jessy Lin.
\newblock Black-box adversarial attacks with limited queries and information.
\newblock In \emph{International Conference on Machine Learning}, volume~80,
  2018.

\bibitem[Ilyas et~al.(2019)Ilyas, Engstrom, and Madry]{ilyas2019prior}
Andrew Ilyas, Logan Engstrom, and Aleksander Madry.
\newblock Prior convictions: Black-box adversarial attacks with bandits and
  priors.
\newblock In \emph{International Conference on Learning Representations}, 2019.

\bibitem[Krizhevsky et~al.(2009)Krizhevsky, Hinton,
  et~al.]{krizhevsky2009learning}
Alex Krizhevsky, Geoffrey Hinton, et~al.
\newblock Learning multiple layers of features from tiny images.
\newblock 2009.

\bibitem[Kurakin et~al.(2016)Kurakin, Goodfellow, and
  Bengio]{kurakin2016adversarial}
Alexey Kurakin, Ian Goodfellow, and Samy Bengio.
\newblock Adversarial examples in the physical world.
\newblock In \emph{International Conference on Learning Representations}, 2016.

\bibitem[Liu(2020)]{liu2020geometry}
Bo~Liu.
\newblock Geometry and topology of deep neural networks' decision boundaries.
\newblock \emph{arXiv preprint arXiv:2003.03687}, 2020.

\bibitem[Liu(2018)]{pytorchtraining}
Kuang Liu.
\newblock Pytorch cifar10 training, 2018.
\newblock URL \url{https://github.com/kuangliu/pytorch-cifar}.

\bibitem[Liu et~al.(2017)Liu, Chen, Liu, and Song]{liu2016delving}
Yanpei Liu, Xinyun Chen, Chang Liu, and Dawn Song.
\newblock Delving into transferable adversarial examples and black-box attacks.
\newblock In \emph{International Conference on Learning Representations}, 2017.

\bibitem[Madry et~al.(2018)Madry, Makelov, Schmidt, Tsipras, and
  Vladu]{madry2017towards}
Aleksander Madry, Aleksandar Makelov, Ludwig Schmidt, Dimitris Tsipras, and
  Adrian Vladu.
\newblock Towards deep learning models resistant to adversarial attacks.
\newblock In \emph{International Conference on Learning Representations}, 2018.

\bibitem[Moon et~al.(2019)Moon, An, and Song]{moon2019parsimonious}
Seungyong Moon, Gaon An, and Hyun~Oh Song.
\newblock Parsimonious black-box adversarial attacks via efficient
  combinatorial optimization.
\newblock In \emph{International Conference on Machine Learning}, pages
  4636--4645, 2019.

\bibitem[Moosavi-Dezfooli et~al.(2019)Moosavi-Dezfooli, Fawzi, Uesato, and
  Frossard]{moosavi2019robustness}
Seyed-Mohsen Moosavi-Dezfooli, Alhussein Fawzi, Jonathan Uesato, and Pascal
  Frossard.
\newblock Robustness via curvature regularization, and vice versa.
\newblock In \emph{IEEE Conference on Computer Vision and Pattern Recognition},
  pages 9078--9086, 2019.

\bibitem[Papernot et~al.(2016)Papernot, Goodfellow, Sheatsley, Feinman, and
  McDaniel]{papernot2016cleverhans}
Nicolas Papernot, Ian Goodfellow, Ryan Sheatsley, Reuben Feinman, and Patrick
  McDaniel.
\newblock cleverhans v1.0.0: an adversarial machine learning library.
\newblock \emph{arXiv preprint arXiv:1610.00768}, 2016.

\bibitem[Sandler et~al.(2018)Sandler, Howard, Zhu, Zhmoginov, and
  Chen]{sandler2018mobilenetv2}
Mark Sandler, Andrew Howard, Menglong Zhu, Andrey Zhmoginov, and Liang-Chieh
  Chen.
\newblock Mobilenetv2: Inverted residuals and linear bottlenecks.
\newblock In \emph{IEEE Conference on Computer Vision and Pattern Recognition},
  2018.

\bibitem[Szegedy et~al.(2014)Szegedy, Zaremba, Sutskever, Bruna, Erhan,
  Goodfellow, and Fergus]{szegedy2013intriguing}
Christian Szegedy, Wojciech Zaremba, Ilya Sutskever, Joan Bruna, Dumitru Erhan,
  Ian Goodfellow, and Rob Fergus.
\newblock Intriguing properties of neural networks.
\newblock In \emph{International Conference on Learning Representations}, 2014.

\bibitem[Szegedy et~al.(2015)Szegedy, Liu, Jia, Sermanet, Reed, Anguelov,
  Erhan, Vanhoucke, and Rabinovich]{szegedy2015going}
Christian Szegedy, Wei Liu, Yangqing Jia, Pierre Sermanet, Scott Reed, Dragomir
  Anguelov, Dumitru Erhan, Vincent Vanhoucke, and Andrew Rabinovich.
\newblock Going deeper with convolutions.
\newblock In \emph{IEEE Conference on Computer Vision and Pattern Recognition},
  2015.

\bibitem[Tu et~al.(2019)Tu, Ting, Chen, Liu, Zhang, Yi, Hsieh, and
  Cheng]{tu2019autozoom}
Chun-Chen Tu, Paishun Ting, Pin-Yu Chen, Sijia Liu, Huan Zhang, Jinfeng Yi,
  Cho-Jui Hsieh, and Shin-Ming Cheng.
\newblock Autozoom: Autoencoder-based zeroth order optimization method for
  attacking black-box neural networks.
\newblock In \emph{AAAI Conference on Artificial Intelligence}, volume~33,
  pages 742--749, 2019.

\bibitem[Vivek et~al.(2018)Vivek, Reddy~Mopuri, and
  Venkatesh~Babu]{vivek2018gray}
BS~Vivek, Konda Reddy~Mopuri, and R~Venkatesh~Babu.
\newblock Gray-box adversarial training.
\newblock In \emph{European Conference on Computer Vision}, 2018.

\bibitem[Wong et~al.(2019)Wong, Rice, and Kolter]{wong2019fast}
Eric Wong, Leslie Rice, and J~Zico Kolter.
\newblock Fast is better than free: Revisiting adversarial training.
\newblock In \emph{International Conference on Learning Representations}, 2019.

\bibitem[Wu et~al.(2018)Wu, Zhu, Tai, and E]{wu2018understanding}
Lei Wu, Zhanxing Zhu, Cheng Tai, and Weinan E.
\newblock Understanding and enhancing the transferability of adversarial
  examples.
\newblock \emph{arXiv preprint arXiv:1802.09707}, 2018.

\bibitem[Xie et~al.(2019)Xie, Zhang, Zhou, Bai, Wang, Ren, and
  Yuille]{xie2019improving}
Cihang Xie, Zhishuai Zhang, Yuyin Zhou, Song Bai, Jianyu Wang, Zhou Ren, and
  Alan~L Yuille.
\newblock Improving transferability of adversarial examples with input
  diversity.
\newblock In \emph{Proceedings of the IEEE Conference on Computer Vision and
  Pattern Recognition}, pages 2730--2739, 2019.

\bibitem[Zhou et~al.(2018)Zhou, Hou, Chen, Tang, Huang, Gan, and
  Yang]{zhou2018transferable}
Wen Zhou, Xin Hou, Yongjun Chen, Mengyun Tang, Xiangqi Huang, Xiang Gan, and
  Yong Yang.
\newblock Transferable adversarial perturbations.
\newblock In \emph{European Conference on Computer Vision}, 2018.

\end{thebibliography}


\begin{thebibliography}{4}
\providecommand{\natexlab}[1]{#1}
\providecommand{\url}[1]{\texttt{#1}}
\expandafter\ifx\csname urlstyle\endcsname\relax
  \providecommand{\doi}[1]{doi: #1}\else
  \providecommand{\doi}{doi: \begingroup \urlstyle{rm}\Url}\fi

\bibitem[Huang et~al.(2019)Huang, Katsman, He, Gu, Belongie, and
  Lim]{huang2019enhancing}
Qian Huang, Isay Katsman, Horace He, Zeqi Gu, Serge Belongie, and Ser-Nam Lim.
\newblock Enhancing adversarial example transferability with an intermediate
  level attack.
\newblock In \emph{IEEE International Conference on Computer Vision}, 2019.

\bibitem[Liu(2018)]{pytorchtraining}
Kuang Liu.
\newblock Pytorch cifar10 training, 2018.
\newblock URL \url{https://github.com/kuangliu/pytorch-cifar}.

\bibitem[Marcel and Rodriguez(2010)]{marcel2010torchvision}
S{\'e}bastien Marcel and Yann Rodriguez.
\newblock Torchvision the machine-vision package of torch.
\newblock In \emph{Proceedings of the 18th ACM international conference on
  Multimedia}, pages 1485--1488, 2010.

\bibitem[Wong et~al.(2019)Wong, Rice, and Kolter]{wong2019fast}
Eric Wong, Leslie Rice, and J~Zico Kolter.
\newblock Fast is better than free: Revisiting adversarial training.
\newblock In \emph{International Conference on Learning Representations}, 2019.

\end{thebibliography}
}

\end{document}

%%%%%%%%% TITLE
\title{Supplementary Material For Query-Free Adversarial Transfer via Undertrained Surrogates}

\maketitle

%%%%%%%%% BODY TEXT
\setcounter{table}{0}
\renewcommand{\thetable}{S\arabic{table}}%
\setcounter{figure}{0}
\renewcommand{\thefigure}{S\arabic{figure}}%
\renewcommand{\thesection}{S\arabic{section}}%

\section{Transfer accuracies for other attacks}

\begin{figure}[H]
    \centering
    \begin{subfigure}[b]{\graphsubsize\textwidth}
        \raggedleft
        \includegraphics[width=\columnwidth]{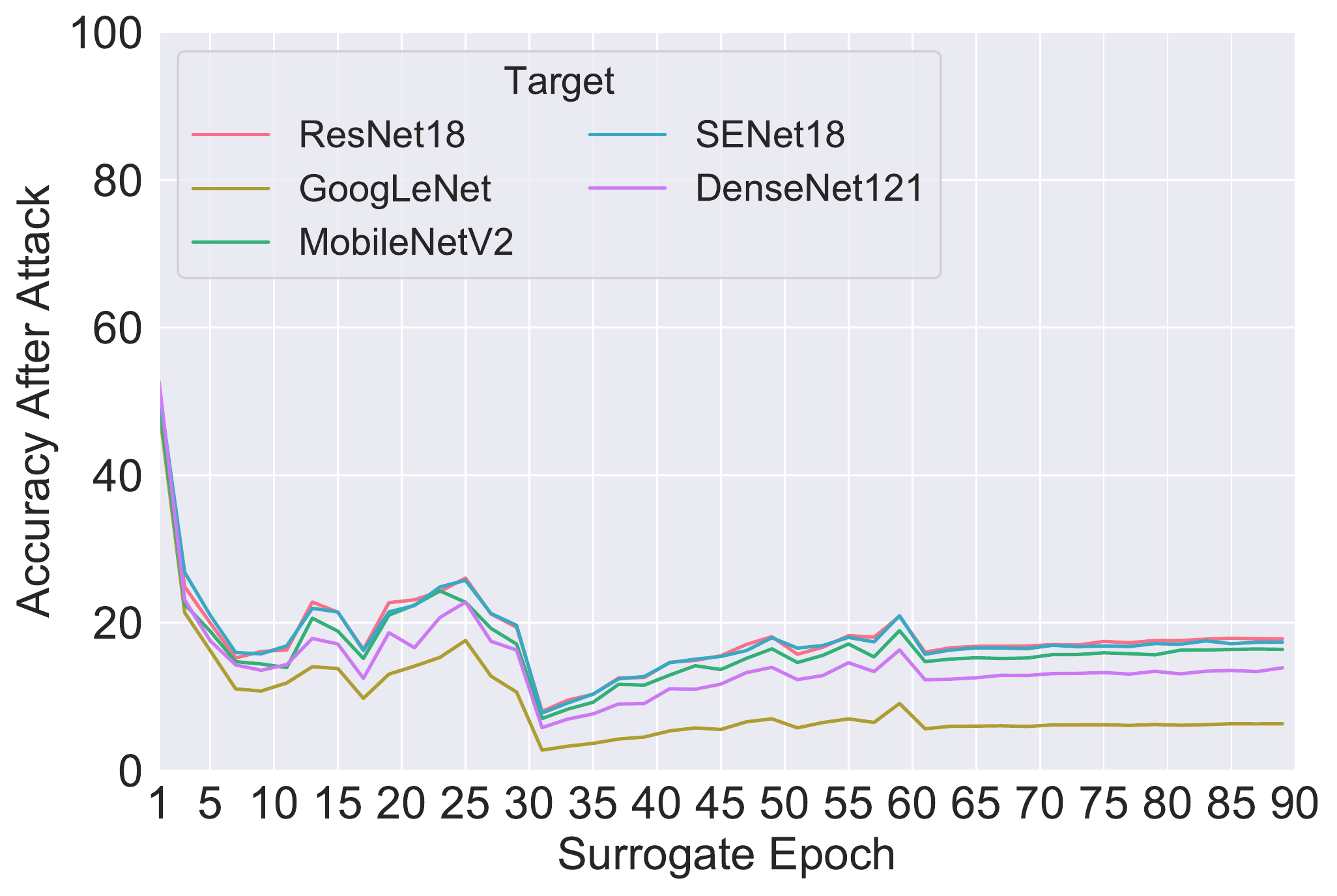}
        \subcaption{GoogLeNet surrogate model}
        \label{fig:googlenet_ilamifgsm_transfer}
    \end{subfigure}
    \hfill
    \begin{subfigure}[b]{\graphsubsize\textwidth}
        \raggedright
        \includegraphics[width=\columnwidth]{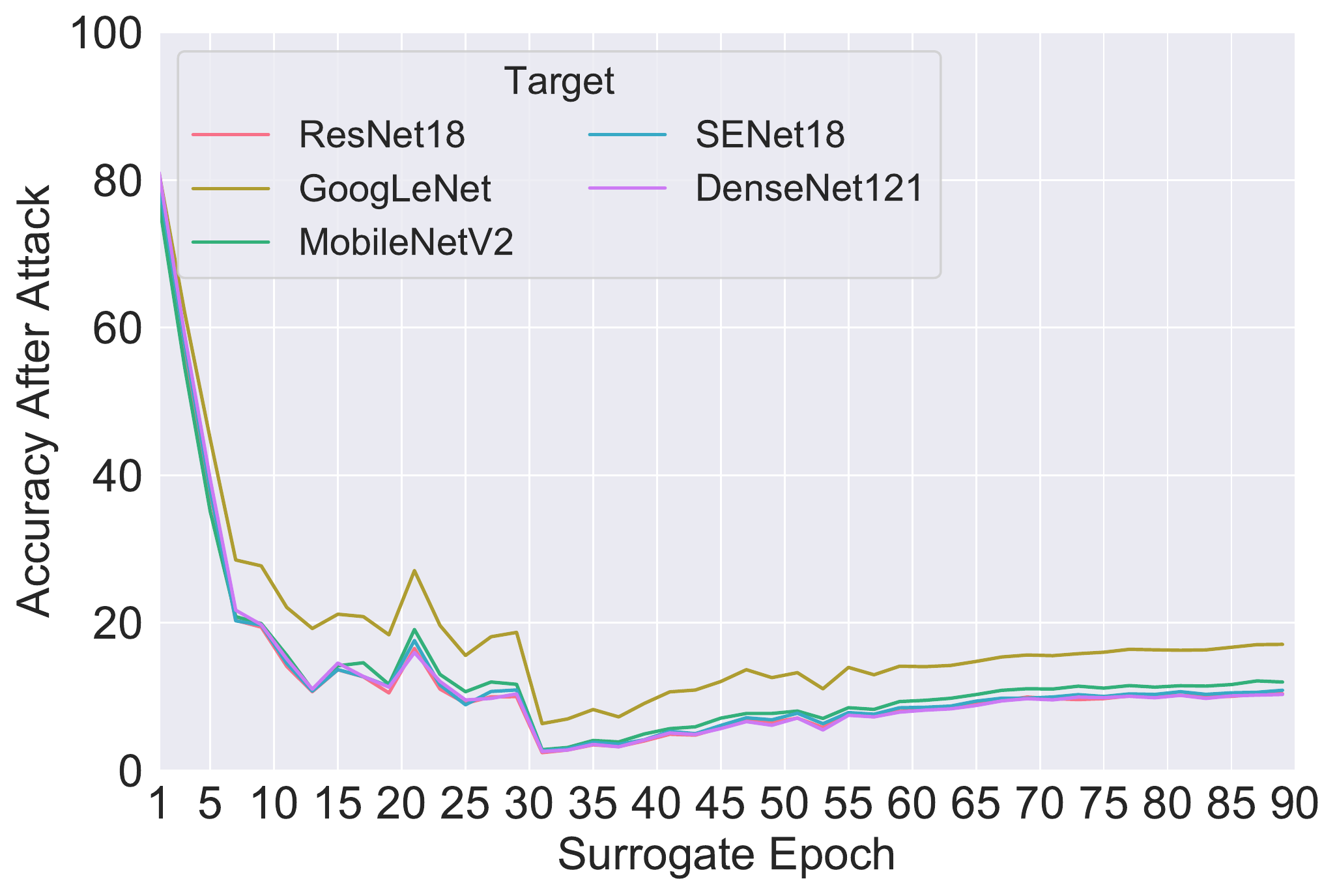}
        \subcaption{ResNet18 surrogate model}
        \label{fig:resnet_ilamifgsm_transfer}
    \end{subfigure}
    \newline 
    \begin{subfigure}[b]{\graphsubsize\textwidth}
        \raggedleft
        \includegraphics[width=\columnwidth]{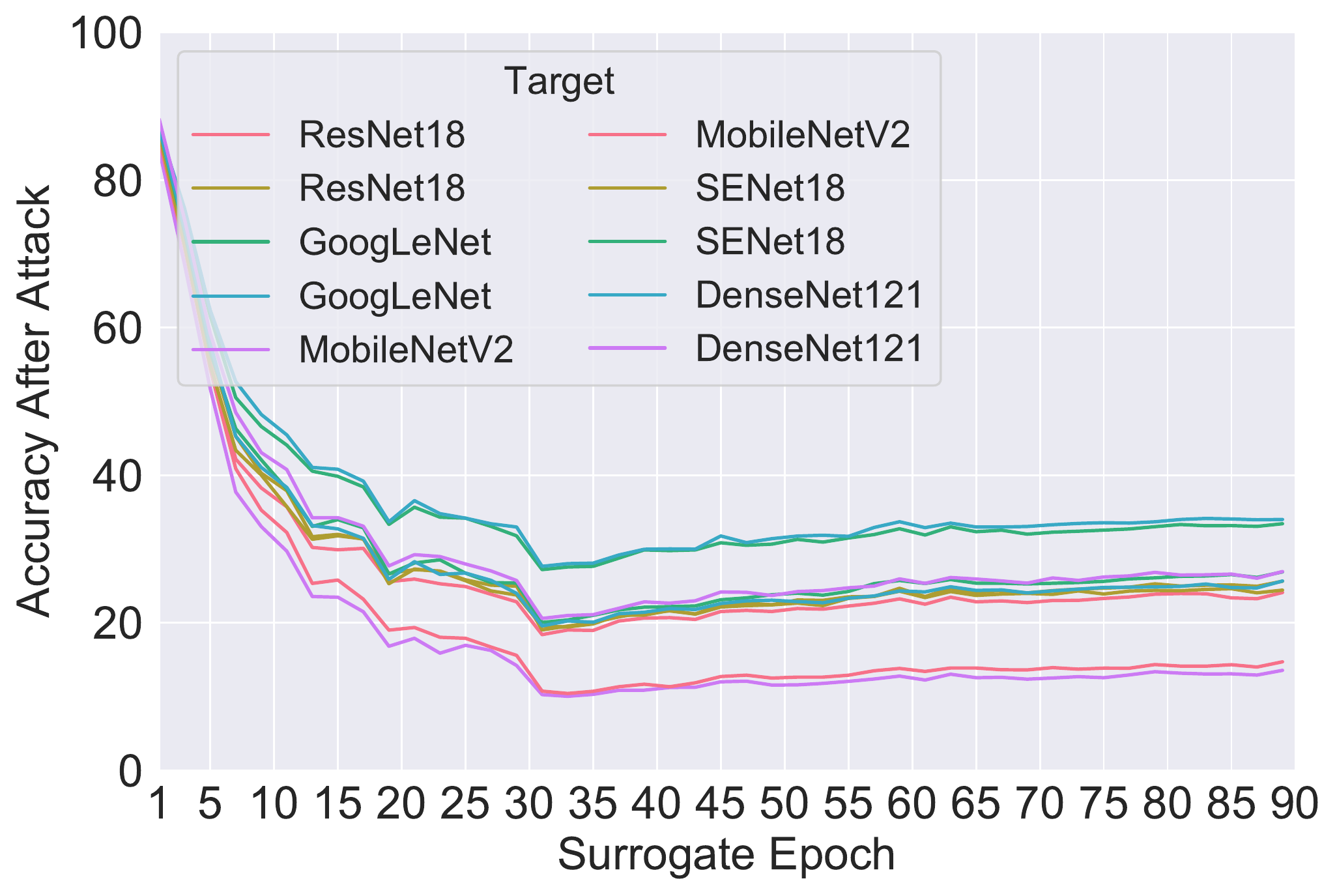}
        \subcaption{MobileNetV2 surrogate model}
        \label{fig:mobilenet_ilamifgsm_transfer}
    \end{subfigure}
    \hfill
    \begin{subfigure}[b]{\graphsubsize\textwidth}
        \raggedright
        \includegraphics[width=\columnwidth]{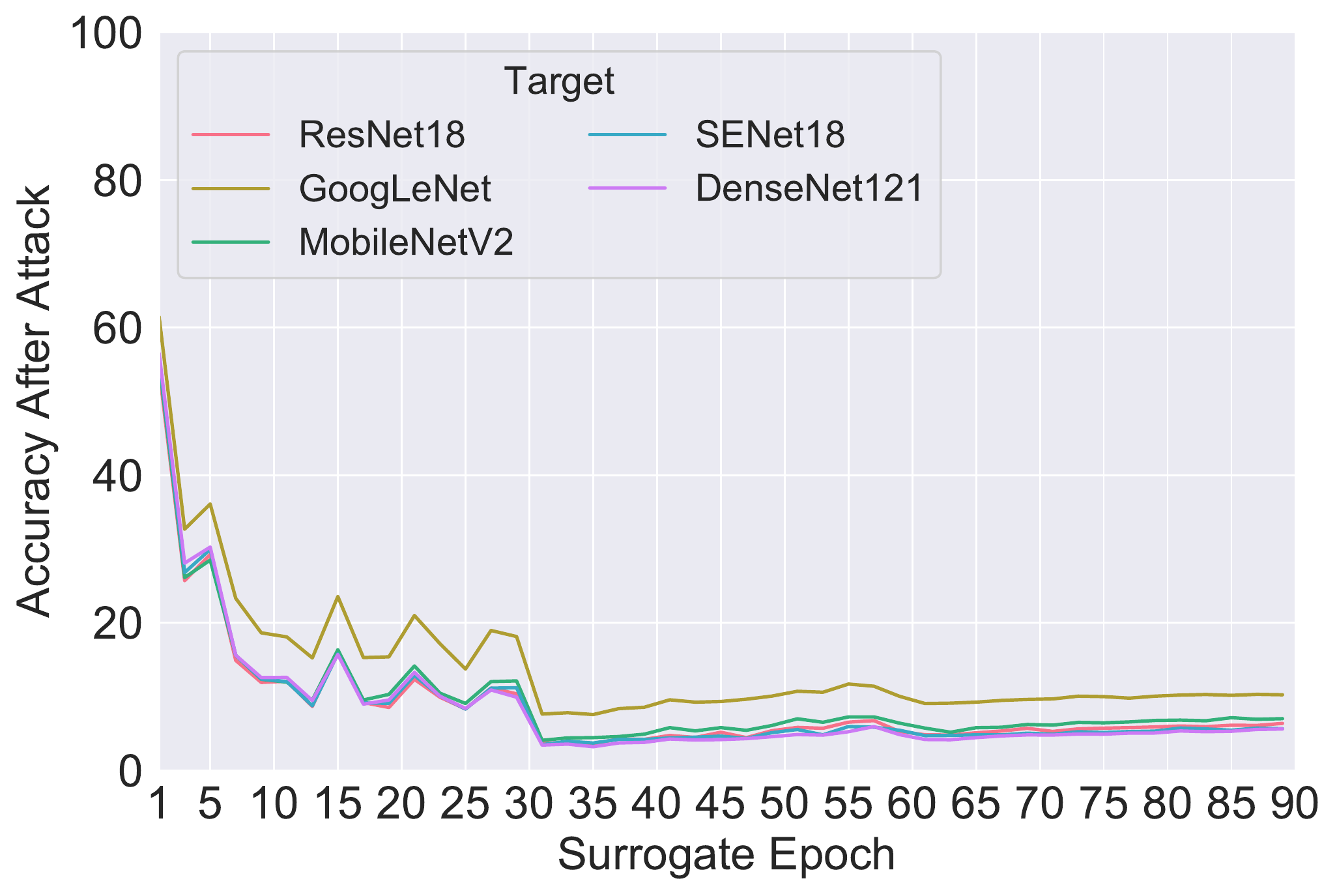}
        \subcaption{SENet18 surrogate model}
        \label{fig:senet_ilamifgsm_transfer}
    \end{subfigure}
    \newline 
    \begin{subfigure}[b]{\graphsubsize\textwidth}
        \centering
        \includegraphics[width=\columnwidth]{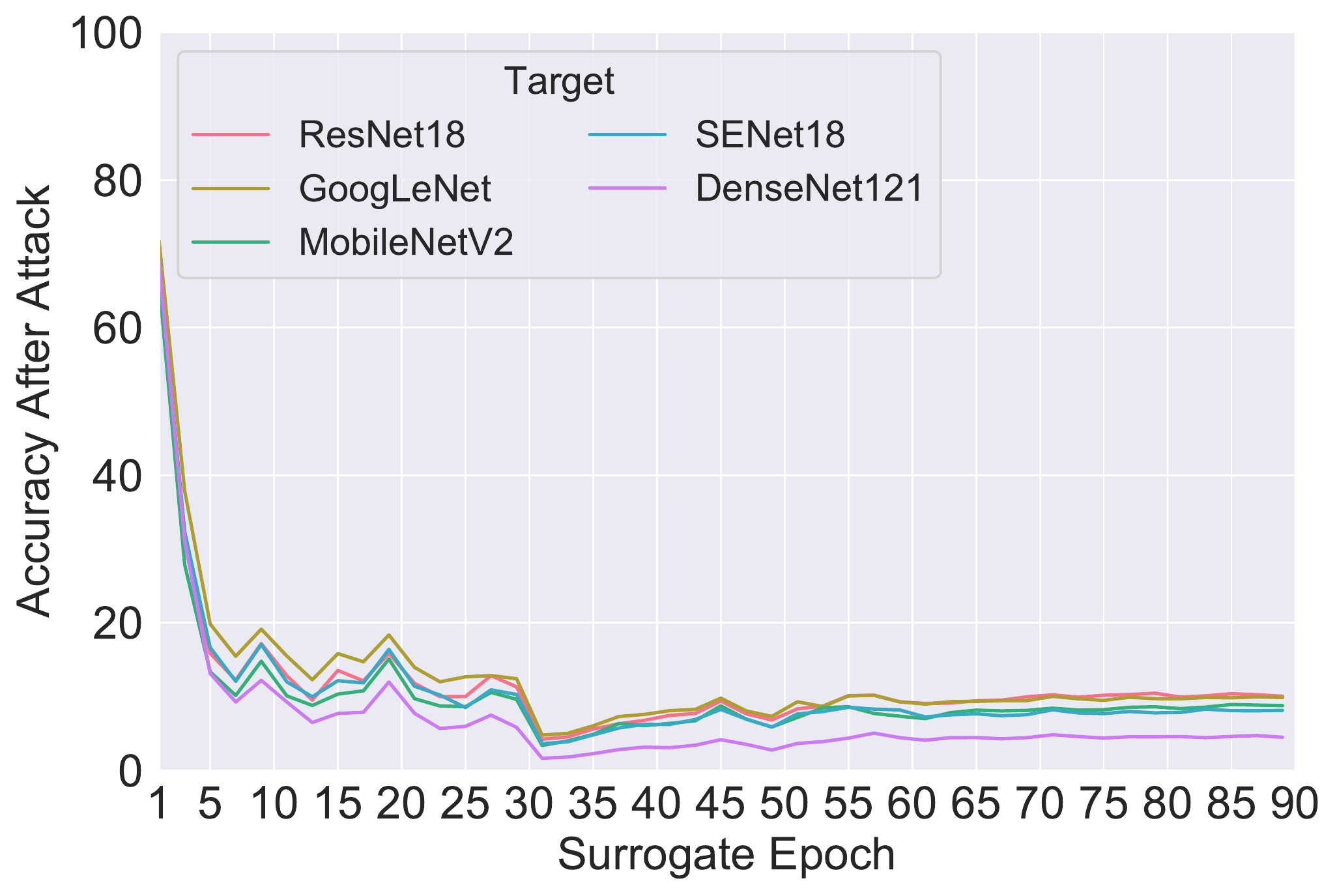}
        \subcaption{DenseNet121 surrogate model}
        \label{fig:densenet_ilamifgsm_transfer}
    \end{subfigure}
    
    \caption{Post-attack accuracy for ILA enhanced MI-FGSM transfer attacks from naturally trained models to target models, ($\epsilon=.05$). Lower accuracy indicates better transfer. The dashed lines indicate the surrogate epoch with the best transferability for most attacks.}
    \label{fig:ILA_mifgsm_transfer}
\end{figure}

\begin{figure}[H]
    \centering
    \begin{subfigure}[b]{\graphsubsize\textwidth}
        \raggedleft
        \includegraphics[width=\columnwidth]{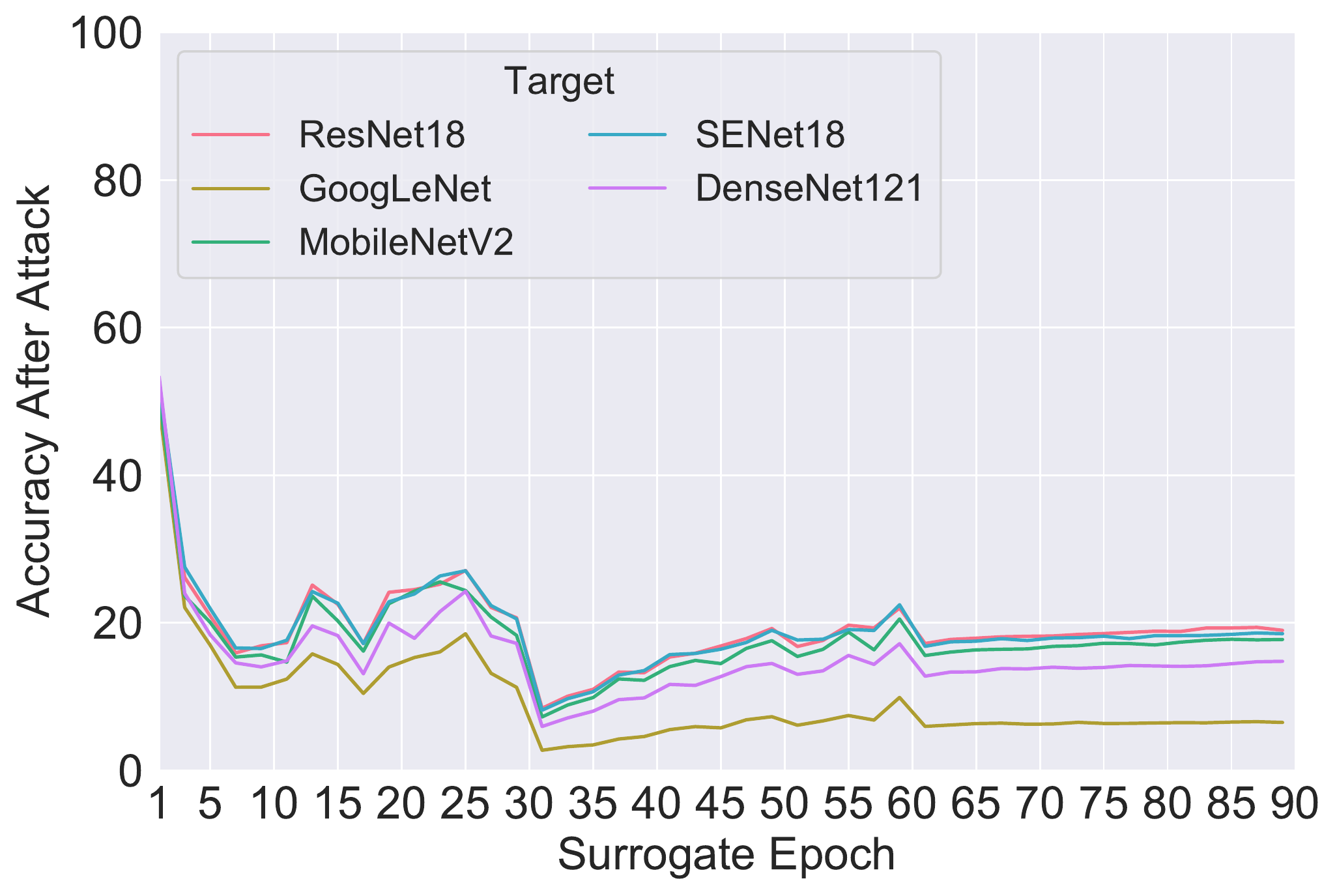}
        \subcaption{GoogLeNet surrogate model}
        \label{fig:googlenet_ilaifgsm_transfer}
    \end{subfigure}
    \hfill
    \begin{subfigure}[b]{\graphsubsize\textwidth}
        \raggedright
        \includegraphics[width=\columnwidth]{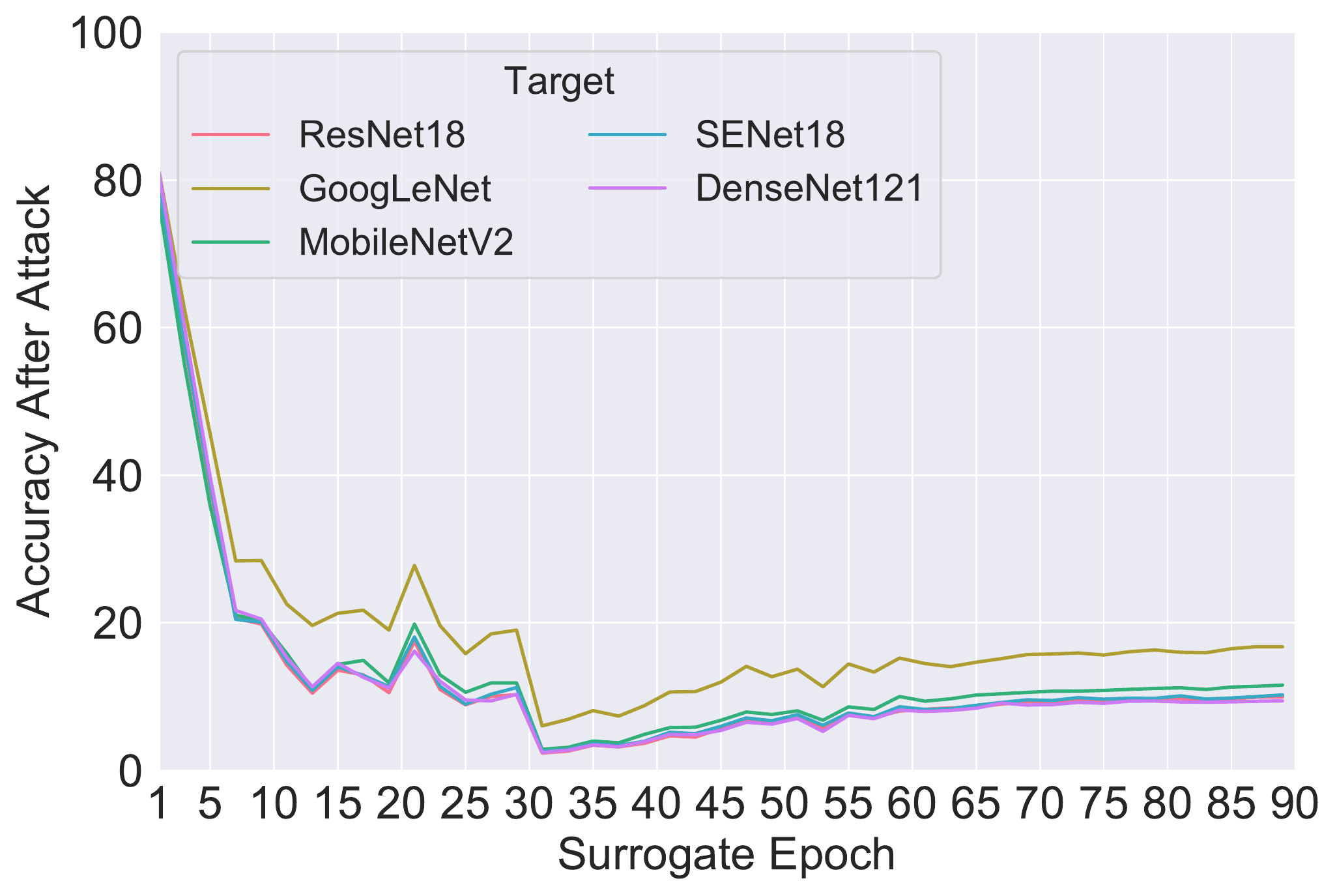}
        \subcaption{ResNet18 surrogate model}
        \label{fig:resnet_ilaifgsm_transfer}
    \end{subfigure}
    \newline 
    \begin{subfigure}[b]{\graphsubsize\textwidth}
        \raggedleft
        \includegraphics[width=\columnwidth]{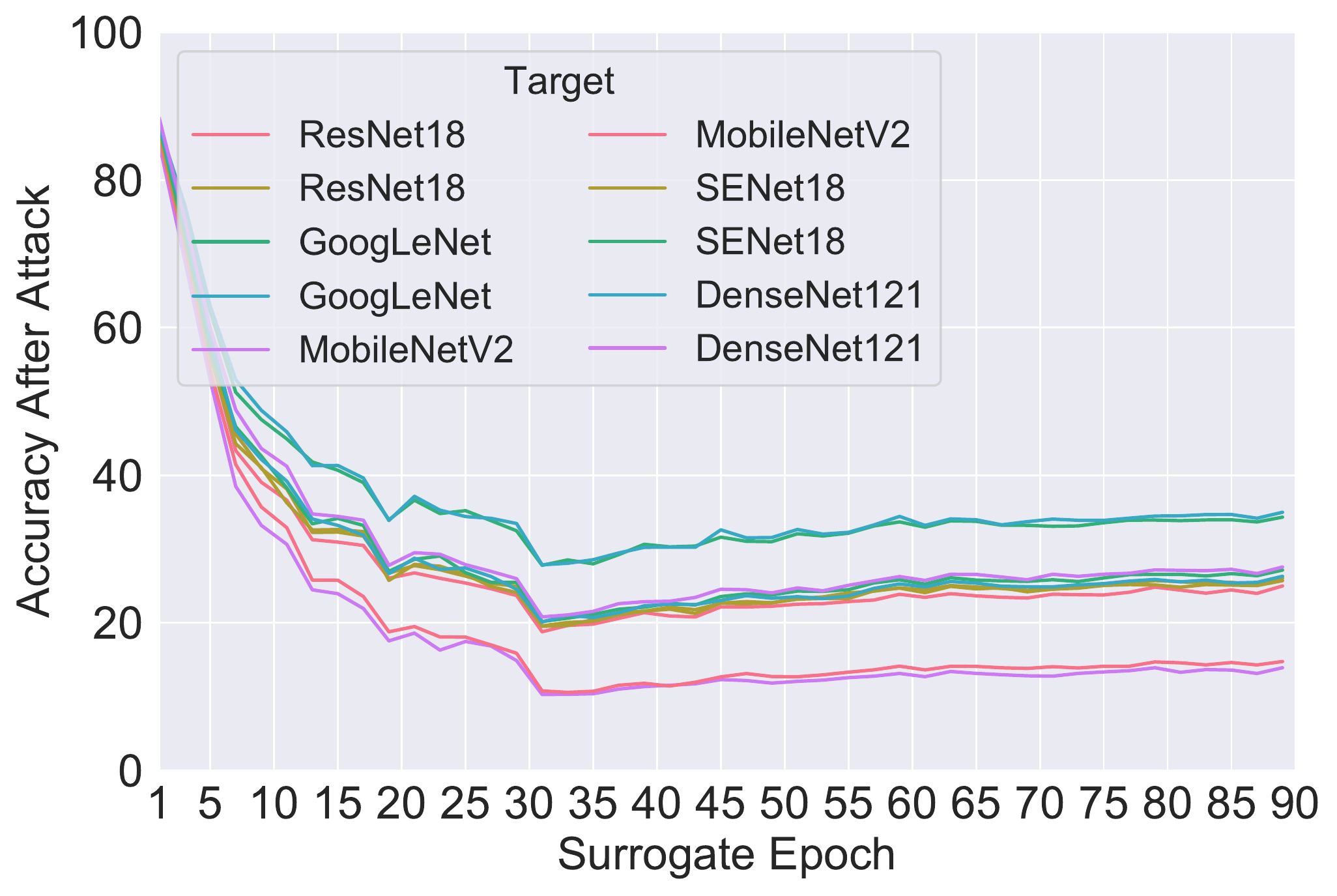}
        \subcaption{MobileNetV2 surrogate model}
        \label{fig:mobilenet_ilaifgsm_transfer}
    \end{subfigure}
    \hfill
    \begin{subfigure}[b]{\graphsubsize\textwidth}
        \raggedright
        \includegraphics[width=\columnwidth]{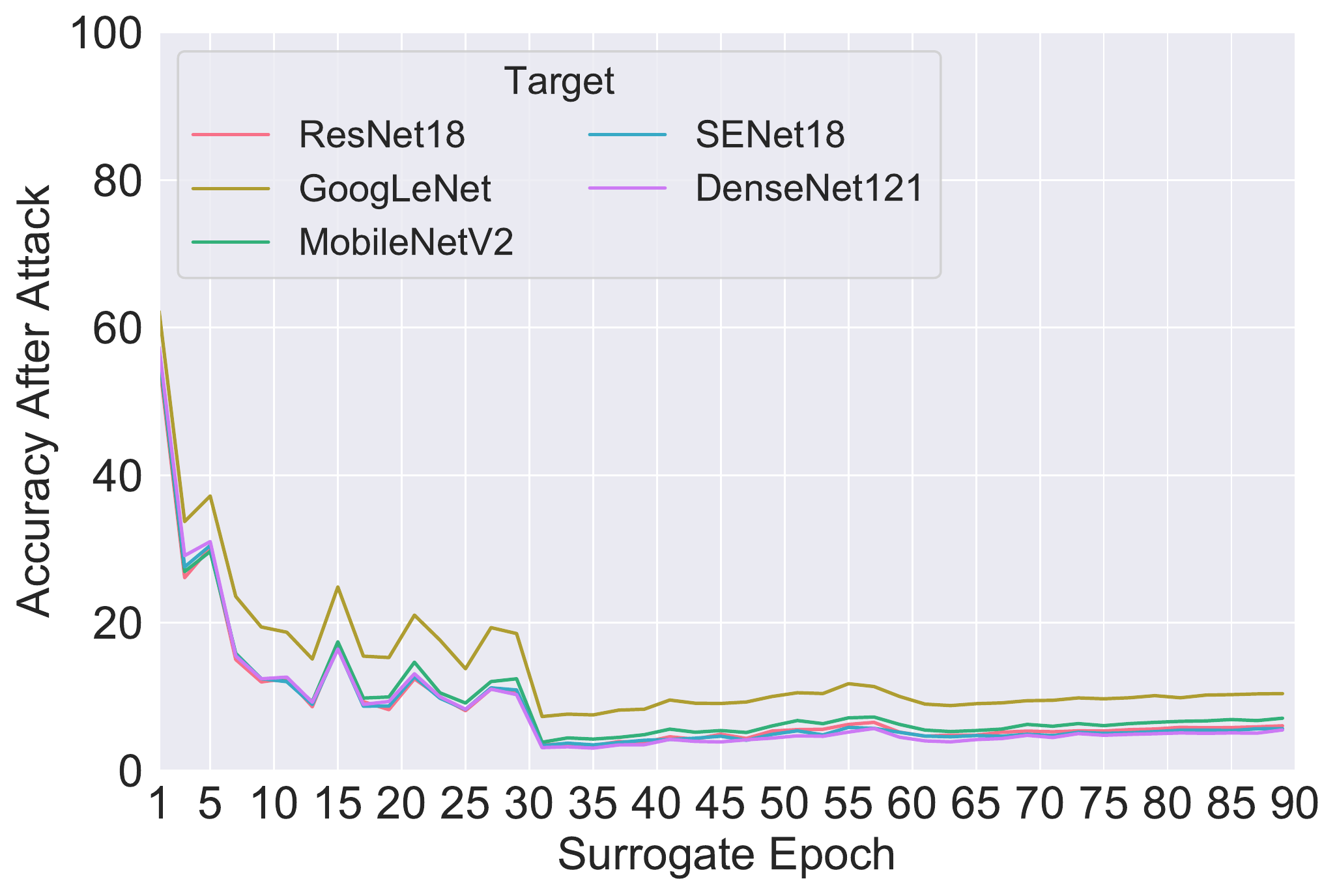}
        \subcaption{SENet18 surrogate model}
        \label{fig:senet_ilaifgsm_transfer}
    \end{subfigure}
    \newline 
    \begin{subfigure}[b]{\graphsubsize\textwidth}
        \centering
        \includegraphics[width=\columnwidth]{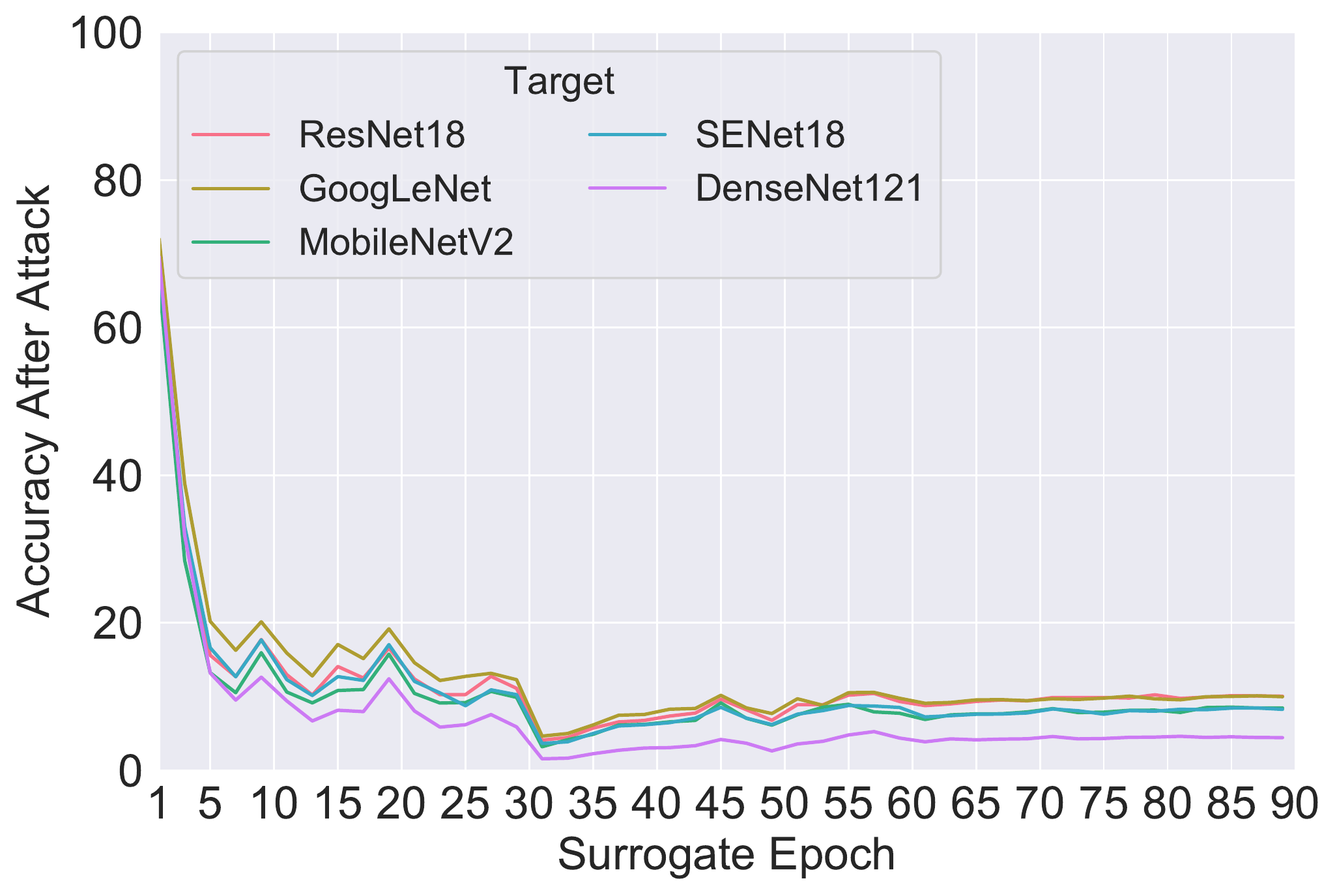}
        \subcaption{DenseNet121 surrogate model}
        \label{fig:densenet_ilaifgsm_transfer}
    \end{subfigure}
    
    \caption{Post-attack accuracy for ILA enhanced I-FGSM transfer attacks from naturally trained models to target models, ($\epsilon=.05$). Lower accuracy indicates better transfer. The dashed lines indicate the surrogate epoch with the best transferability for most attacks.}
    \label{fig:ILA_ifgsm_transfer}
\end{figure}

\begin{figure}[H]
    \centering
    \begin{subfigure}[b]{\graphsubsize\textwidth}
        \raggedleft
        \includegraphics[width=\columnwidth]{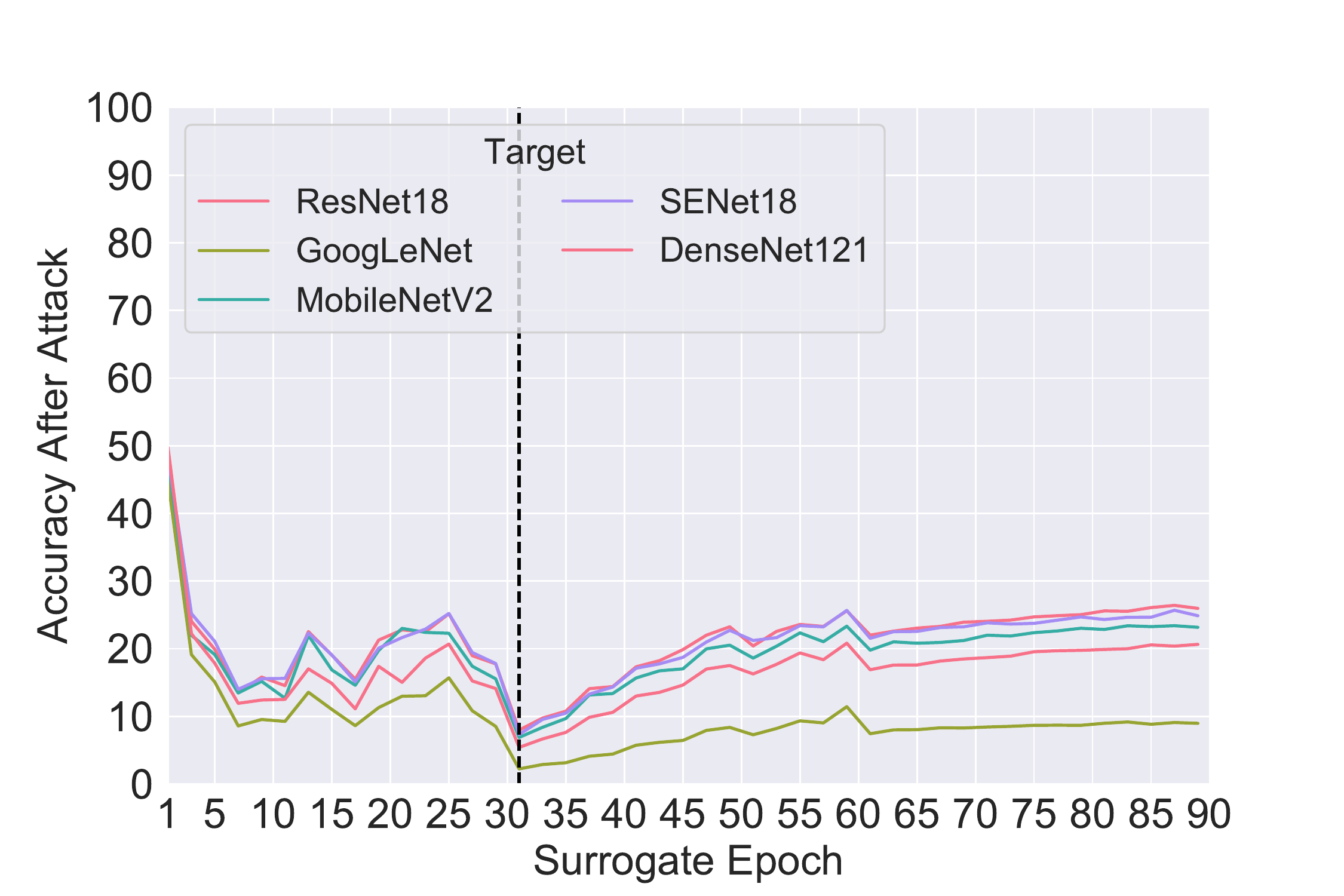}
        \subcaption{GoogLeNet surrogate model}
        \label{fig:googlenet_ifgsm_transfer}
    \end{subfigure}
    \hfill
    \begin{subfigure}[b]{\graphsubsize\textwidth}
        \raggedright
        \includegraphics[width=\columnwidth]{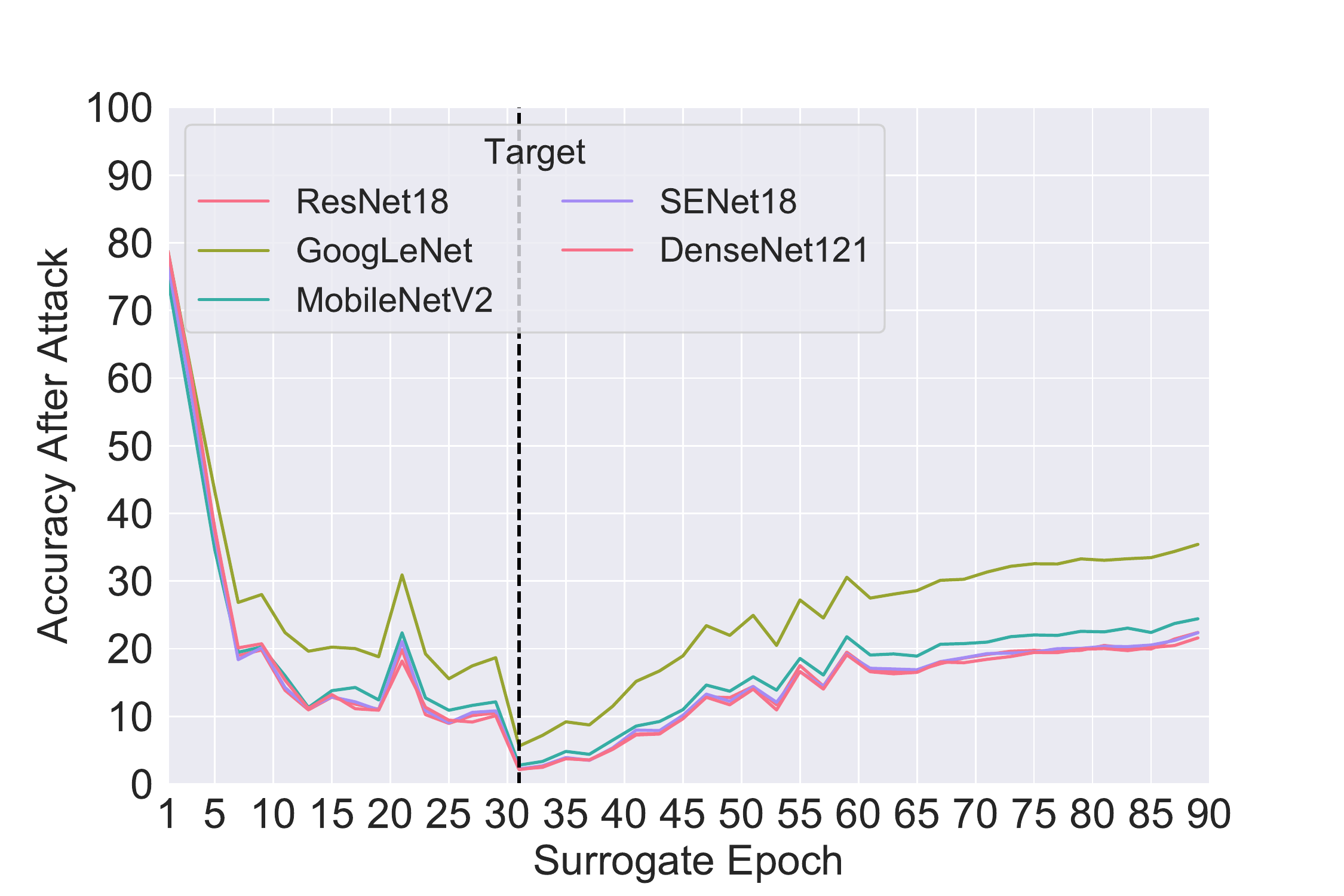}
        \subcaption{ResNet18 surrogate model}
        \label{fig:resnet_ifgsm_transfer}
    \end{subfigure}
    \newline 
    \begin{subfigure}[b]{\graphsubsize\textwidth}
        \raggedleft
        \includegraphics[width=\columnwidth]{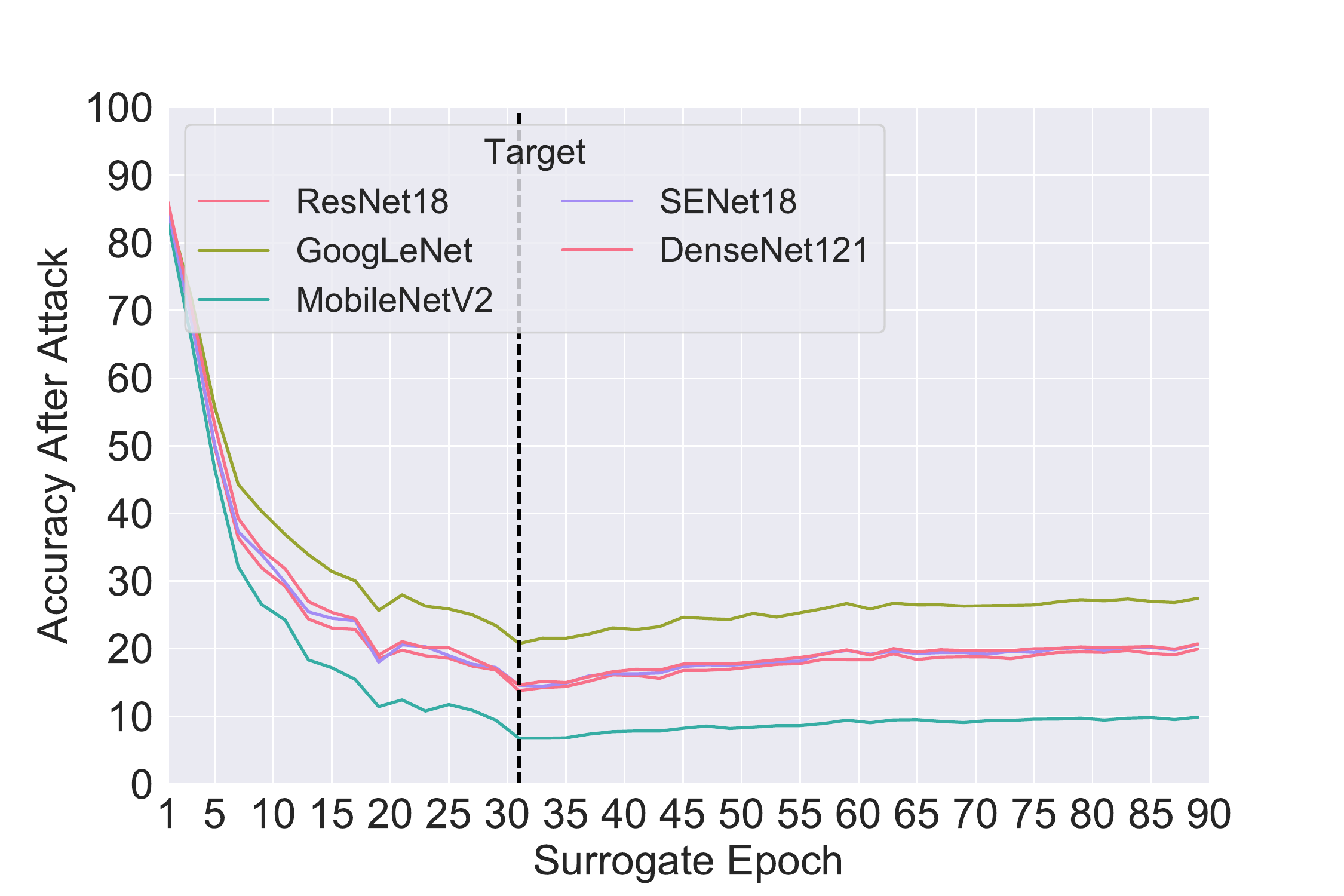}
        \subcaption{MobileNetV2 surrogate model}
        \label{fig:mobilenet_ifgsm_transfer}
    \end{subfigure}
    \hfill
    \begin{subfigure}[b]{\graphsubsize\textwidth}
        \raggedright
        \includegraphics[width=\columnwidth]{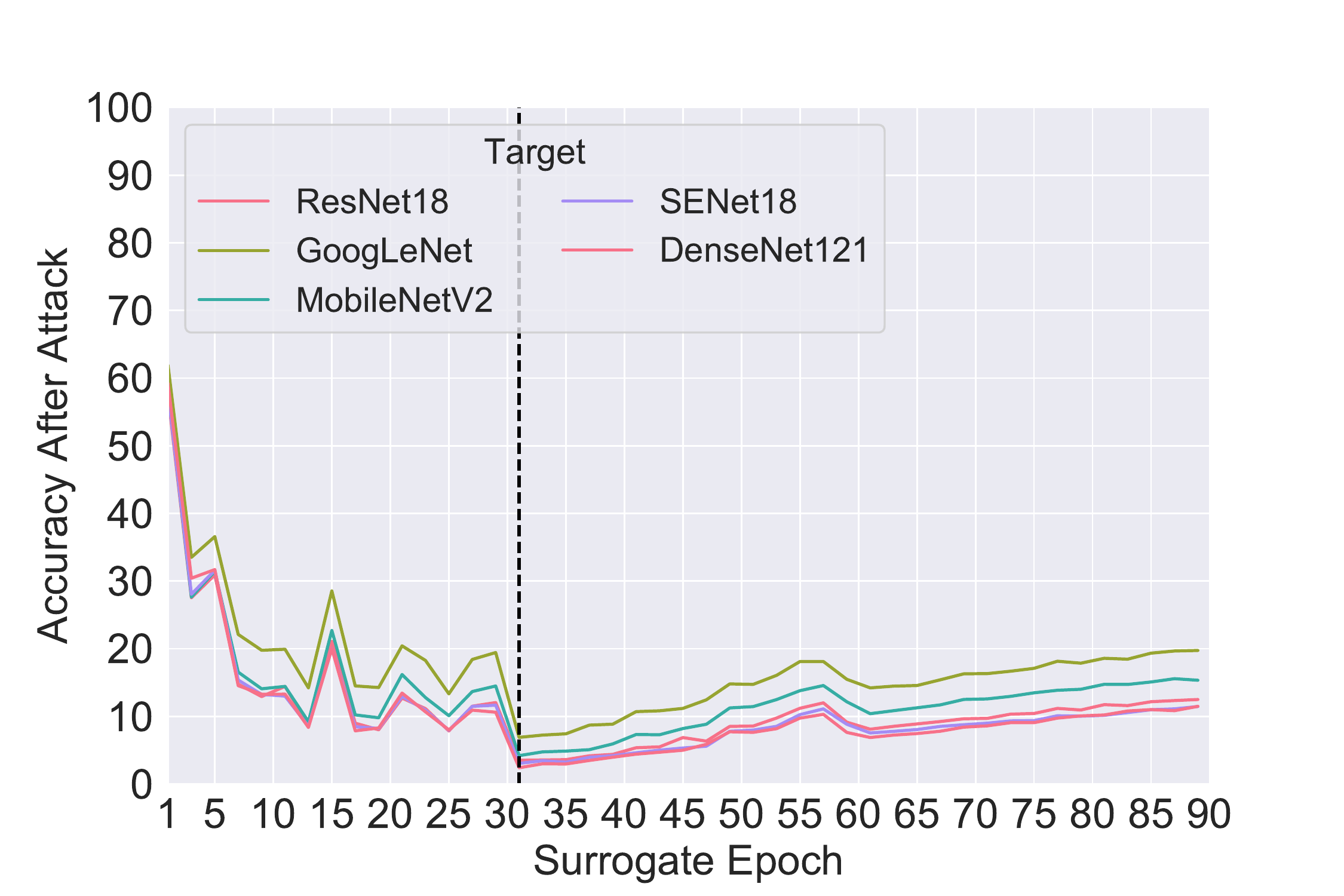}
        \subcaption{SENet18 surrogate model}
        \label{fig:senet_ifgsm_transfer}
    \end{subfigure}
    \newline 
    \begin{subfigure}[b]{\graphsubsize\textwidth}
        \centering
        \includegraphics[width=\columnwidth]{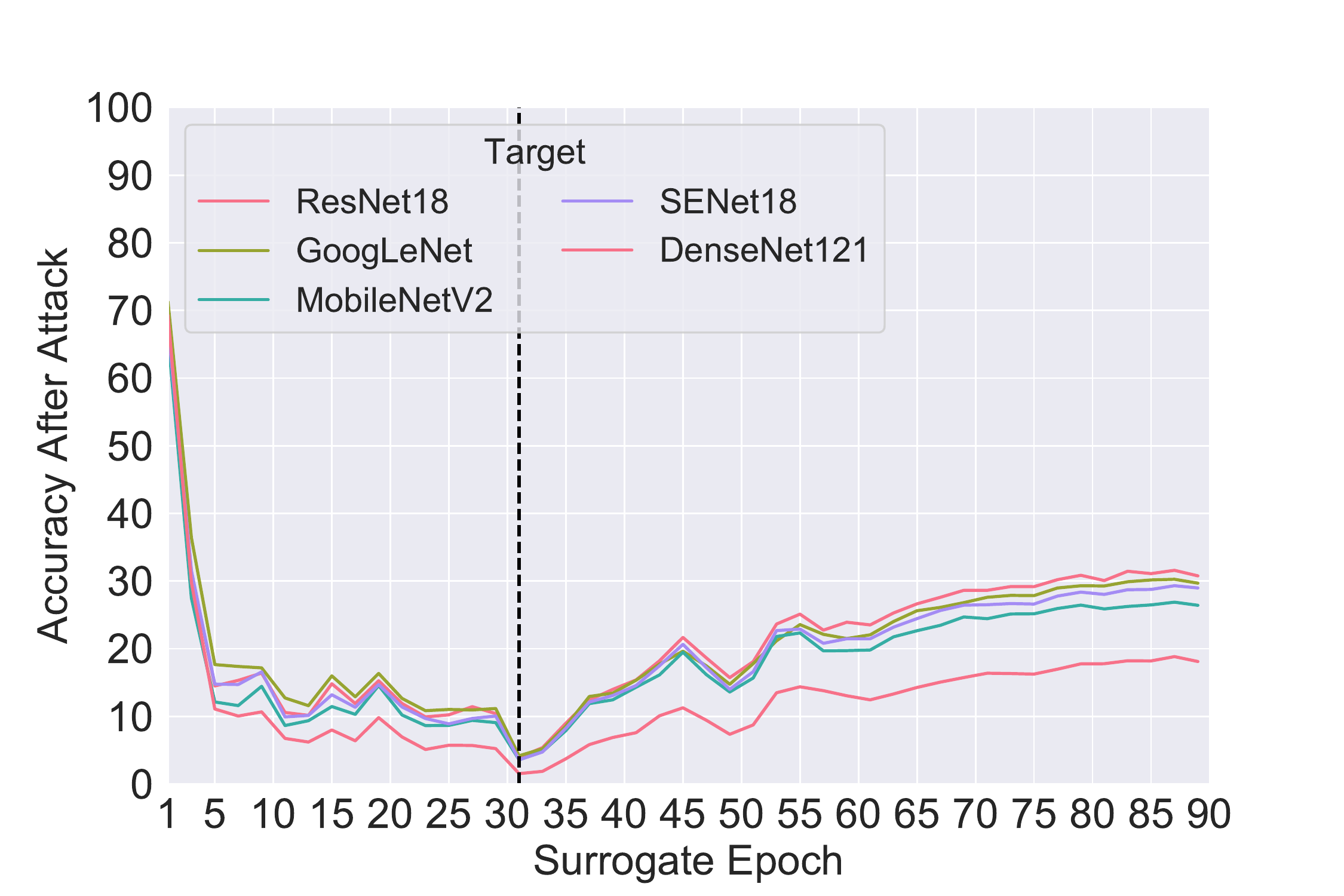}
        \subcaption{DenseNet121 surrogate model}
        \label{fig:densenet_ifgsm_transfer}
    \end{subfigure}
    
    \caption{Post-attack accuracy for I-FGSM transfer attacks from naturally trained models to target models, ($\epsilon=.05$). Lower accuracy indicates better transfer. The dashed lines indicate the surrogate epoch with the best transferability for most attacks.}
    \label{fig:ifgsm_transfer}
\end{figure}

\begin{figure}[H]
    \centering
    \begin{subfigure}[b]{\graphsubsize\textwidth}
        \raggedleft
        \includegraphics[width=\columnwidth]{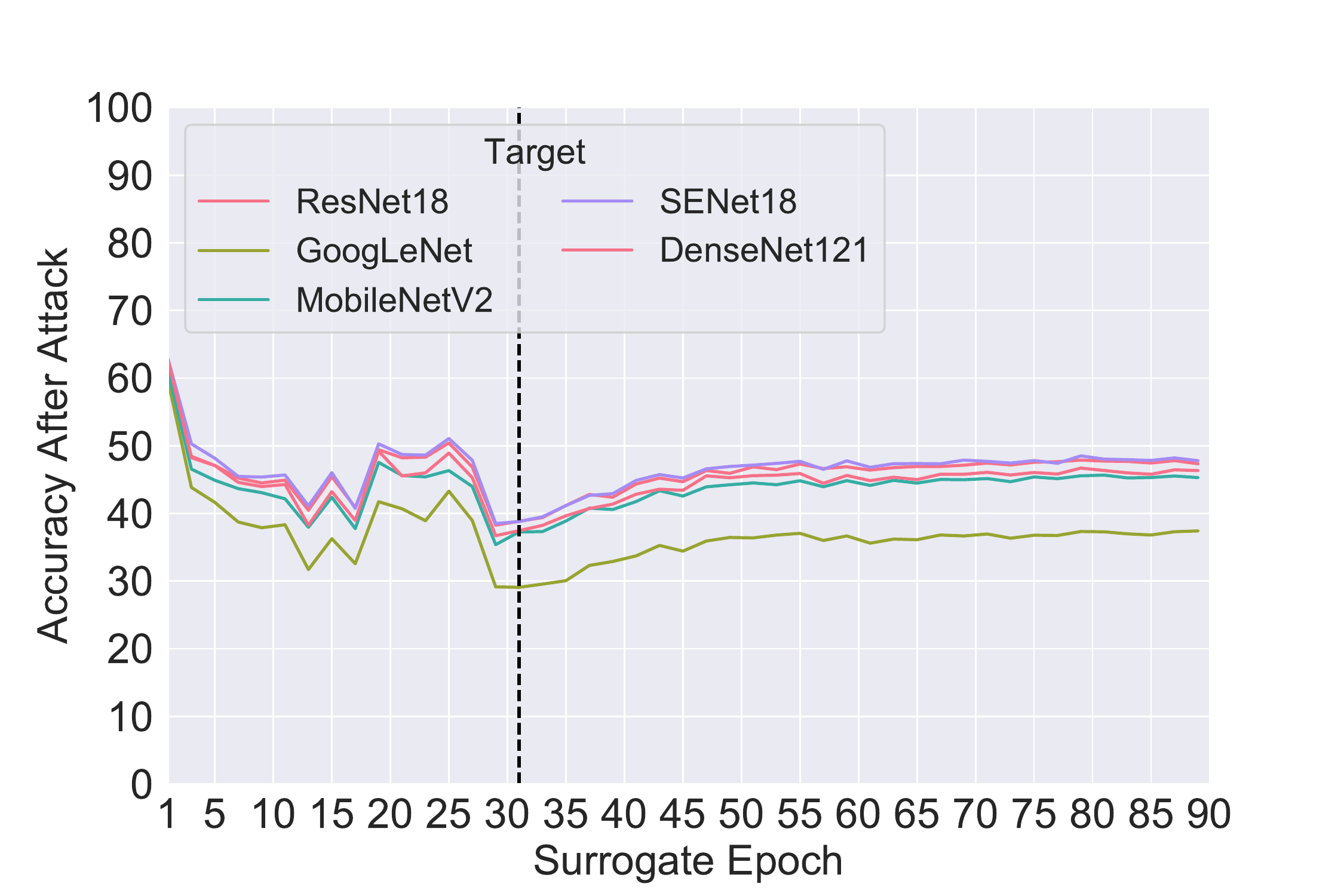}
        \subcaption{GoogLeNet surrogate model}
        \label{fig:googlenet_fgsm_transfer}
    \end{subfigure}
    \hfill
    \begin{subfigure}[b]{\graphsubsize\textwidth}
        \raggedright
        \includegraphics[width=\columnwidth]{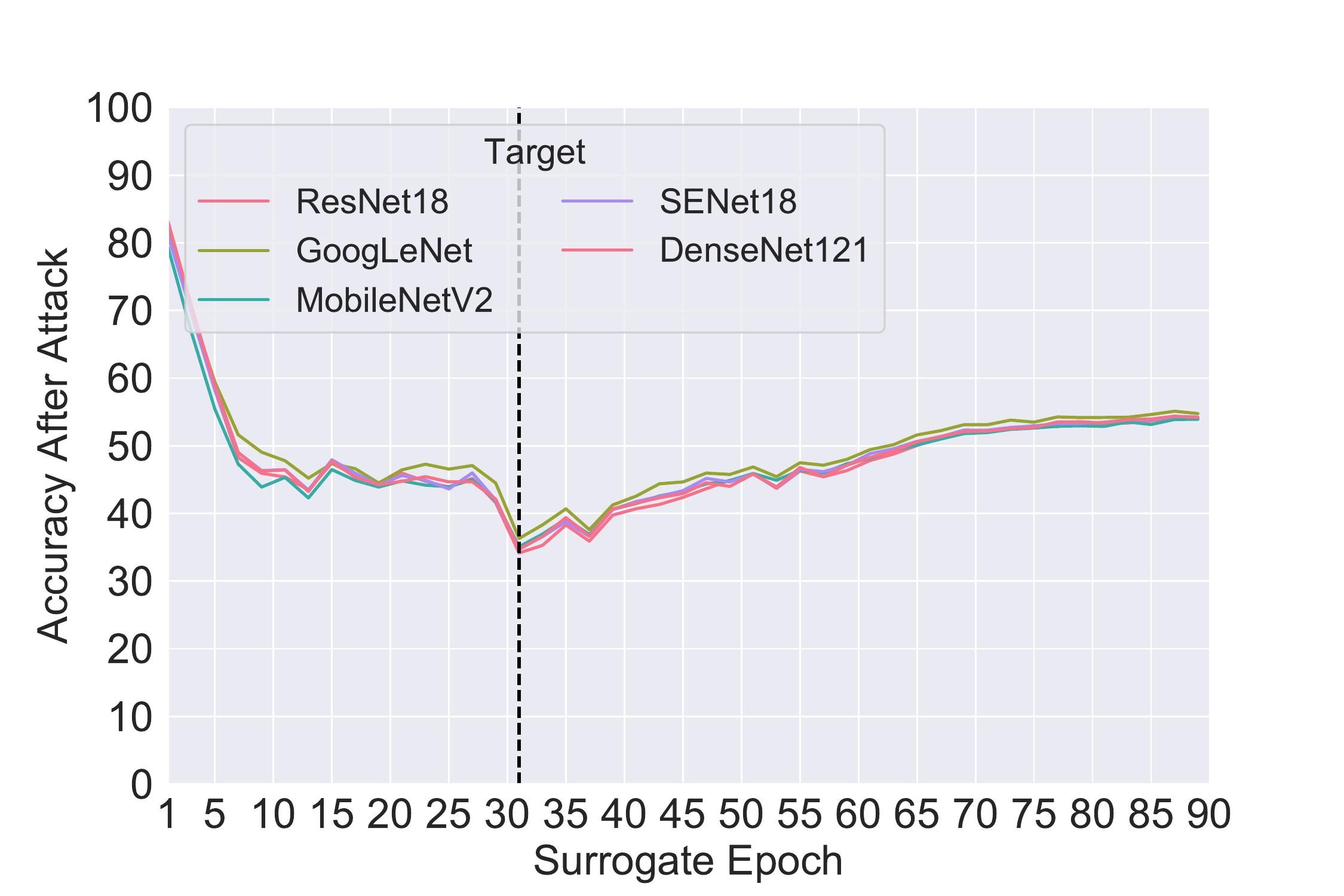}
        \subcaption{ResNet18 surrogate model}
        \label{fig:resnet_fgsm_transfer}
    \end{subfigure}
    \newline 
    \begin{subfigure}[b]{\graphsubsize\textwidth}
        \raggedleft
        \includegraphics[width=\columnwidth]{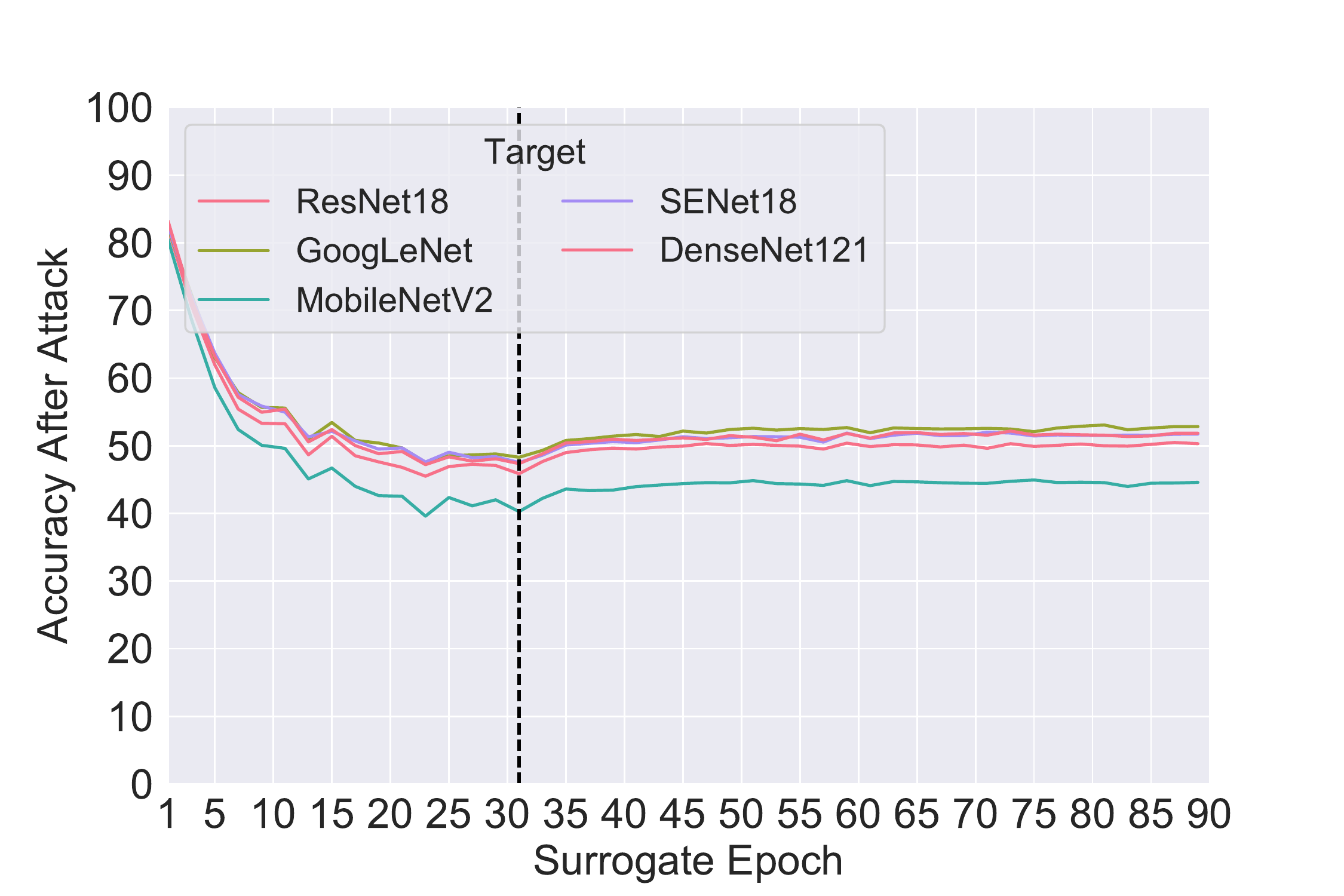}
        \subcaption{MobileNetV2 surrogate model}
        \label{fig:mobilenet_fgsm_transfer}
    \end{subfigure}
    \hfill
    \begin{subfigure}[b]{\graphsubsize\textwidth}
        \raggedright
        \includegraphics[width=\columnwidth]{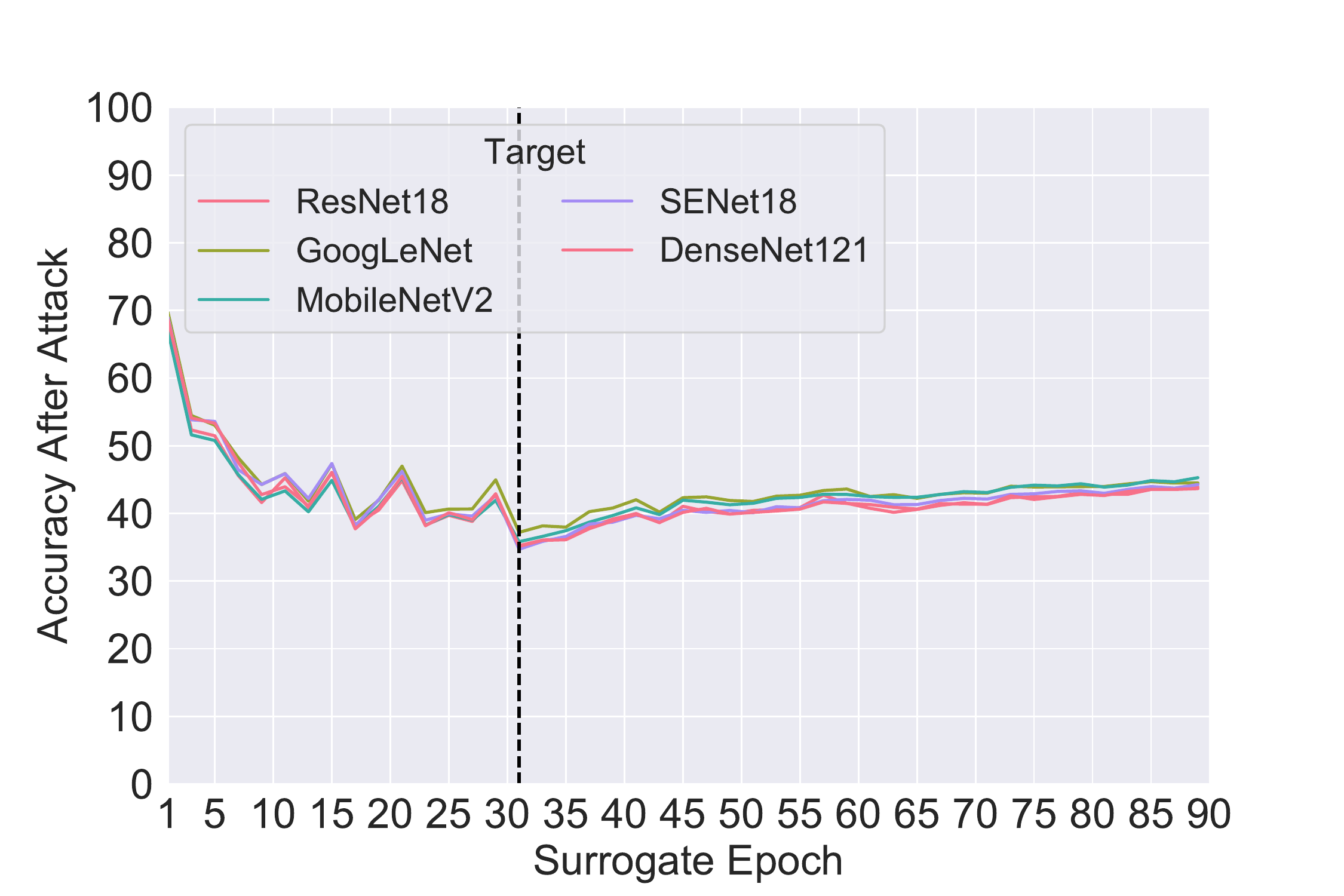}
        \subcaption{SENet18 surrogate model}
        \label{fig:senet_fgsm_transfer}
    \end{subfigure}
        \newline 
    \begin{subfigure}[b]{\graphsubsize\textwidth}
        \centering
        \includegraphics[width=\columnwidth]{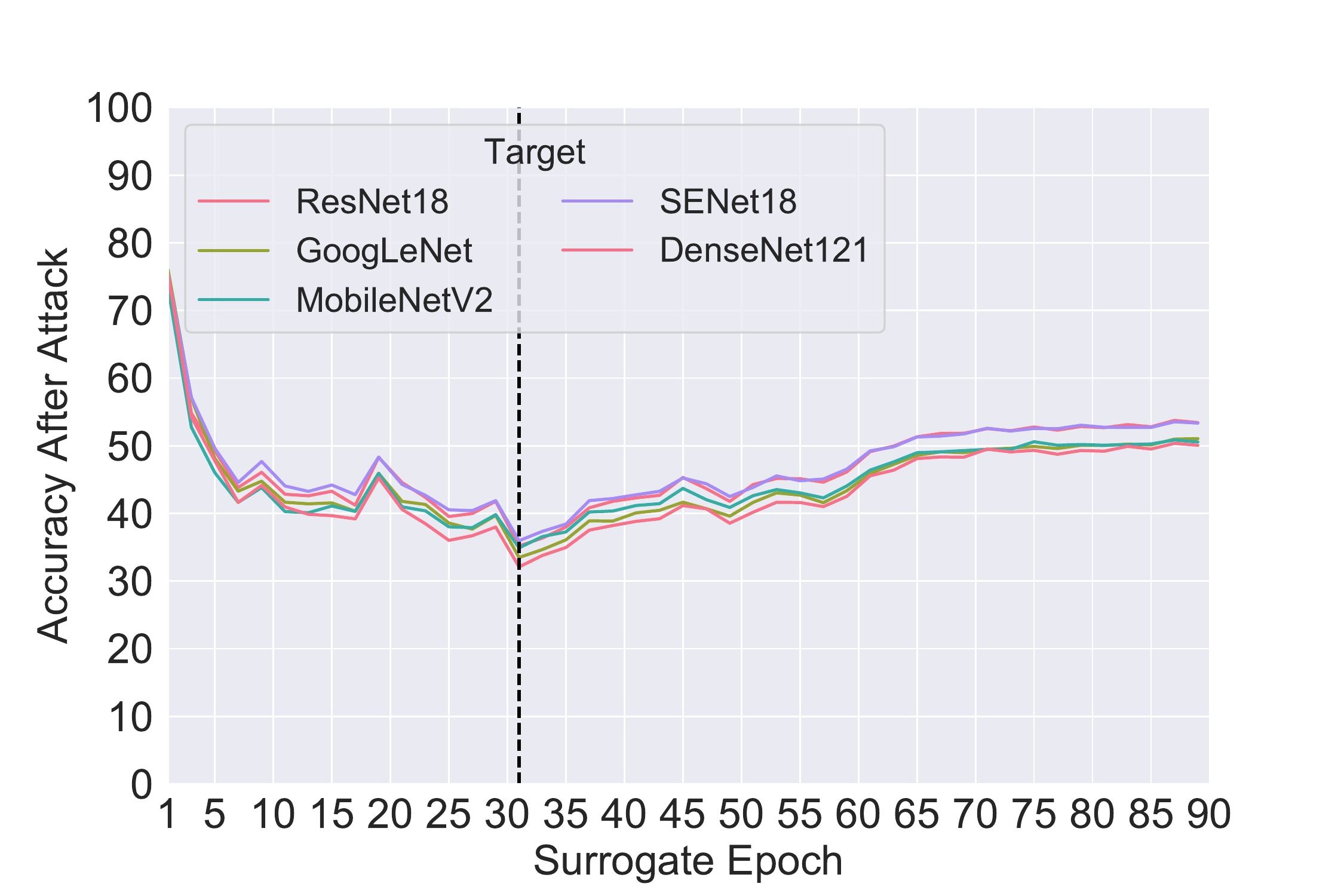}
        \subcaption{DenseNet121 surrogate model}
        \label{fig:densenet_fgsm_transfer}
    \end{subfigure}
    
    \caption{Post-attack accuracy for FGSM transfer attacks from naturally trained models to target models, ($\epsilon=.05$). Lower accuracy indicates better transfer. The dashed lines indicate the surrogate epoch with the best transferability for most attacks.}
    \label{fig:fgsm_transfer}
\end{figure}

\begin{figure}[H]
    \centering
    \begin{subfigure}[b]{\graphsubsize\textwidth}
        \raggedleft
        \includegraphics[width=\columnwidth]{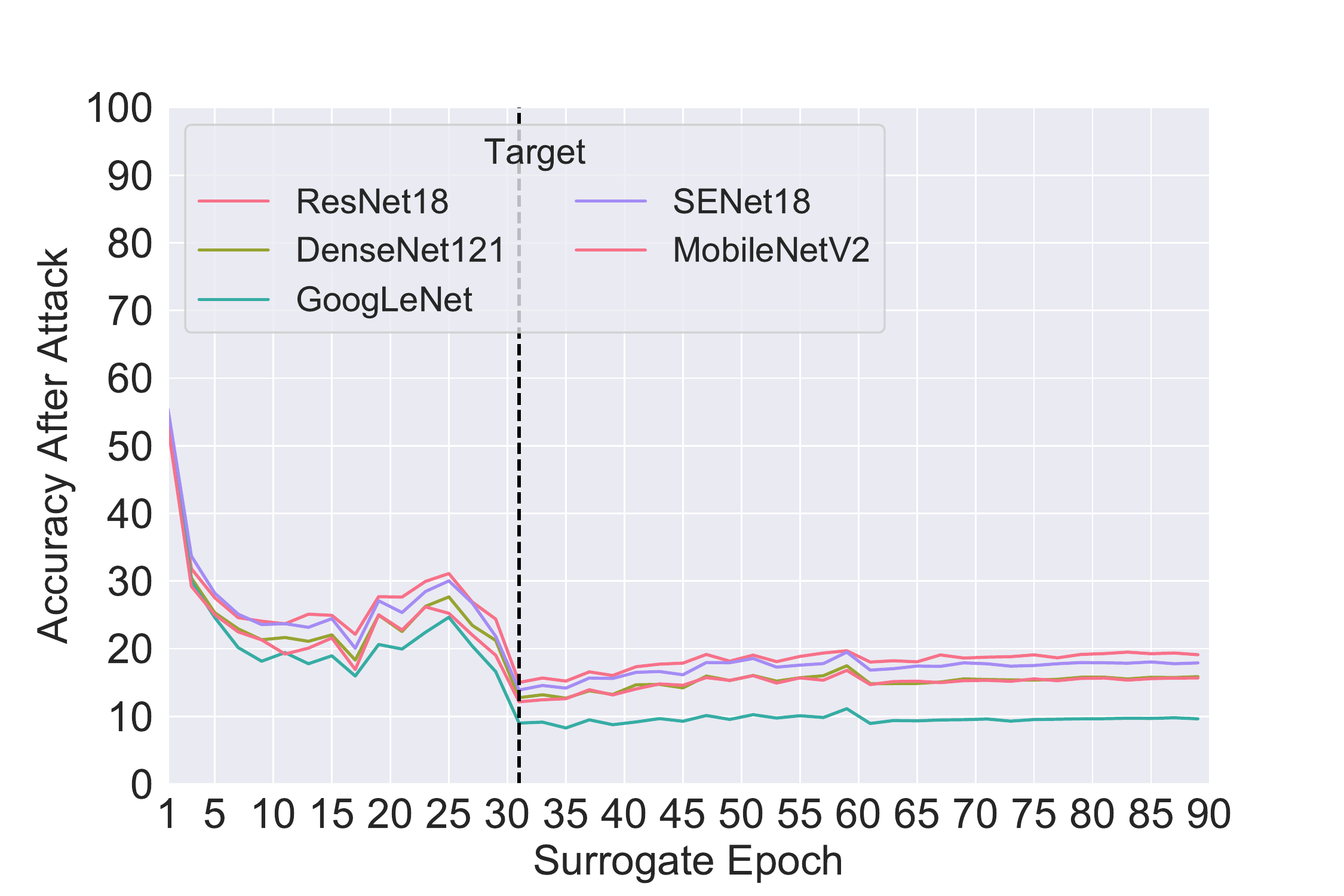}
        \subcaption{GoogLeNet surrogate model}
        \label{fig:googlenet_tap_transfer}
    \end{subfigure}
    \hfill
    \begin{subfigure}[b]{\graphsubsize\textwidth}
        \raggedright
        \includegraphics[width=\columnwidth]{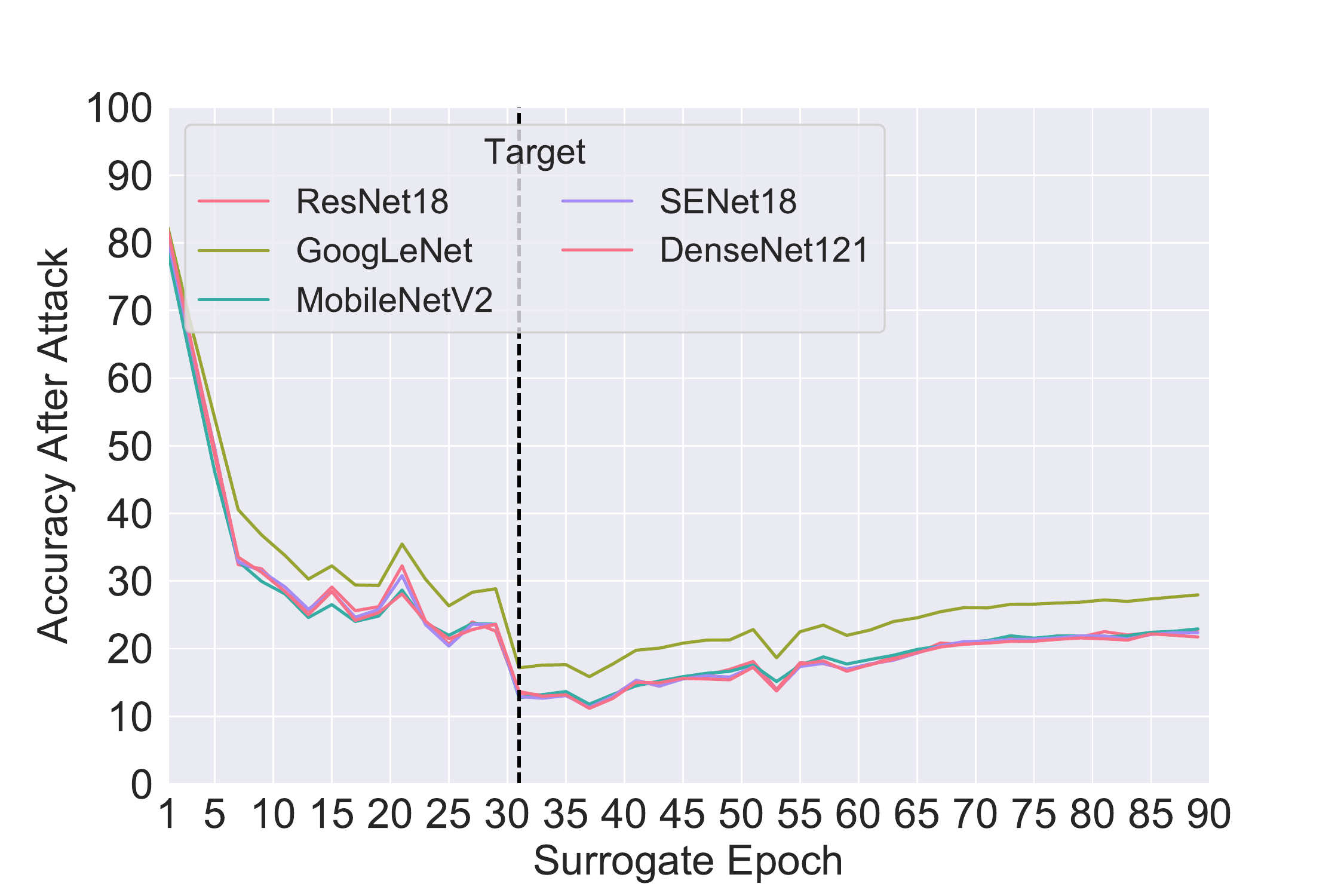}
        \subcaption{ResNet18 surrogate model}
        \label{fig:resnet_tap_transfer}
    \end{subfigure}
    \newline 
    \begin{subfigure}[b]{\graphsubsize\textwidth}
        \raggedleft
        \includegraphics[width=\columnwidth]{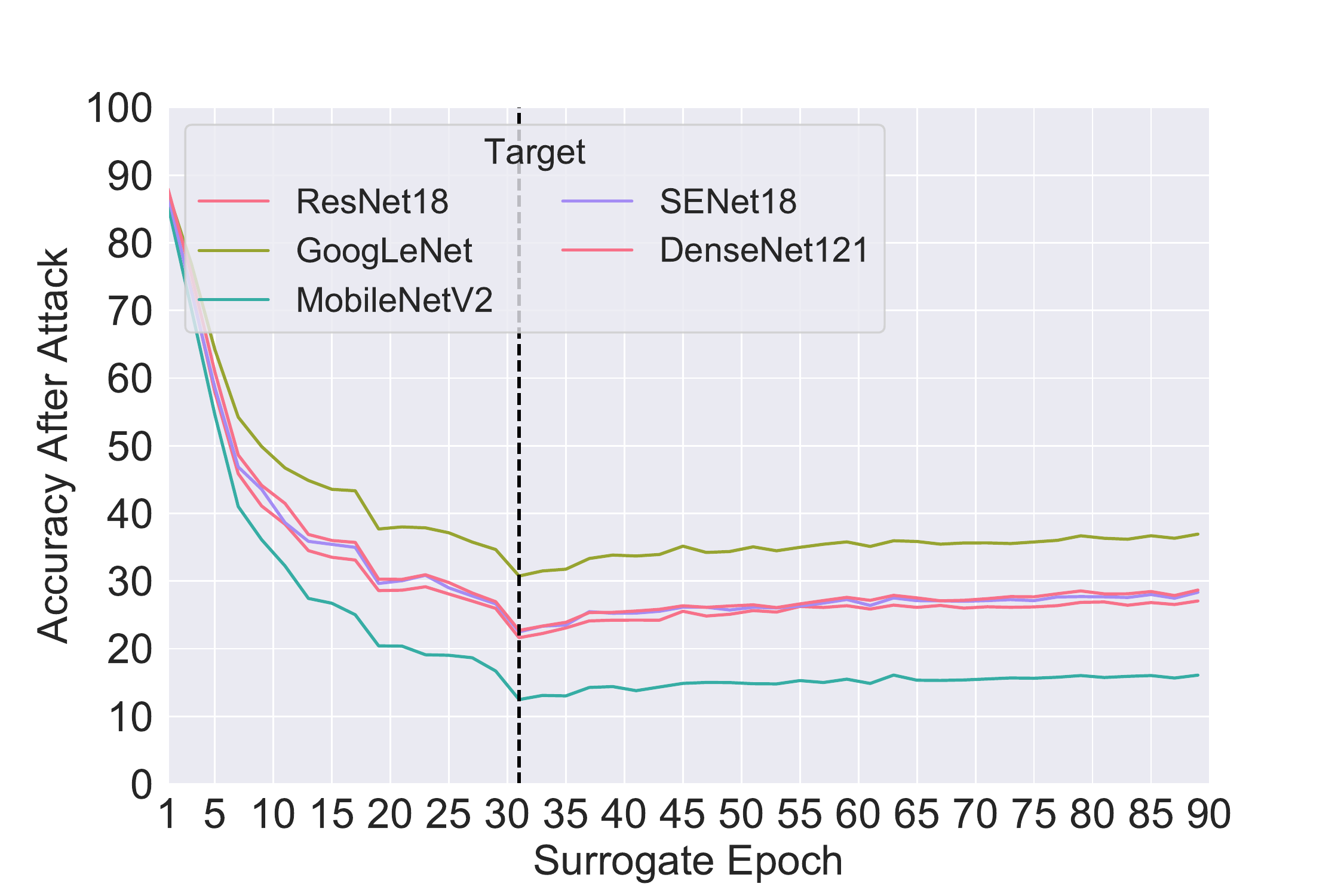}
        \subcaption{MobileNetV2 surrogate model}
        \label{fig:mobilenet_tap_transfer}
    \end{subfigure}
    \hfill
    \begin{subfigure}[b]{\graphsubsize\textwidth}
        \raggedright
        \includegraphics[width=\columnwidth]{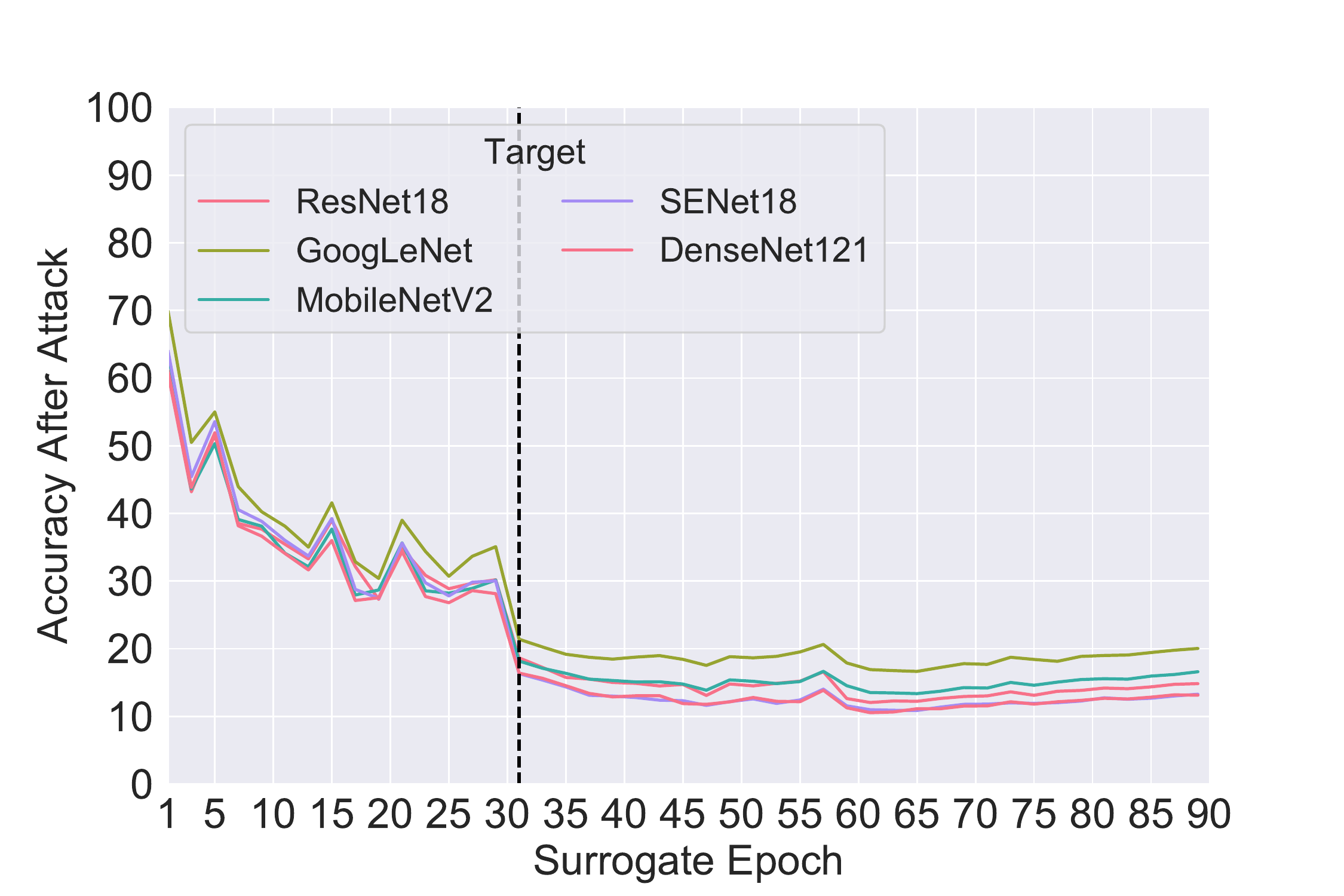}
        \subcaption{SENet18 surrogate model}
        \label{fig:senet_tap_transfer}
    \end{subfigure}
    \newline 
    \begin{subfigure}[b]{\graphsubsize\textwidth}
        \centering
        \includegraphics[width=\columnwidth]{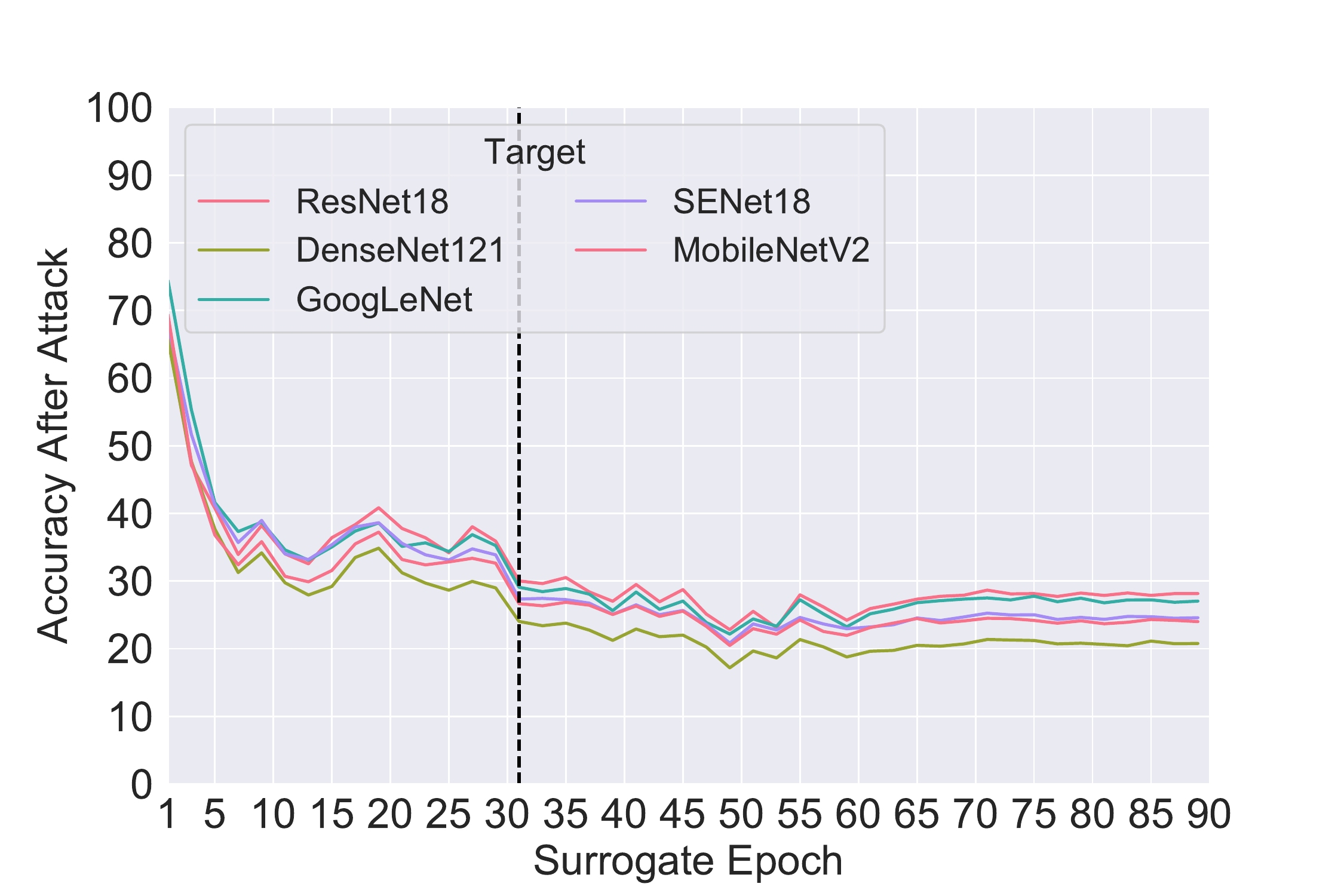}
        \subcaption{DenseNet121 surrogate model}
        \label{fig:densenet_tap_transfer}
    \end{subfigure}
    \caption{Post-attack accuracy for TAP transfer attacks from naturally trained models to target models, ($\epsilon=.05$). Lower accuracy indicates better transfer. The dashed lines indicate the surrogate epoch with the best transferability for most attacks. Note that though the undertraining effect is still present for TAP, the optimal epoch may differ from that for other attacks and not match the dashed lines.}
    \label{fig:tap_transfer}
\end{figure}

\section{Transfer accuracies for ImageNet}

 The ImageNet dataset is a large scale dataset consisting of 1.2 million training images and 150,000 validation and test images divided into 1000 mutually exclusive classes. The images are variably sized, but are resized to 224 by 224 for training and evaluation. We train models with the same architectures as discussed in Section~5, with minor changes to reflect the different input size of ImageNet vs CIFAR10.

\begin{figure}[H]
    \centering
    \begin{subfigure}[b]{\graphsubsize\textwidth}
        \raggedleft
        \includegraphics[width=\columnwidth]{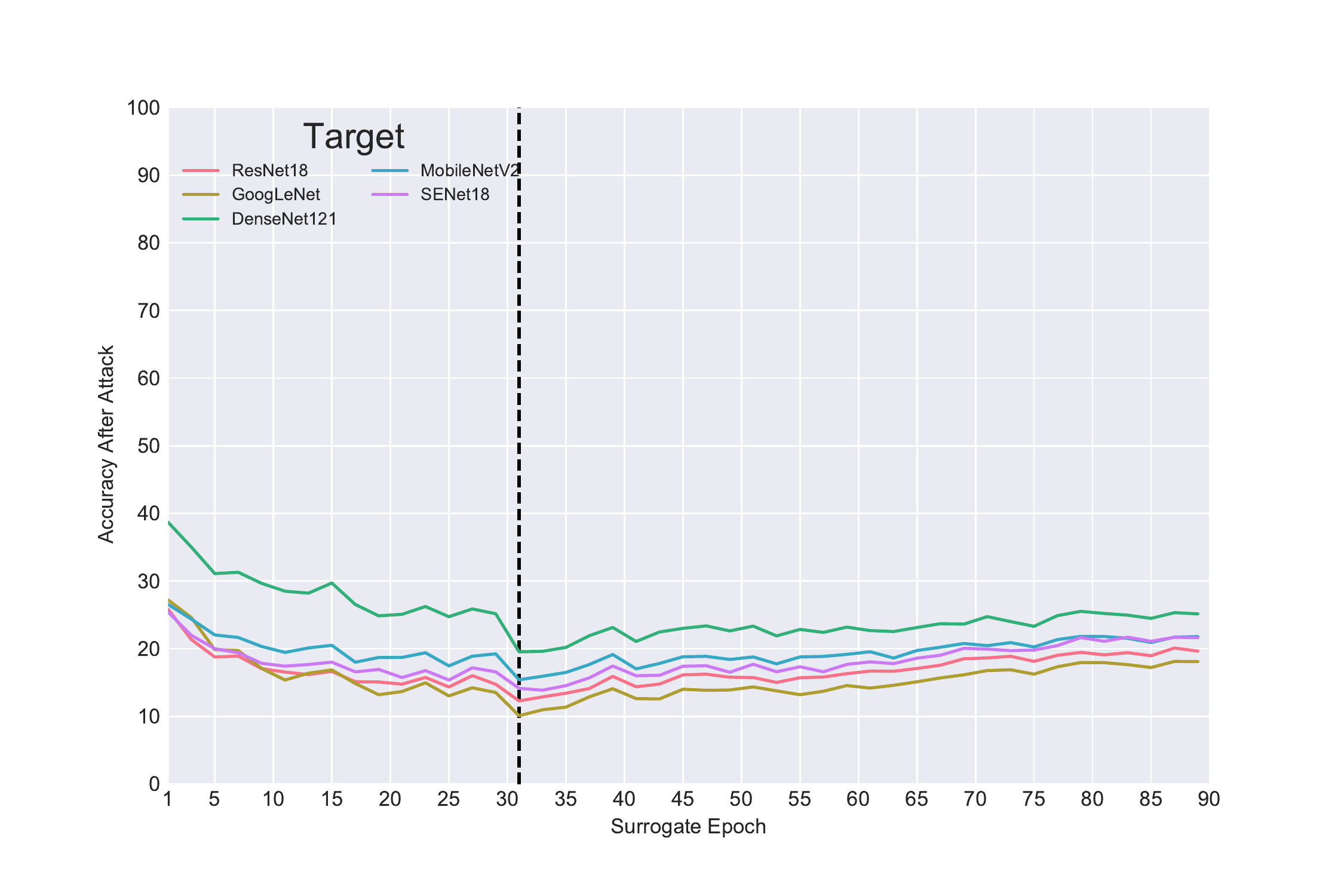}
        \subcaption{GoogLeNet surrogate model}
        \label{fig:googlenet_imagenet}
    \end{subfigure}
    \hfill
    \begin{subfigure}[b]{\graphsubsize\textwidth}
        \raggedright
        \includegraphics[width=\columnwidth]{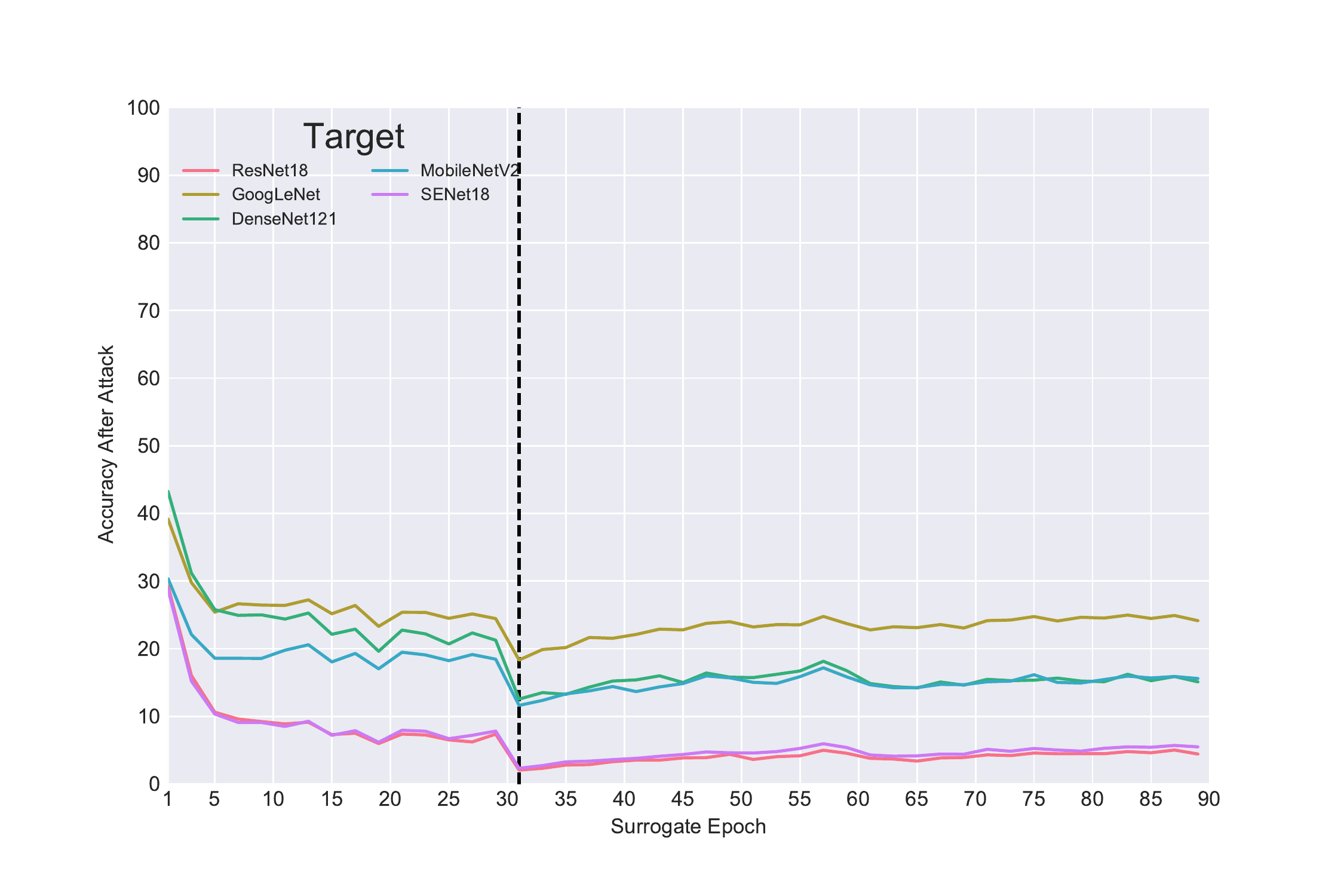}
        \subcaption{ResNet18 surrogate model}
        \label{fig:resnet_imagenet}
    \end{subfigure}
    \newline 
    \begin{subfigure}[b]{\graphsubsize\textwidth}
        \raggedleft
        \includegraphics[width=\columnwidth]{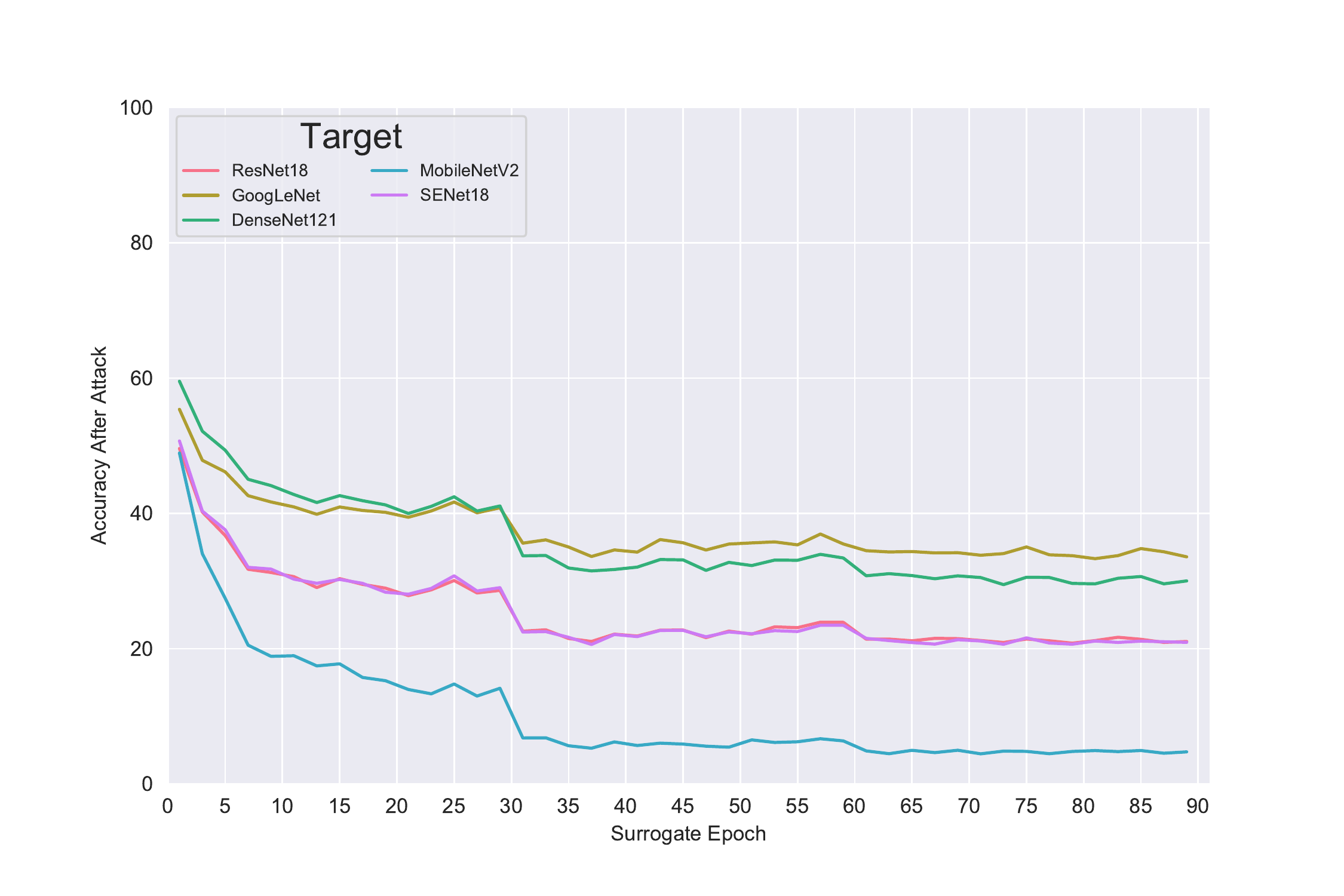}
        \subcaption{MobileNetV2 surrogate model}
        \label{fig:mobilenet_imagenet}
    \end{subfigure}
    \hfill
    \begin{subfigure}[b]{\graphsubsize\textwidth}
        \raggedright
        \includegraphics[width=\columnwidth]{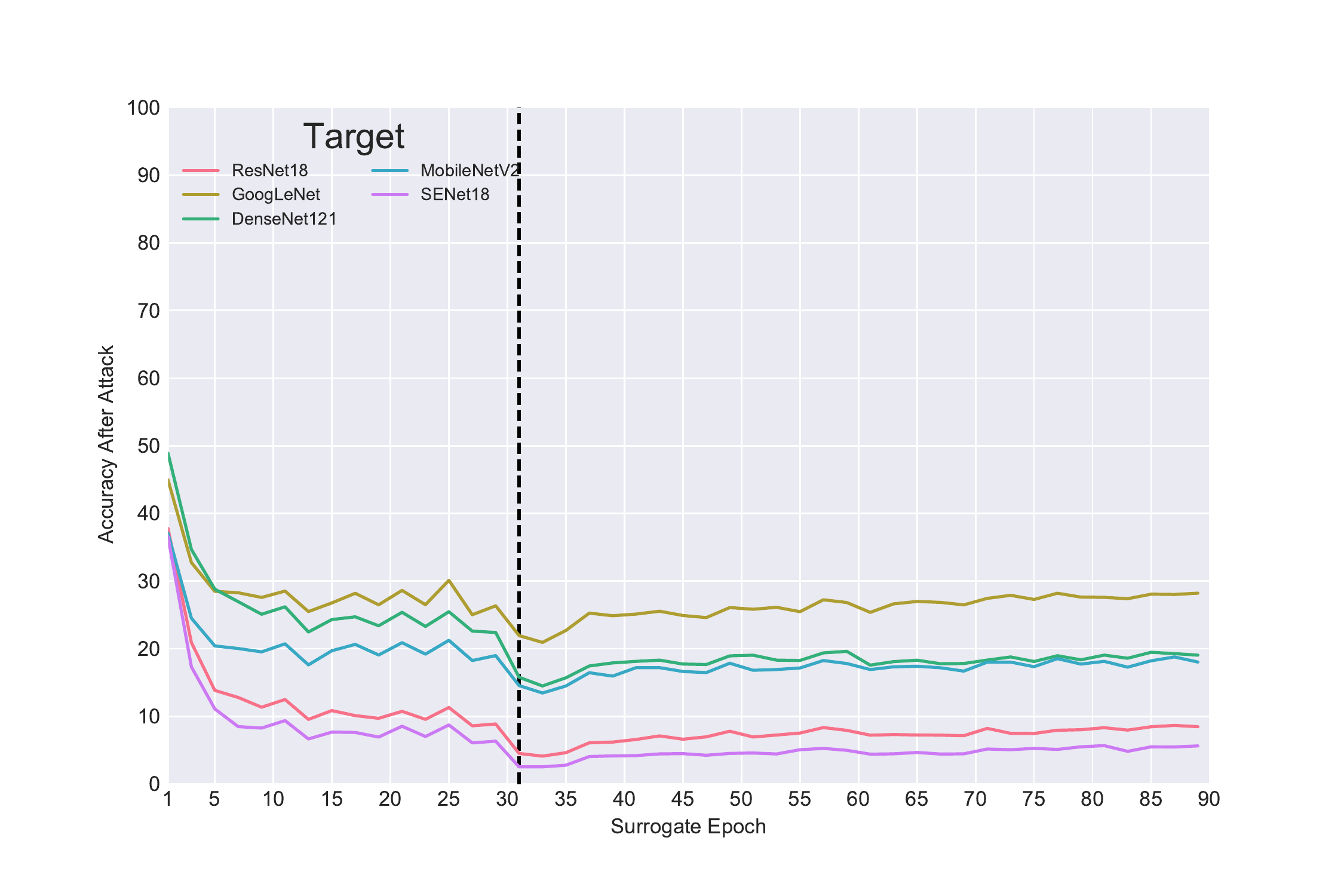}
        \subcaption{SENet18 surrogate model}
        \label{fig:senet_imagenet}
    \end{subfigure}
    \newline 
    \begin{subfigure}[b]{\graphsubsize\textwidth}
        \centering
        \includegraphics[width=\columnwidth]{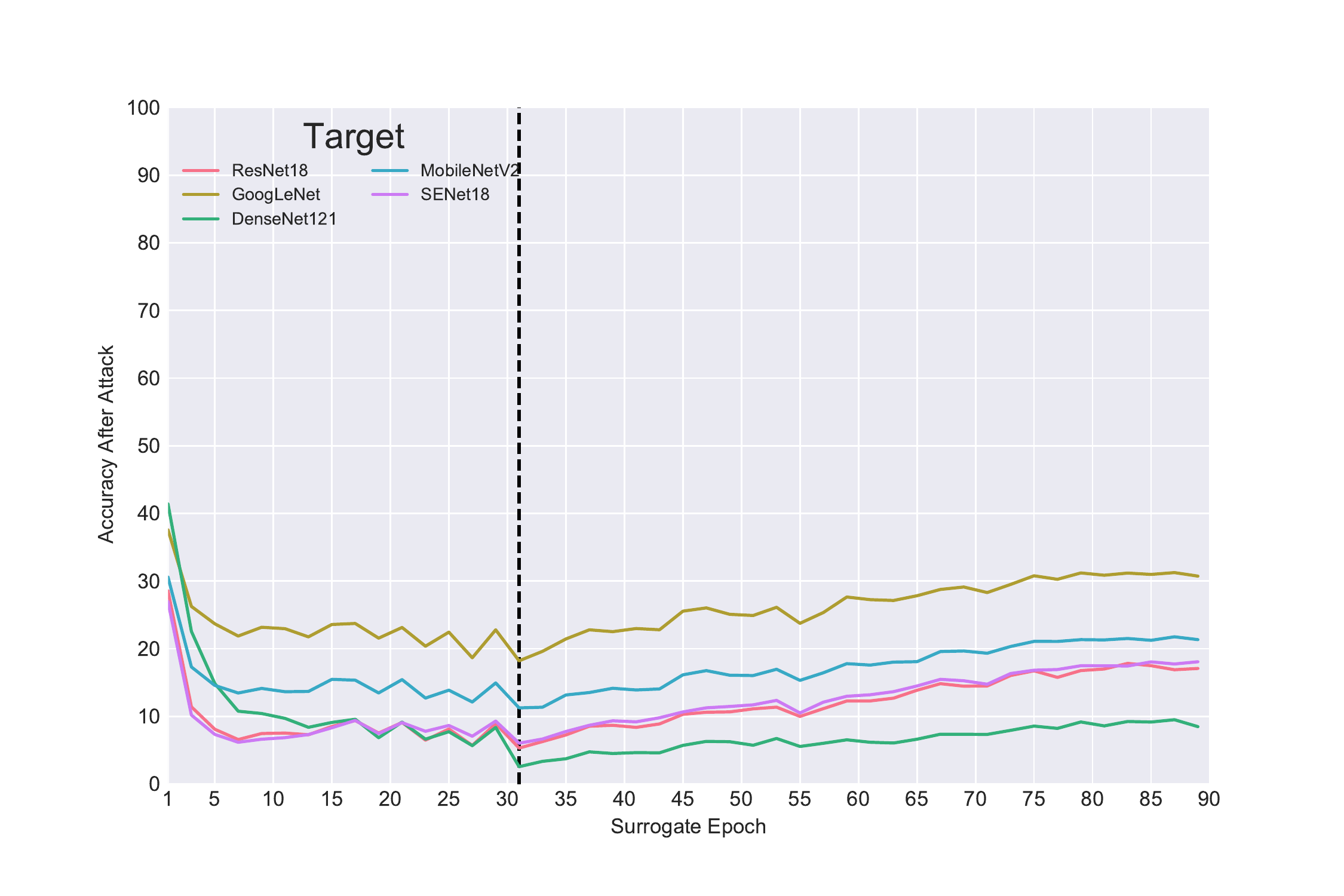}
        \subcaption{DenseNet121 surrogate model}
        \label{fig:densenet_imagenet}
    \end{subfigure}
    
    \caption{Post-attack accuracy for ImageNet MI-FGSM transfer attacks from naturally trained models to target models, ($\epsilon=.05$). Lower accuracy indicates better transfer. The dashed lines indicate the surrogate epoch with the best transferability for most attacks.}
    \label{fig:imagenet_transfer}
\end{figure}

\section{Training Procedure}

\subsection{CIFAR10}
We train surrogate models for $90$ epochs using stochastic gradient descent with momentum. We use a stepped learning rate schedule with an initial learning rate of $0.1$, decreased by a factor of $10$ at epochs $30$ and $60$.

Target models were trained by Huang \etal. for $350$ total epochs, using a learning rate schedule which starts at $0.1$ and reduces by a factor of $10$ at epochs $150$ and $250$. To introduce an additional architecture, we also train a MobileNetV2 model for 350 epochs using the same learning rate schedule. We report validation set performance of all models in Table~\ref{tab:cif_acc}.

\begin{table}[H]
\caption{Model validation set performance of fully trained CIFAR-10 surrogate and target models on clean (non-adversarial) inputs.}
\centering
\begin{tabular}{ lcccc } \\
 \toprule 
 Architecture & Accuracy (Surrogate) & Accuracy (Target)\\
 \midrule
 ResNet18 & 94.29 & 94.77 \\
 SENet18 & 94.44 & 94.59 \\
 MobileNetV2 & 92.23 & 94.06 \\
 GoogLeNet & 94.43 & 94.86 \\
 DenseNet121 & 94.99 & 95.61 \\
 \bottomrule
\end{tabular}
\label{tab:cif_acc}
\end{table}

Robust adversarially trained models were trained following the parameters introduced by \cite{wong2019fast}, we train two robust models for 45 epochs using cyclic learning rates. Our first model achieves clean test accuracy of $79.66\%$, and our second achieves test accuracy of $79.58\%$. The first model achieves white box adversarial accuracy of $41.03\%$ against a 20-iteration I-FGSM attack, and the second gets $40.98\%$ accuracy.

\subsection{ImageNet}
We train the same model architectures for ImageNet as for CIFAR10, with adjustments to accommodate the adjusted image size. We train all surrogate models models for 90 epochs using stochastic gradient descent with momentum, decaying learning rate by a factor of $10$ at epochs $30$ and $60$, using an initial learning rate of $.01$. We use pretrained models provided by Torchvision as target models for all architectures except SENet18, which is not available in the Torchvision library \cite{marcel2010torchvision}. For SENet18, we train a target model using the same training procedure as used for the surrogate models. We report validation set performance (Top-1 and Top-5 accuracy) in Table~\ref{tab:im_acc}.

\begin{table}[H]
\caption{Model validation set performance of fully trained ImageNet surrogate and target models on clean (non-adversarial) inputs.}
\centering
\begin{tabular}{ lcccc } \\
 \toprule % TODO: Fill in
 Architecture & Top 1 (Surrogate) & Top 5 (Surrogate) & Top 1 (Target) & Top 5 (Target)\\
 \midrule
 ResNet18 & 70.16 & 87.95 & 69.76 & 89.08 \\
 SENet18 & 71.30 & 88.57 & 71.49 & 88.68 \\
 MobileNetV2 & 66.11 & 85.71 & 71.88 & 90.29 \\
 GoogLeNet & 69.48 & 87.66 & 69.78 & 89.53 \\
 DenseNet121 & 77.01 & 91.72 & 74.65 & 92.17 \\
 \bottomrule
\end{tabular}
\label{tab:im_acc}
\end{table}

\section{Hyperparameters}

We evaluated results for a variety of hyperparameters to determine the impact of hyperparameter selection on intermediate epoch transferability. We evaluate hyperparameters using our strongest model and attack (ResNet18 and Momentum IFGSM), and transfer to all target models, reporting average accuracy across targets.

\subsection{Batch Size}
\begin{figure}[H]
    \centering
    \includegraphics[width=.6\textwidth]{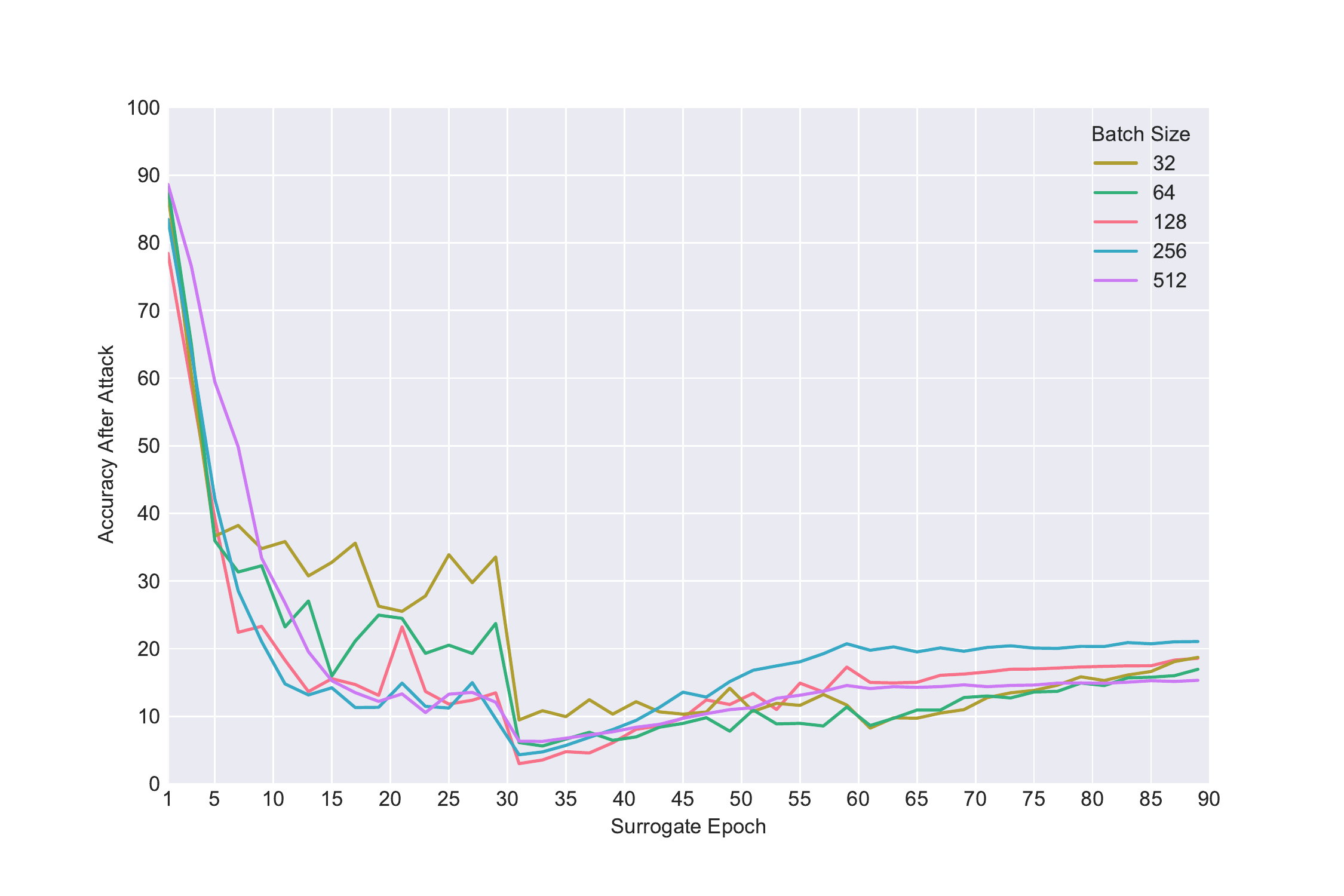}
    \caption{ResNet18 surrogate model using varied training batch sizes}
    \label{fig:batchsize}
\end{figure}

We evaluate ResNet18 surrogate models trained using a variety of batch sizes (Batch Size $\in \{32,\ 64,\ 128,\ 256,\ 512\}$). We find that the effect of intermediate epoch transferability persists across all evaluated surrogate model training batch sizes, and show results in Figure~\ref{fig:batchsize}.

\subsection{Learning Rate}
\begin{figure}[H]
    \centering
    \includegraphics[width=.6\textwidth]{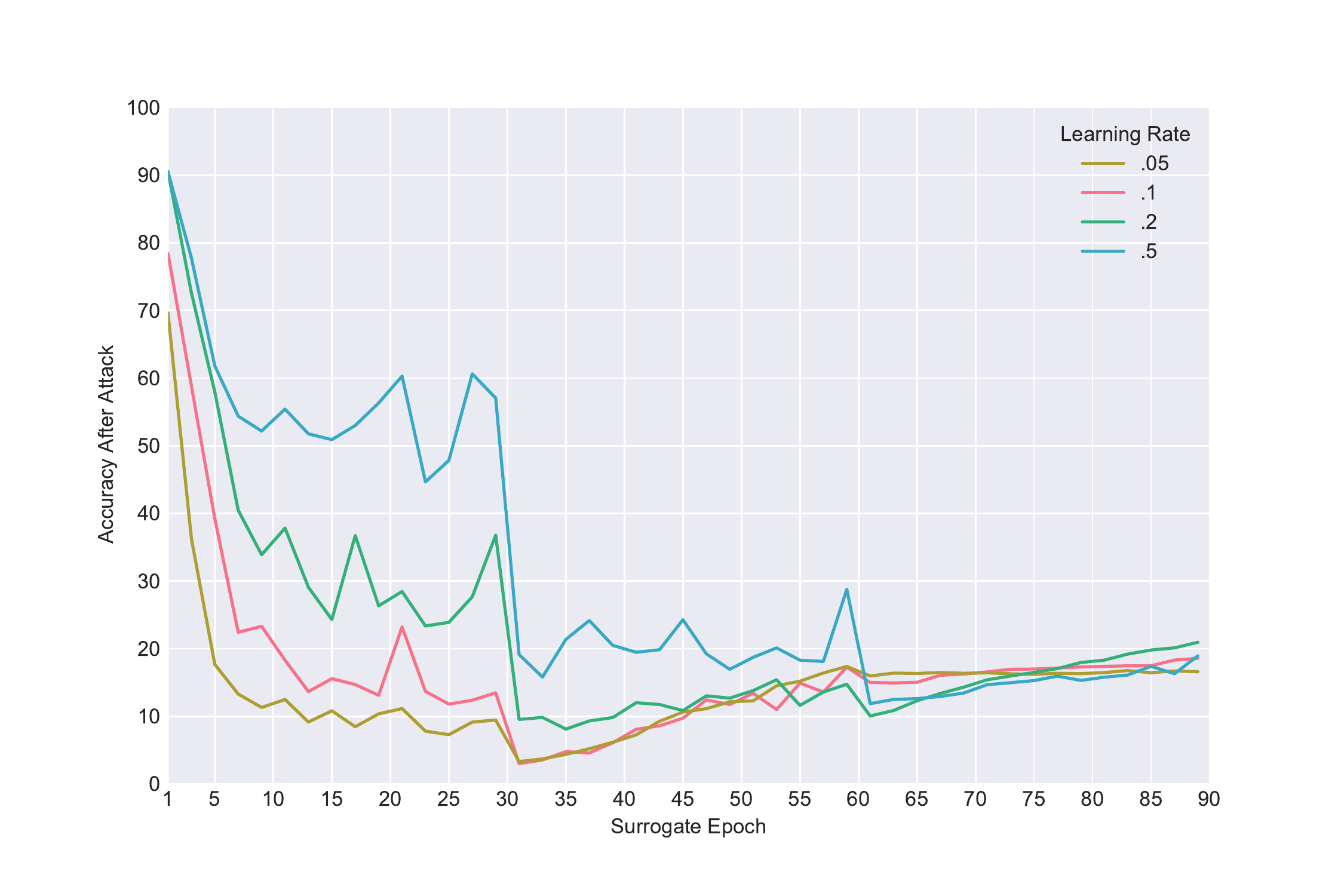}
    \caption{ResNet18 surrogate model using varied training learning rates}
    \label{fig:learningrate}
\end{figure}

We evaluate ResNet18 surrogate models trained using a variety of learning rates (Learning Rate $\in \{.05,\ .01,\ .2,\ .5\}$). Though the transfer curves are more distinct than those depicted for varied batch size in Figure~\ref{fig:batchsize}, all learning rates exhibit the same pattern --- the optimal epoch for transfer (between 30 and 35 or 60 and 65 depending on model) is far before the fully trained epoch (between 80 and 89 for all models). We show results in Figure~\ref{fig:learningrate}.

\subsection{Optimizer}
\begin{figure}[H]
    \centering
    \begin{subfigure}[b]{.48\textwidth}
        \raggedleft
        \includegraphics[width=\columnwidth]{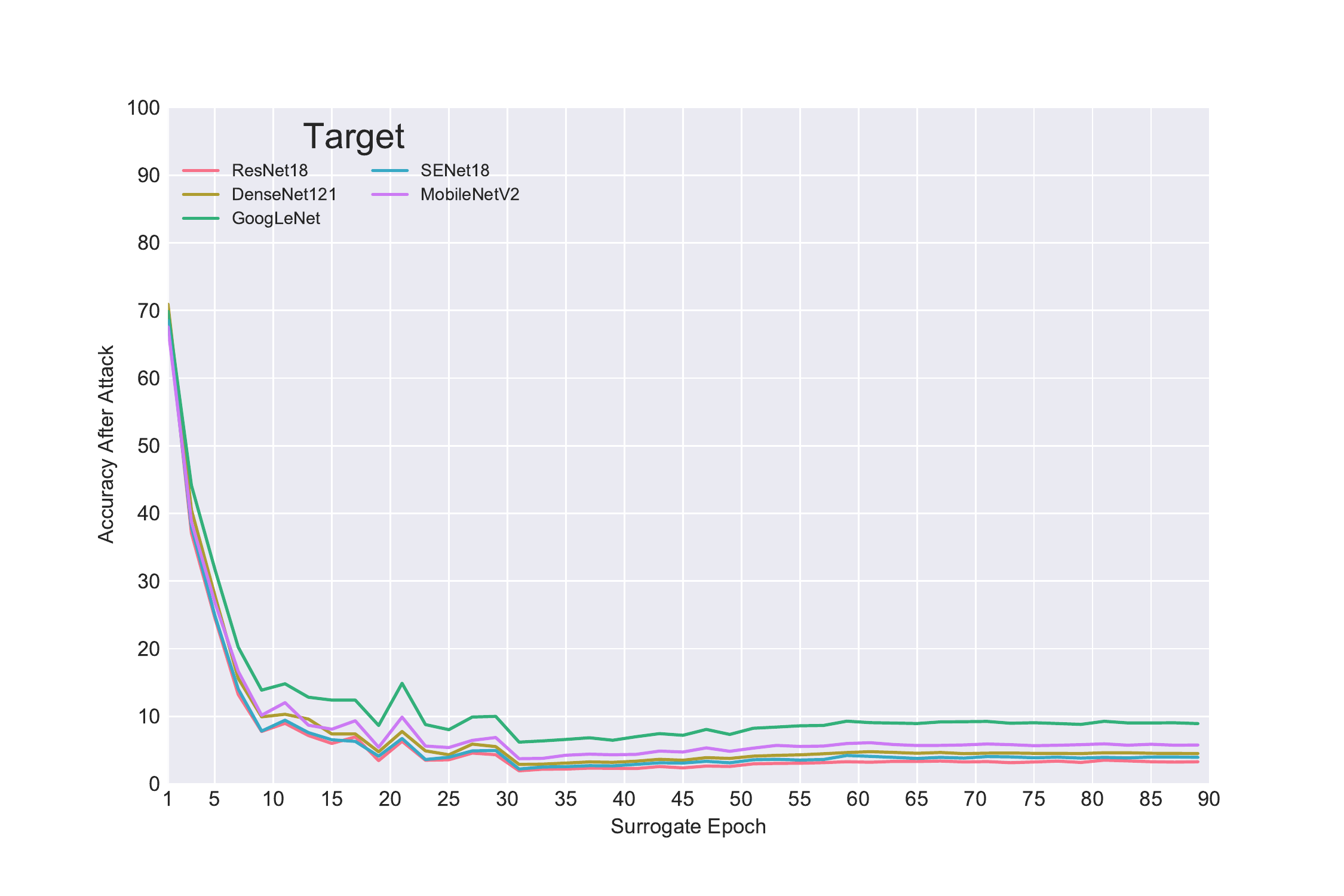}
        \subcaption{ResNet18 trained using SGD with no momentum}
        \label{fig:sgd_no_momentum}
    \end{subfigure}
    \hfill
    \begin{subfigure}[b]{.48\textwidth}
        \raggedright
        \includegraphics[width=\columnwidth]{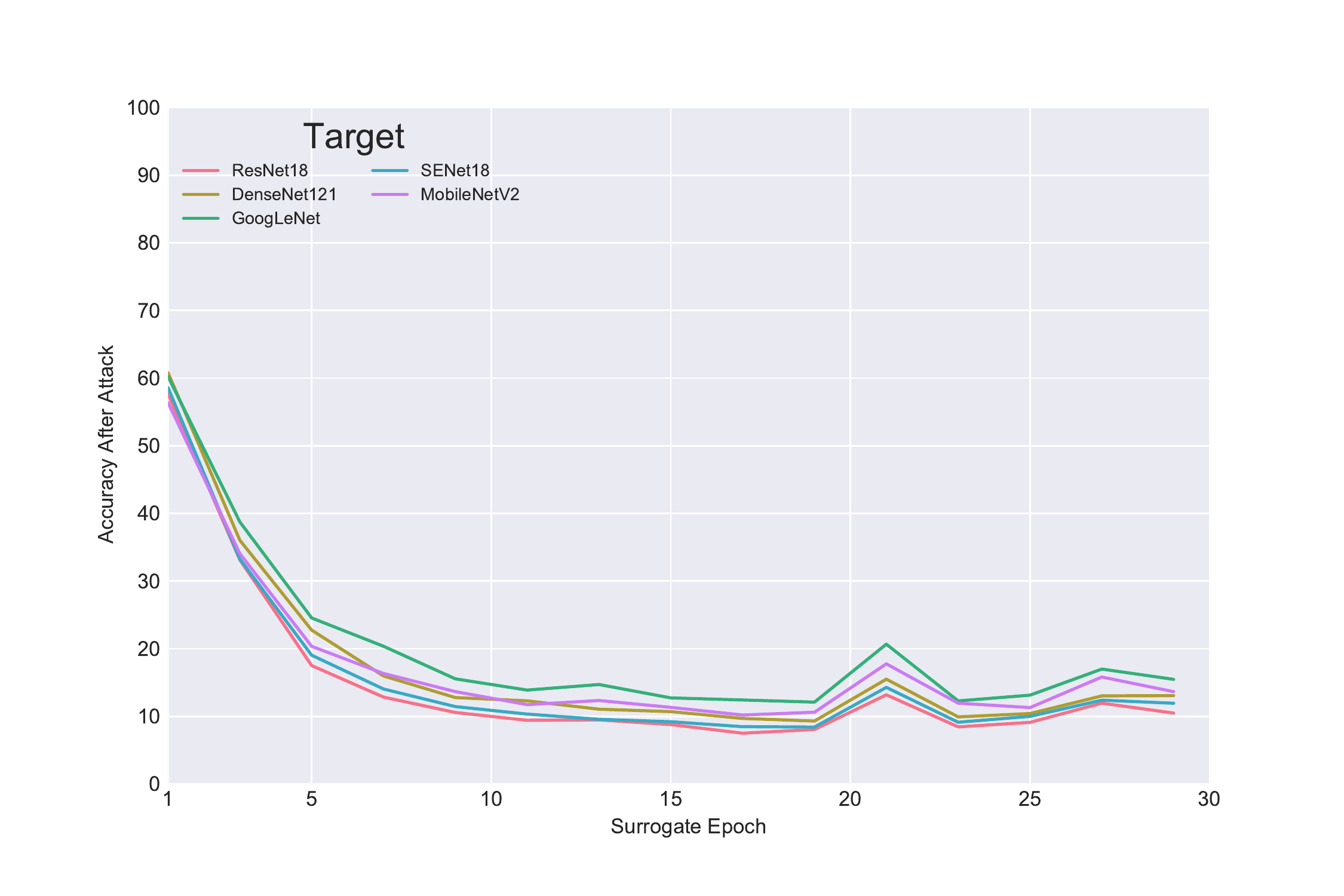}
        \subcaption{ResNet18 trained using the Adam optimizer}
        \label{fig:adam}
    \end{subfigure}
    
    \caption{Post-attack accuracy for transfer attacks from naturally trained ResNet18 models with varied optimizers to target models, ($\epsilon=.05$). Lower accuracy indicates better transfer. The dashed lines indicate the surrogate epoch with the best transferability for most attacks.}
    \label{fig:optimizers}
\end{figure}

We evaluated results using models trained with alternative optimizers. We found that optimizer selection does impact results, and show results in Figure~\ref{fig:optimizers}. Surrogate epoch 87 is the fully trained model for SGD with no momentum, and surrogate epoch 27 is the fully trained model for Adam. We find that intermediate epoch transferability is consistent for SGD with no momentum, but is less present for the Adam optimizer. We suggest that optimizers which use parameter-level learning rates (such as Adam) do not experience this effect strongly. 

\begin{figure}[H]
    \centering
    \includegraphics[width=.6\textwidth]{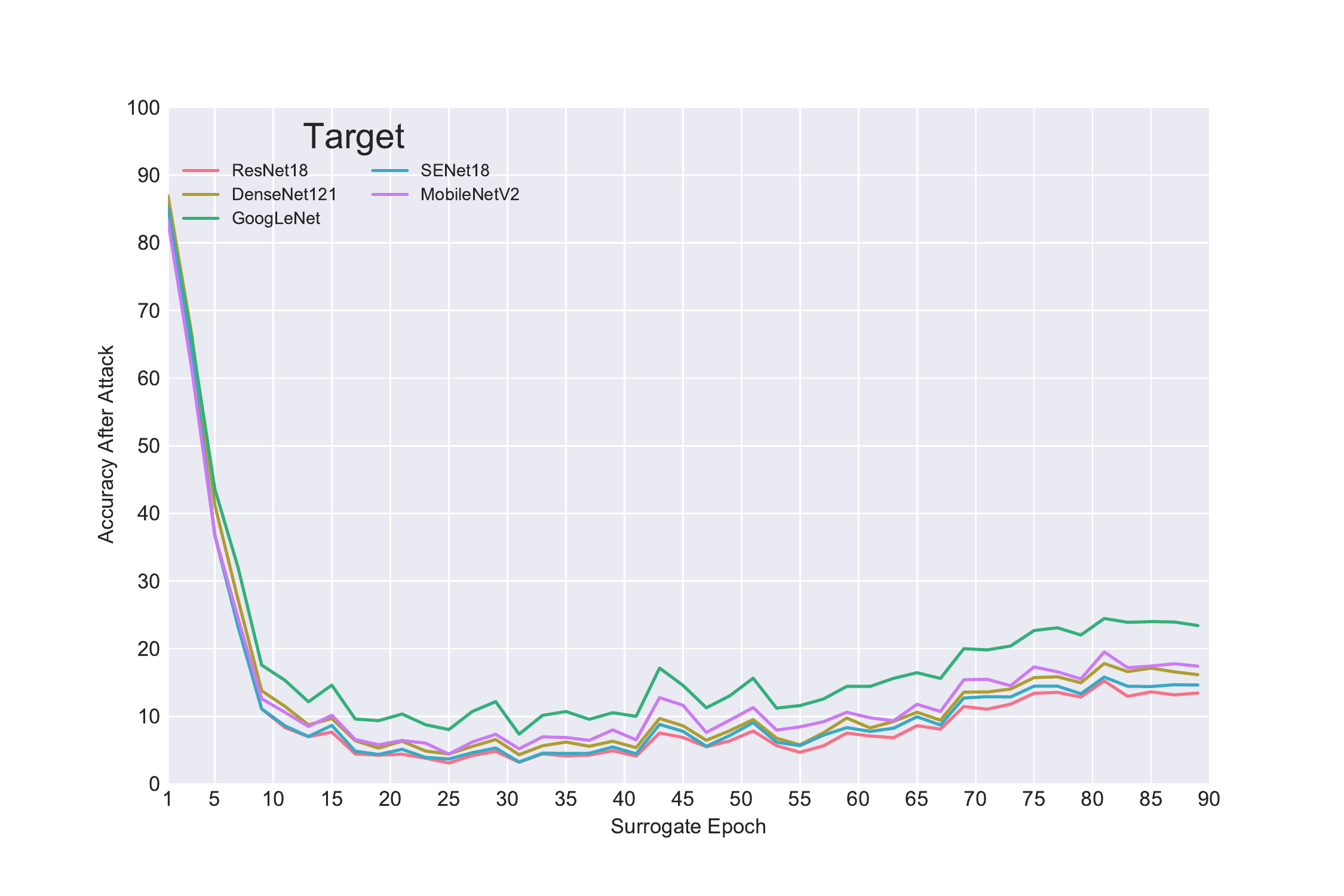}
    \caption{ResNet18 surrogate model using SGD with cyclic learning rates}
    \label{fig:cyclicsgd}
\end{figure}

We evaluate cyclic learning rates to test this effect, and find that optimal surrogate efficacy is impacted by global learning rate decay events. As shown in Figure~\ref{fig:cyclicsgd}, transferability for a model trained using Cyclic learning rates goes up and down in a pattern following the learning rate. 

We suggest that this is because during SGD training, learning rate decay after a plateau in validation set accuracy typically produces a stronger model (as the model enters a better local optimum in the loss landscape). After the decay, the model begins to overfit to the local loss landscape, and adversarial images generated from the surrogate model become less transferable.

% cite adam and cyclic

\section{Correlation between gradient similarity and transferability}
\begin{table}[H]
\caption{Pearson correlation coefficients between gradient similarity and post-attack accuracy for all epochs of each surrogate/target pair. * indicates $p < .05$, ** indicates $p < .01$, and *** indicates $p < .001$.}
\centering
\begin{tabular}{ lccccc } \\
 \toprule % TODO: Fill in
 Source & & & Target & & \\
 \midrule
 & ResNet18 & SENet18 & MobileNetV2 & GoogLeNet & DenseNet121\\
 \midrule
 ResNet18 & -0.58*** & -0.57*** & -0.75*** & -0.76*** & -0.84*** \\
 SENet18 & -0.68*** & -0.63*** & -0.70*** & -0.69*** & -0.76*** \\
 MobileNetV2 & -0.95*** & -0.96*** & -0.97*** & -0.99*** & -0.97*** \\
 GoogLeNet & -0.64*** & -0.63*** & -0.60*** & -0.70*** & -0.55*** \\
 DenseNet121 & -0.58*** & -0.60*** & -0.78*** & -0.53*** & -0.59*** \\
 \bottomrule
\end{tabular}
\label{tab:gradient_correlation}
\end{table}

%\subsection*{Linear Model}
%We include gradient similarity, squared gradient similarity, loss curvature, and a source model constant. We include squared gradient similarity due to the presence of nonlinear residuals in a model without squared gradient similarity. 

\section{Comparison to other surrogates}
To address the possibility that our final epoch models are overtrained, and do not represent typical models, we also evaluate attacks between our target models. We use each target model as a surrogate, and evaluate the transferability of attacks generated using that target model to other target models. All models except for the MobileNetV2 architecture were released as pretrained models by \citet{huang2019enhancing}, who used these models as their target and surrogate models, ensuring that our comparison here is a fair comparison to prior work. All models were trained using a open-sourced training code and hyperparameters provided by \cite{pytorchtraining}. When evaluated as surrogates, these models perform very similarly to the best loss epochs of our separately trained surrogate models, and our undertrained surrogate models outperform them by a wide margin. We report the results of this analysis in SM in Table~S4. This indicates that our best loss models are representative of typical surrogate models for CIFAR-10.

\begin{table*}[h]
\scriptsize
%\footnotesize 
\caption{Accuracy (\%) after attack by given CIFAR-10 target model (column) for attacks generated using each of the other target models. We omit the diagonal, since the diagonal is a white box attack. $\epsilon = .05$. Lower accuracy indicates better transfer.}
\centering
\begin{tabular}{ llccccc } \\
 \toprule 
 Attack & Source & & & Target & & \\
 \midrule
 & & ResNet18 & SENet18 & MobileNetV2 & GoogLeNet & DenseNet121\\
 \midrule
 MI-FGSM & ResNet18 & --- & 17.90 & 21.87 & 29.06 & 19.64 \\
 & SENet18 & 12.28 & --- & 16.37 & 20.10 & 12.87 \\
 & MobileNetV2 & 17.64 & 18.31 & --- & 21.89 & 16.78 \\
 & GoogLeNet & 26.80 & 18.31 & 27.87 & --- & 22.93 \\
 & DenseNet121 & 11.42 & 10.78 & 12.13 & 12.67 & --- \\
 \midrule
 I-FGSM & ResNet18 & --- & 27.35 & 31.45 & 42.81 & 29.34 \\
 & SENet18 & 16.06 & --- & 21.00 & 25.88 & 17.26 \\
 & MobileNetV2 & 23.41 & 24.60 & --- & 28.33 & 22.07 \\
 & GoogLeNet & 36.90 & 35.61 & 36.73 & --- & 31.29 \\
 & DenseNet121 & 21.65 & 20.72 & 21.95 & 24.10 & --- \\
 \midrule
 ILA MI-FGSM & ResNet18 & --- & 7.95 & 9.60 & 12.96 & 8.41 \\
 & SENet18 & 7.07 & --- & 8.56 & 11.29 & 7.36 \\
 & MobileNetV2 & 27.24 & 29.00 & --- & 33.45 & 26.60 \\
 & GoogLeNet & 11.47 & 11.56 & 12.19 & --- & 8.56 \\
 & DenseNet121 & 6.91 & 6.59 & 7.01 & 7.87 & --- \\
 \midrule
 ILA I-FGSM & ResNet18 & --- & 7.67 & 9.35 & 13.86 & 7.74 \\
 & SENet18 & 6.66 & --- & 8.14 & 10.98 & 6.88 \\
 & MobileNetV2 & 28.04 & 29.66 & --- & 34.17 & 26.94 \\
 & GoogLeNet & 12.52 & 12.14 & 13.05 & --- & 9.05 \\
 & DenseNet121 & 6.43 & 6.16 & 6.73 & 7.62 & --- \\
 \midrule
  TAP & ResNet18 & --- & 20.66 & 21.55 & 25.70 & 20.72 \\
 & SENet18 & 15.14 & --- & 18.37 &  21.58 & 15.81 \\
 & MobileNetV2 & 28.92 & 29.64 & --- & 34.65 & 27.42 \\
 & GoogLeNet & 17.79 & 17.82 & 17.10 & --- & 15.37 \\
 & DenseNet121 & 20.37 & 19.49 & 20.15 & 20.46 & --- \\
 \bottomrule
\end{tabular}
\label{tab:comparison_transfer}
\end{table*}

{\small
\bibliographystyle{plainnat}
\bibliography{egbib}
}